%% file: main.tex
\documentclass[fleqn]{article}
\usepackage[english]{babel}
\usepackage[utf8]{inputenc}
\usepackage{amsmath}
\usepackage{multirow}
\usepackage{amssymb}
\usepackage{booktabs}
\usepackage{johd}
\usepackage{algorithm}
\usepackage{algpseudocode}
\usepackage{cancel}
\usepackage{graphicx}
\usepackage{subcaption}
\usepackage{subcaption}
\usepackage{pgffor}
\usepackage{lipsum} 
\usepackage{soul}
\usepackage{pifont}
\usepackage{cancel}
\usepackage{xcolor}
\usepackage{colortbl}
\usepackage[toc]{appendix}
\newcommand{\cmark}{\ding{51}}  
\newcommand{\xmark}{\ding{55}}  
\PassOptionsToPackage{hidelinks}{hyperref}

\title{Know Unreported Roadway Incidents in Real-time: Early Traffic Anomaly Detection}

\author{Haocheng Duan$^{a}$, Hao Wu$^{a}$, Sean Qian$^{a}$$^{b}$$^{*}$\\
        \small $^{a}$Department of Civil and Environmental Engineering, Carnegie Mellon University \\
        \small $^{b}$Heinz College, Carnegie Mellon University \\
        \small $^{*}$Corresponding author: Sean Qian; \tt{seanqian@cmu.edu} \\
}
\date{}
\begin{document}

\pagenumbering{roman}

\pagenumbering{arabic}  
\setcounter{page}{1}    
\maketitle
\begin{abstract}
This research aims to know traffic anomalies as early as possible. A traffic anomaly refers to a generic incident on the road that influences traffic flow and calls for urgent traffic management measures. `Knowing'' the occurrence of a traffic anomaly is twofold: the ability to detect this anomaly before it is reported anywhere, or it may be such that an anomaly can be predicted before it actually occurs on the road (e.g., non-recurrent traffic breakdown). In either way, the objective is to inform traffic operators of unreported incidents in real time and as early as possible. The key is to stay ahead of the curve. Time is of the essence. 

Conventional automatic incident detection (AID) methods often struggle with early detection due to their limited consideration of spatial effects and early-stage characteristics. Therefore, we propose a deep learning framework utilizing prior domain knowledge and model-designing strategies. This allows the model to detect a broader range of anomalies, not only incidents that significantly influence traffic flow but also early characteristics of incidents along with historically unreported anomalies. We specially design the model to target the early-stage detection/prediction of an incident. Additionally, unlike most conventional AID studies, our method is highly scalable and generalizable, as it is fully automated with no manual selection of historical reports required, relies solely on widely available low-cost data, and requires no additional detectors. The experimental results across numerous road segments on different maps demonstrate that our model leads to more effective and early anomaly detection. 

\end{abstract}

\noindent\keywords{Automatic Incident Detection, Non-recurrent Traffic, Anomaly Detection, Incident Management}\\
\begin{flushleft}
\section{Introduction}
Traffic congestion grievously disrupts the everyday lives of urban residents and causes substantial economic losses. It can be categorized into recurring and non-recurring congestion. Recurring congestion occurs periodically when traffic volume exceeds the road's capacity. In contrast, non-recurring congestion, also known as \textit{traffic anomalies}, is caused by incidents such as crashes, work zones, and special events. In the U.S., non-recurring congestion accounts for over half of the total congestion \citep{fhwa_congestion_2023}. Mitigating non-recurring impacts requires accurately knowing roadway incidents in advance and proactive management strategies as opposed to being reactive. Figure \ref{fig: incident-timeline} illustrates the FHWA timeline for managing non-recurring events, highlighting that the effectiveness of proactive strategies heavily relies on the timely detection and verification of incidents. However, incident reports in the real world are typically delayed and subject to multiple layers of verification, not to mention a large number of unreported incidents. Acknowledging that incident reports are oftentimes late or missing, traffic operators may find their actions too late by the time they are informed of non-recurrent congestion. To assist traffic operators in implementing real-time control measures \citep{ahmed2015integrated, sheu2002stochastic, ke2024real} and travelers in planning their journeys \citep{ke2024interpretable, xie2020deep}, it is crucial to verify incident reports more promptly, or better yet, detect incidents before they are reported and effectively identify unreported incidents. Our goal is to alert traffic operators of an anomaly before $T_1$, the current practice, by as much as we can, and ideally even before $T_0$ referred in Figure \ref{fig: incident-timeline} in situations where the risk of roadway incidents or non-recurrent traffic breakdown is extremely high.
\begin{figure}
      \centering
      \includegraphics[width=0.5\linewidth]{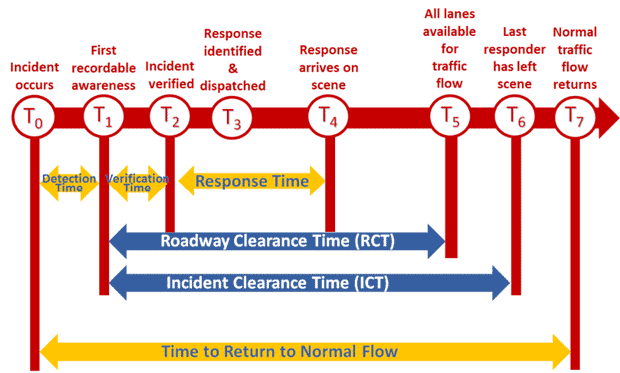}
      \caption{Incident Management Timeline \citep{fhwa2015step1}}
      \label{fig: incident-timeline}
\end{figure}

Conventional automatic incident detection (AID) methods may assist in verifying incident reports, but it is unclear if AID is able to detect incidents earlier than the reports themselves \textit{after the fact} and efficiently identify unreported incidents. This is because they are typically tuned/trained and evaluated based solely on incident reports, thereby inheriting the same delays and missing issues. These issues will be further discussed in detail later in this section. This paper aims to overcome the bottlenecks caused by incident reports in conventional AID methods and management strategies, with the goal of detecting or predicting anomalies as early as possible. Here, we use ``early anomaly detection'' to indicate that our problem is fundamentally different from conventional AID and from network traffic speed/counts prediction. An anomaly in this paper is defined not only as a reported incident but also as any incident that may not be reported at the time, potentially reported later, or not reported at all, yet significantly impacts traffic flow in the near term. Efficiently detecting all these anomalies, especially unreported incidents and incidents that have not yet been reported, is essential for the timely implementation of proactive control measures and the overall incident management process.

Given the exact anomaly occurrence time (namely $T_0$ referred in Figure \ref{fig: incident-timeline}) is usually unknown in practice, we do not necessarily differentiate ``detecting'' an ongoing but unreported anomaly and ``predicting'' an anomaly that has not occurred yet. The goal is to know an unreported anomaly in advance, regardless of whether it has occurred. It is critical to clarify we do not intend to ``predict'' the occurrence of an incident before it actually occurs, since generally the incident occurrence is rare and random. Our intention is to know the anomaly as early as we can, even if there is very little sign of an incident yet. This means that at the time of anomaly detection, it is likely that an incident has occurred but not reported anywhere (including Waze and social media, which is considered more timely than other traditional manners). It is also possible (but unlikely) that an incident may not have occurred, but a certain traffic/weather condition may lead to high risks resulting in its occurrence in the near future. Again, our algorithm does not differentiate this ``detection'' or ``prediction'' nature, thus lending its name to ``early anomaly detection''. We also do not intend to detect or predict the attributes of an incident, its type, duration, etc. 

Conventional AID methods struggle with early anomaly detection. Over the decades, AID has evolved from traditional statistical and comparative approaches to AI-driven methods. However, regardless of the technique used, these methods rely on mining differences in traffic states with and without incidents. Incident reports, being the most intuitive reference for incidents, thus are almost universally used to calibrate or train AID models. Yet, incident reports are not equivalent to anomalies. Real-world incident reports often contain numerous false positives, inaccurate information, or incidents that do not impact traffic conditions. Directly using them to calibrate models can introduce significant bias and even lead to convergence issues during model training. Therefore, existing research often involves manual review \citep{dia1997development, abdulhai1999enhancing, jin2002development} and even camera calibration \citep{chakraborty2019data} to filter out false or insignificant reports. This approach is neither feasible for scaling nor capable of addressing other significant inherent issues, especially delays and missing.

The issue of delayed incident reports arises from the long reaction time between the incident occurrence and when they are reported \citep{gu2016twitter, elsahly2022systematic, coursey2024ft}. Figure \ref{fig: report_late} illustrates the delay of incident reports by displaying the relationship between Waze reports and speed on an I-70E highway segment in Howard County, MD. The blue line represents the speed on the day of the incident report, while the yellow and green dotted lines represent the speed on the same days of the week and at the same time, providing a recurrent reference. It can be seen from the figure that Waze incident reports are significantly delayed compared to the impact on speed. As a result of overlooking delays in incident reports, those features prior to incident reports are often missing from training an AID model. Due to the lack of training samples in the early stage of an incident, the model is unable to learn from these features signing an upcoming or ongoing incident, hindering timely detection. It is worth noting that many studies emphasize achieving near-zero mean time to detection (MTTD) as evidence of timely detection: 
\begin{eqnarray}
\text{MTTD} & = & \frac{\sum_{i=1}^{N_{\text{detected}}} (t_i^{\text{alarm}} - t_{i}^{\text{report}})}{N_{\text{detected}}}\label{eqn: MTTD}
\end{eqnarray}
where $N_{\text{detected}}$ refers to the number of incident reports being detected, $t_i^{\text{report}}$ refers to the start time of the report, and $t_i^{\text{alarm}}$ refers to the time when the model begins to trigger an alert. However, this is because their evaluation process compares detection times to delayed incident reports, so a near-zero MTTD simply indicates that the alarm is triggered shortly after the incident is reported, not when it actually occurs.
\begin{figure}[!htb]
    \centering
    \begin{minipage}{0.61\textwidth}
        \centering
        \includegraphics[width=\textwidth]{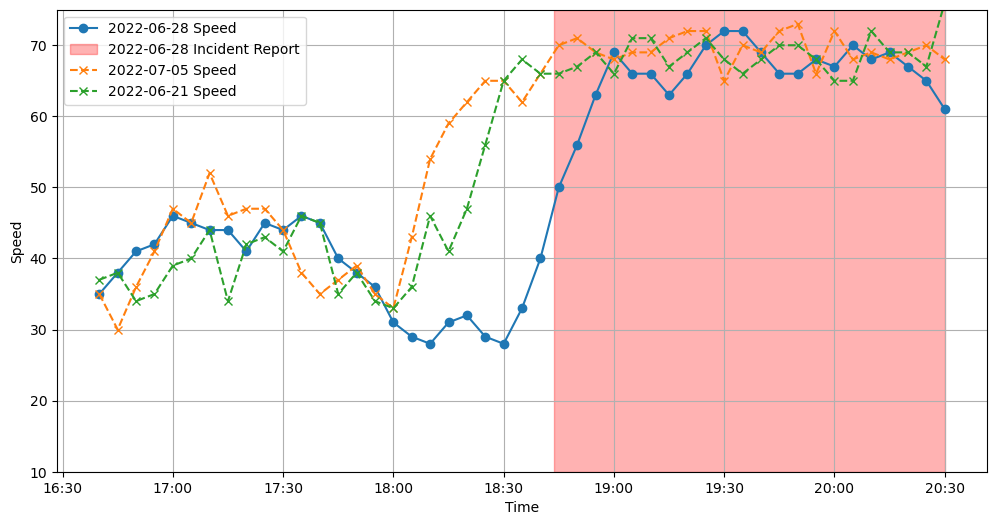}
        \subcaption{2022-06-28 Case}
    \end{minipage}
    \hfill
    \begin{minipage}{0.61\textwidth}
        \centering
        \includegraphics[width=\textwidth]{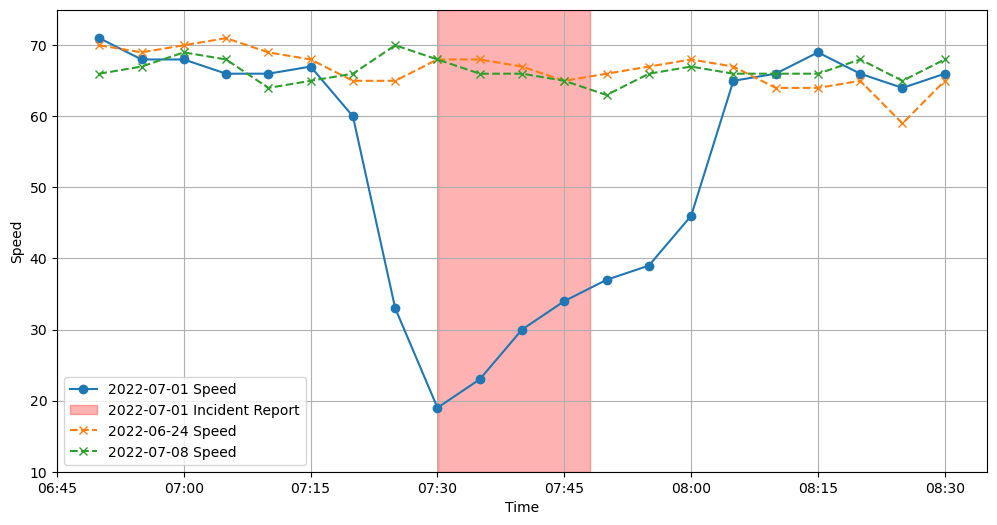}
        \subcaption{2022-07-01 Case}
    \end{minipage}
    \hfill
    \begin{minipage}{0.61\textwidth}
        \centering
        \includegraphics[width=\textwidth]{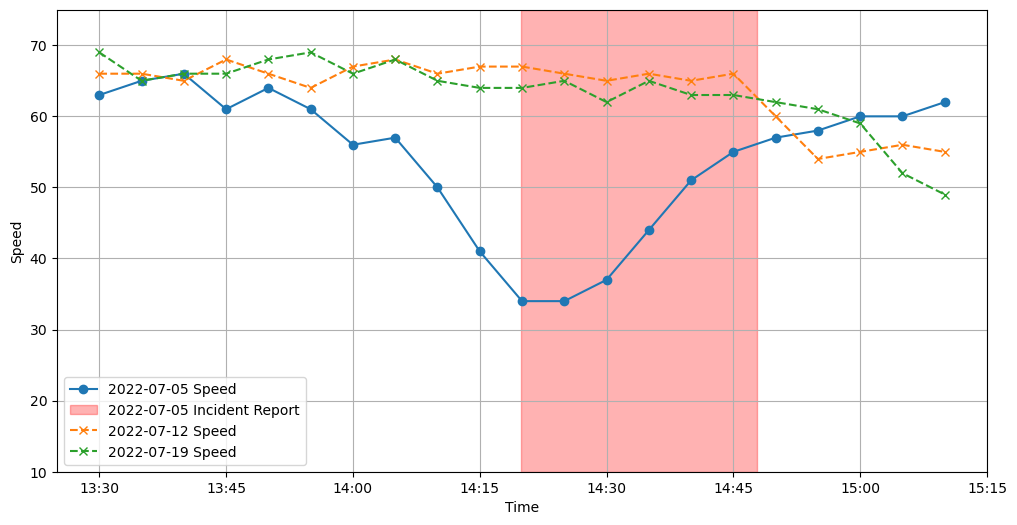}
        \subcaption{2022-07-05 Case}
    \end{minipage}
        \hfill
    \begin{minipage}{0.61\textwidth}
        \centering
        \includegraphics[width=\textwidth]{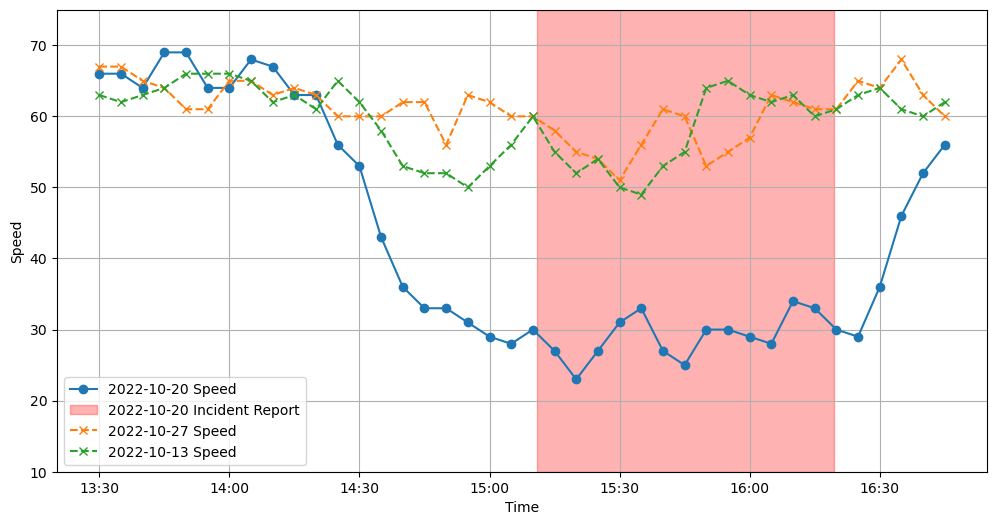}
        \subcaption{2022-10-20 Case}
    \end{minipage}
    \caption{Time-varying speeds on a road segment: Waze incident reports were significantly delayed}
    \label{fig: report_late}
\end{figure}

\begin{figure}[!htb]
    \centering
    \begin{minipage}{0.61\textwidth}
        \centering
        \includegraphics[width=\textwidth]{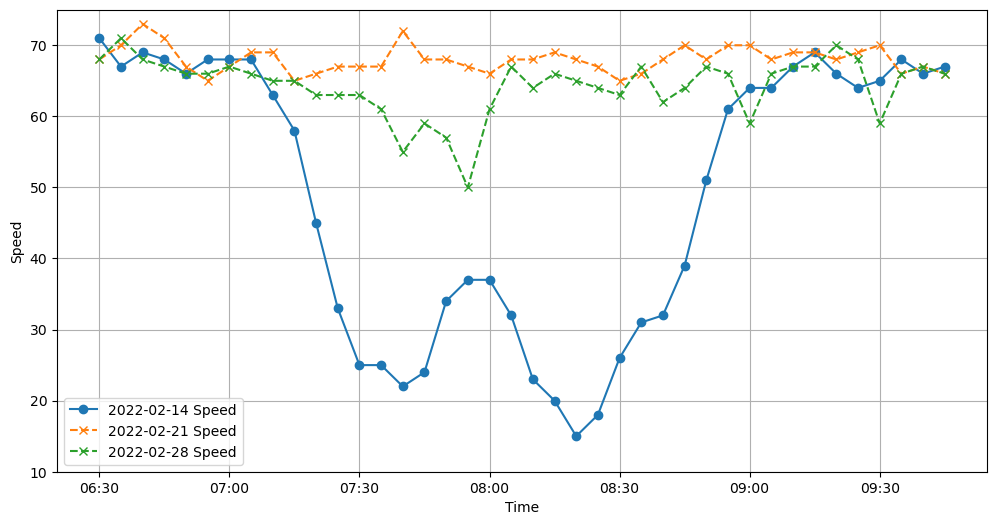}
        \subcaption{2022-02-14 Case}
    \end{minipage}
    \hfill
    \begin{minipage}{0.61\textwidth}
        \centering
        \includegraphics[width=\textwidth]{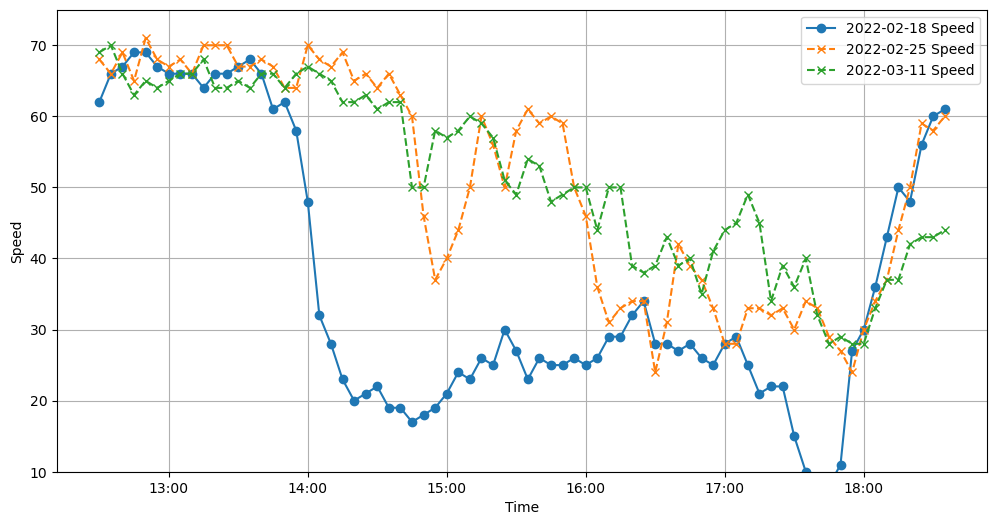}
        \subcaption{2022-02-18 Case}
    \end{minipage}
    \hfill
    \begin{minipage}{0.61\textwidth}
        \centering
        \includegraphics[width=\textwidth]{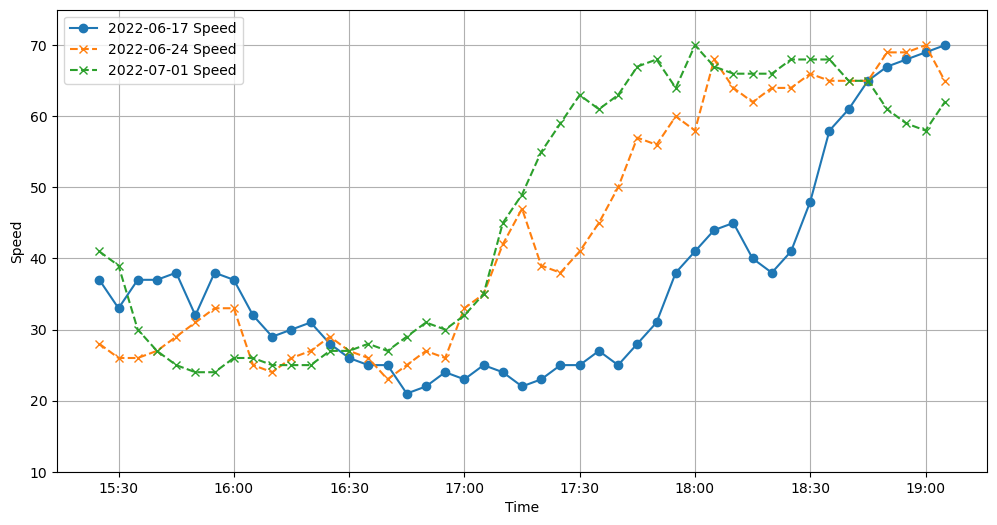}
        \subcaption{2022-06-17 Case}
    \end{minipage}
        \hfill
    \begin{minipage}{0.61\textwidth}
        \centering
        \includegraphics[width=\textwidth]{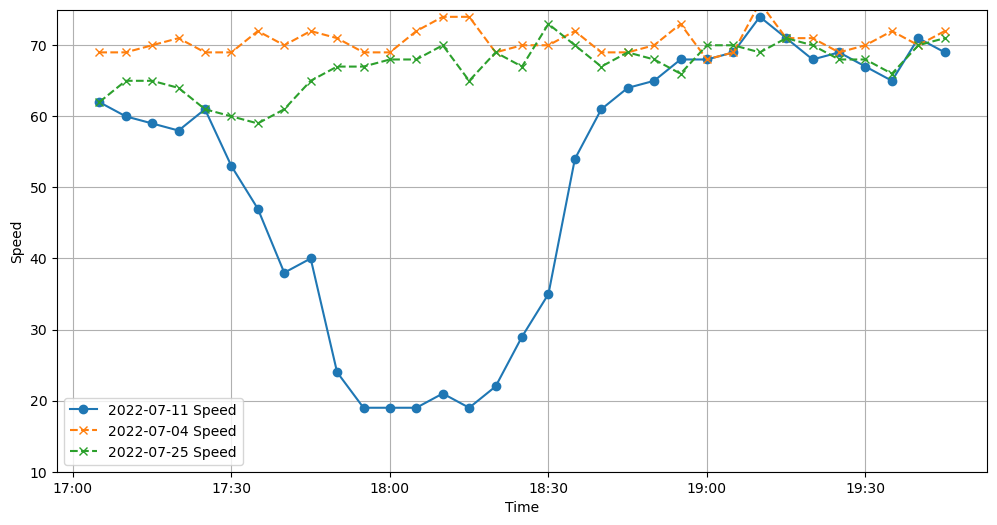}
        \subcaption{2022-07-11 Case}
    \end{minipage}
    \caption{Time-varying speeds on a road segment: Incident reports were missing}
    \label{fig: report_missing}
\end{figure}

In addition, the issue of missing incident reports is mainly because incident reports still largely rely on either fixed location sensors that are extremely limited or passing drivers to report actively (e.g., Waze, X, etc.). In fact, many incidents or traffic anomalies go unreported \citep{turner2015road}. Figure \ref{fig: report_missing} illustrates examples of significant anomalies that were not reported on the same segment as in Figure \ref{fig: report_late}. Significant anomalies can be observed even without the presence of an incident report. Those data under unreported anomalies are otherwise not included in the incident detection/prediction models. Even worse, they are labeled as data under recurrent traffic conditions, which could mislead the training of AID models. 

In summary, overlooking the issues related to incident reports leads existing AID methods to fit their results to the reports themselves (which may be delayed, false, or missing) rather than to true and timely anomalies, particularly the early stages of an anomaly or unreported anomalies, which potentially are incidents, limiting conventional AID models' performance.  

Besides performance issues, another key challenge for conventional AID methods today is their scalability and generalizability. Most of the data used in AID methods, such as loop detector data, traffic camera data, and high-precision GPS or radar sensor data, are not widely available or can be very costly in real-time. For instance, real-time 24/7 loop detector data is limited to a very small number of segments or none in most regions, let alone high-precision radar sensor data. Social media data used by a few studies have also faced more scraping restrictions nowadays. On the other hand, data sources such as INRIX (\url{www.inrix.com}), HERE(\url{www.here.com}), and TomTom (\url{www.tomtom.com}) provide probe vehicle speed data, while Waze (\url{www.waze.com}) and State DOT offer incident reports. These data sources cover almost all major roads, providing the foundation for large-scale generalization of AID methods. The more ubiquitous data coverage of spaces and time enables a better understanding of network traffic flow and, thus, early anomaly detection on most road segments. It is essential to develop an incident detection/prediction model relying on those data so that we can afford to detect/predict incidents virtually anywhere and anytime. However, very few models are built exclusively on such data. 

To address the above problems, we will propose an early anomaly detection system that can learn and capture a wider range of anomalies, i.e., not only fits incident reports but also detects early incident features and unreported anomalies, especially those affecting broader areas. In addition, to ensure scalability and generalizability, the inputs to the new model should be limited to widely available low-cost data rather than data from extremely selected locations or manually selected data. Our contributions can be summarized as: 
\begin{itemize}
    \item We propose an anomaly detection model specifically designed for early-stage incident detection and prediction. Our model not only aligns well with true incident reports but also detects or predicts incidents before delayed reporting or in cases where incidents remain unreported. This model is fundamentally different from conventional AID models.
    \item  In contrast to most methods in the literature, the entire process is designed in an automated framework, which does not require manual selection of anomaly reports during training, thereby enhancing scalability and efficiency.
    \item Our model uses ubiquitously available and low-cost data with no additional detector required, increasing its scalability and generalizability. We conducted experiments on ten segments across two road networks, which can be generalized to any size of road segments in the network.
    \item Even with low-cost data, our model effectively balances the detection rate and false alarm rate, outperforming the baseline models.
\end{itemize}

The main structure of this paper is as follows: Section \ref{section: literature} reviews existing AID research and identifies the research gaps. Section \ref{section: data} introduces the data we used and the data processing methods. Section \ref{section: method} describes our methodology in detail, including how to generate a new set of labels, address data imbalance issues, and employ design and training techniques to achieve data mining and effective model training. Section \ref{section: results} demonstrates our framework's data mining capabilities by comparing models trained with anomaly labels, followed by a comparison with incident reports. Section \ref{section: conclusion} provides a summary and outlines future research directions.

\section{Literature Review}
\label{section: literature}
Research on Automatic Incident Detection (AID) began in the 1970s. Early traditional methods included statistical and competitive approaches. Statistical methods compare observed values with historical data, using statistical metrics such as standard normal derivative (SND) \citep{dudek1974incident} and inter-quartile distance (IQD) \citep{chakraborty2019data} to measure differences. An alarm is triggered when an outlier is observed. However, without direct supervision by incident labels, outliers identified may not be anomalies, resulting in high false alarm rates. Another statistical approach measures the difference between predicted values and real observed values. Alarms are triggered when predictions significantly differ from observations. This is based on the assumption that traffic patterns can be accurately predicted under recurring conditions while not in non-recurring conditions. The forecasting models used are generally time-series, such as autoregressive integrated moving average (ARIMA) \citep{ahmed1982application} and long short-term memory (LSTM) \citep{pan2022development}. This method performs well when observations are accurate, and the prediction model has low errors in recurring conditions. However, the same issue persists: these models lack supervised learning with anomaly labels, so outliers may not be anomalies. In fact, the observations are inherently noisy, and recurring conditions cannot ensure low prediction errors, especially during peak hours, resulting in poor detection performance. 

Comparative methods collect traffic data such as flow, speed, and occupancy through detectors and compare them with past incident patterns. Alarms are triggered when the collected data matches a pre-defined incident pattern. Representative comparative methods include the California \citep{payne1978freeway}, Manchester \citep{persaud1989catastrophe}, and Minnesota \citep{stephanedes1993application} algorithms. These methods are highly inspiring and have pioneered the analysis and extraction of incident characteristics. However, they largely remain at the level of manually summarizing incident characteristics, relying on pre-defined simple structures or patterns to determine whether to trigger an alert. The limited data mining capacity results in a high false alarm rate and mean time to detection. Recent work, such as \citet{10287856}, proposed comparing observed data with the monthly profile under similar traffic states. By computing the first-order derivative of the normalized Chebyshev distance, they designed a threshold-based method to capture significant temporal deviations from historical traffic patterns. Nevertheless, this method fundamentally depends on heuristic assumptions about incident characteristics.% Additionally, their reliance on specific sensors limits their generalizability.

Due to the richness and complexity of traffic data, researchers have begun to consider using artificial intelligence for data mining to detect incidents. \citet{cheu1995automated} used multi-layer feedforward neural networks (MLF) for incident detection, outperforming traditional methods on simulated data and small field data containing 9 incidents. To further validate feasibility on real-world data, \citet{dia1997development} used data collected from Tullamarine Highway, Melbourne, demonstrating their feasibility on real datasets. \citet{abdulhai1999enhancing} optimized the model, proposing a Bayesian-based probabilistic neural network (BPNN), and trained the model using data from I-880, California, and I-35W, Minnesota. \citet{jin2002development} further optimized the model by proposing the constructive probabilistic neural network (CPNN) and conducted studies on the same experimental section of I-880. Given the strong performance of support vector machines (SVM) on binary classification problems, \citet{yuan2003incident} proposed an SVM-based model and validated it on the datasets used by \citet{jin2002development} and \citet{abdulhai1999enhancing}. To address the lack of incident samples, \citet{lin2020automated} used generative adversarial networks (GANs) to generate more incident samples and use them combined with real incident samples to train an SVM classifier. Similarly, \citet{li2022real} used GANs on the same dataset as \citet{lin2020automated} but replaced the SVM classifier with a temporal and spatially stacked autoencoder (TSSAE). \citet{he2023autonomous} proposed a reinforcement learning-based approach that assigns rewards based on temporal differences and demonstrated its effectiveness on synthetic data. However, the lack of real-world incident reports limited their analysis to only two real-world cases.
% https://ieeexplore.ieee.org/stamp/stamp.jsp?arnumber=8569402 

All the above research has utilized inductive-loop detectors, which can provide comprehensive data on traffic flow, occupancy, and speed. However, the reliance on loop detectors reduces the generalizability of the methods. For example, in the U.S., most highway segments do not have loop detectors installed. Even when installed, the density of the detectors is often insufficient to meet effective detection requirements. For instance, the loop detectors used by \citet{yuan2003incident, lin2020automated} and \citet{li2022real} are spaced about every 0.5 miles, which limits the applicability of these methods to only the road segments near the installed detectors. Other studies using traffic cameras \citep{ki2007traffic, singh2018deep, chakraborty2018freeway} or high-precision GPS \citep{han2020traffic, sermons1996use} also face similar issues. Very few studies have leveraged easily accessible data. \citet{gu2016twitter} and \citet{zhang2018deep} utilized social media data for incident detection. However, social media is inherently a highly noisy form of manual reporting, facing issues such as irrelevance, delays, and missing, similar to traditional incident reports. In contrast, the speed data integrated from probe vehicles used by \citet{sethi1995arterial} provides an intuitive reflection of traffic conditions and is also easily accessible. Though \citet{balke1996using} pointed out that the limited penetration rate of probe vehicles leads to less prominent results, recent advancements in data collection methods have made the penetration rate of probe vehicle speed data increase. \citet{cheu2004sampling} and \citet{asakura2017incident} also explored the required penetration rate of probe vehicles to achieve efficient detection through simulation data. It should be acknowledged that probe vehicle speed data still faces high noise issues compared to detector data. As a result, related research remains quite limited. Recent progress using probe vehicle data in AID includes the method proposed by \citet{chakraborty2019data}, which combines bilateral filtering and total variation with statistical methods. However, there is still a lack of research utilizing other techniques, such as AI, which achieve better mining results on detector data than traditional statistical-based AID methods.

In addition to input data, another long-neglected issue is anomaly labels. Anomaly labels typically come from incident reports. However, incident reports often contain numerous manual errors \citep{10287856, he2023autonomous, 10287217}, which can negatively impact prediction performance and even lead to model convergence issues. Therefore, it has long been necessary to manually filter out real and impactful incidents. \citet{dia1997development}  manually selected 60 out of 385 incidents, \citet{abdulhai1999enhancing} manually filtered and organized incident reports on I-880 California and I-35 Minnesota, \citet{petty1996freeway} used driver manually calibrated incident data, and \citet{chakraborty2019data} combined manual selection with traffic cameras to filter incident reports. Subsequent researchers have widely used these datasets or manually filtered their datasets \citep{zhu2021dynamic, coursey2024ft}. Filtering and organizing incident reports requires significant manual effort, limiting the model's generalizability. To overcome this issue, \citet{10287217} introduced Temporal Positioning of Flow-Density Samples (TP-FDS) derived from studying the flow-density relationship. They utilized loop-detector data at a 15-minute resolution and validated their approach through comparison with expert-labeled data. In order to avoid reliance on loop detector data and thereby enhance the generalizability, \citet{10287856} proposed an automatic method that filters incident reports based on the first-order derivative of temporal speed differences. Although the approach relies solely on speed data, its primary objective is to improve incident duration prediction. As a result, it primarily targets temporal anomalies and may struggle to identify spatial anomalies. Moreover, it does not address key challenges in traffic anomaly detection, such as detecting early-stage disruptions that fall just below predefined thresholds or accounting for missing incident reports. To the best of our knowledge, current speed-based automatic labeling methods generally fall short of systematically addressing diverse anomaly types, such as spatiotemporal anomalies, early-stage disruptions, and unreported incidents. Moreover, most current research still aligns prediction results with selected incident reports rather than anomalies. It is still unclear how newly generated labels can contribute to incident detection, especially in terms of timeliness. We need a model capable of achieving broader anomaly detection.

There are more approaches based on synthetic data, which we will not detail here due to challenges in real-world data collection. For more comprehensive literature reviews on automatic incident detection, please refer to \citet{parkany2005complete} and \citet{elsahly2022systematic}.

\section{Data Preparation \& Feature Engineering}
\label{section: data}
We first discuss multiple data sources that are commonly available to use for early anomaly detection. Those data are most general in the sense they can be obtained to cover any geographic region and thus generally applicable to all locations. The data can be publicly available or offered by a number of data vendors with relatively mature technologies in the present market. In particular, we use incident reports, probe speed (segment level), and weather conditions for this study. Notably, all the information we use is gathered without the need to install additional detectors on the road network.

\subsection{Incident Reports}
Incident data can come from Waze and the State Departments of Transportation, which typically provide incident data feeds (real-time and historical). In our experiments, we use both Waze data and State DOT data, specifically from Pennsylvania DOT and Maryland DOT, depending on the region studied. We select incident types that typically cause non-recurrent changes in traffic. Due to system discrepancies, the selected incident types vary slightly among various systems. The selected incidents mainly include accidents, hazardous weather, special events, and crashes. Those incident reports come with geographic locations and report time for each incident, which is translated to a binary indicator, $INC^i(t)$, which denotes whether there is an incident report on link $i$ at time $t$. 

\subsection{Speed Data}
\label{section: speed}
\textit{\textbf{Probe Vehicle Speed}}: In our experiments, speed data comes from INRIX, which calculates speeds through real-time monitoring of probe vehicles. Similar data can also be obtained from other sources, e.g., HERE, AirSage, TomTom, or telematics data. We use four types of INRIX speed data: 1-minute granularity speed for private vehicles, trucks, and all vehicles, as well as 5-minute granularity speed for all vehicles. To describe the speed on a link, we primarily refer to the 5-minute granularity speed for all vehicles, as it exhibits a lower noise level and provides a comprehensive measurement across different vehicle types.
\begin{eqnarray}
\label{eq: 5-min}
v^{i}(t)  := v^{i, \text{5-min}}(t) 
\end{eqnarray}
where link $i$ is in the segment defined by TMC (Traffic Message Channels), which is unified across multiple data vendors,  $v^{i}(t)$ is the corresponding segment speed, and $v^{i, \text{5-min}}(t)$ is the 5-minute granularity speed for all vehicles of the segment. To impute missing 5-minute granularity speed data for all vehicles, we initially attempt to fill the gaps using the space mean speed from 1-minute granularity data for all vehicles, as described in Equation \ref{eq: SpaceMean}. If all the 1-minute data during that 5-minute interval are also missing, we infer that traffic flow is minimal so that no probe vehicles are present. In such cases, we use the free flow speed, defined as the 85th percentile speed, to fill in the missing data. This approach allows us to compile a complete dataset of 5-minute speed data for all vehicles. 
\begin{eqnarray}
\label{eq: SpaceMean}
v^{i, \text{5-min}}(t) = \frac{\sum_{j=0}^{4}\mathbb{I}(v^{i, \text{1-min}}(t-j\text{-min}) \neq \text{NaN})}{\sum_{j=0}^{4}\{\mathbb{I}(v^{i, \text{1-min}}(t-j\text{-min}) \neq \text{NaN})/v^{i, \text{1-min}}(t-j\text{-min})\}}
\end{eqnarray}
The 5-minute speed data for trucks and private vehicles are also calculated from 1-minute granularity using Equation \ref{eq: SpaceMean}. To impute missing speed data for trucks and private vehicles, we initially calculate the ratio of non-missing speed data for trucks and private vehicles to that of all vehicles. We then multiply these ratios by the complete speed dataset for all vehicles to fill the gaps. 

\textit{\textbf{Slowdown Speed}}: We further perform feature engineering on the speed data. The slowdown speed as defined in Equation \ref{eq: SD} is calculated,
 \begin{eqnarray}
\label{eq: SD}
SD^{i}(t) = \max(\frac{\sum_{j\in \tau^{i}}v^j(t)}{|\tau^{i}|}-v^{i}(t), 0)
\end{eqnarray}
where $\tau$ represents the set of upstream link segments within a certain distance range, and $|.|$ denotes the number of elements in the set. Equation (\ref{eq: SD}) means that for a target link segment $i$, its slowdown speed is the difference between a target link and the speed of all its upstream links. A large slowdown speed indicates the presence of a sudden back-of-sequence slowdown in the spatial domain \citep{doi:10.1177/0361198120917668} and has a high correlation with the occurrence of incidents according to the incident shock-wave theory \citep{wirasinghe1978determination}.

\textit{\textbf{Travel Time Index}}: We also calculate the travel time index (TTI) using the formula specified in Equation \ref{eq: TTI}, where $V^i$ denotes the speed distribution of segment $i$, and $\mathbb{P}_{0.85}(V^i)$ represents the 85th percentile speed. While the slowdown speed measures anomalies in the spatial dimension, the travel time index provides a comparison in the temporal dimension. Both the slowdown speed and travel time index calculations utilize the complete 5-minute granularity speed data for all vehicles.
\begin{eqnarray}
\label{eq: TTI}
TTI^{i}(t) = \max(\frac{\mathbb{P}_{0.85}(V^i)}{v^i(t)}, 1)
\end{eqnarray}
\textit{\textbf{Seasonal Recurrent Speed}}: The travel time index measures speed anomalies in the temporal dimension but lacks seasonal and day-of-week patterns that the speed clearly exhibits. Therefore, we defined a seasonal recurrent speed (SRS) using Equation \ref{eq: SRS} to provide the model with a reference for recurrent speed under the current season and day-of-week. Equation \ref{eq: SRS} represents the mean speed on the same day of the week over the past three weeks, excluding any periods with incident reports. $INC_i(t) = 1$ when there is an incident report on link segment $i$ at time $t$, and $0$ otherwise.
\begin{eqnarray}
\label{eq: SRS}
SRS^i(t) = \frac{\sum_{j=1}^{3}v^i(t-j*\text{week})\mathbb{I}(INC^i(t-j*\text{week})\neq 1)}{\sum_{j=1}^{3}\mathbb{I}(INC^i(t-j*\text{week})\neq 1)}
\end{eqnarray}

\textit{\textbf{Sampling rate}}: Fixed location detectors (such as loop detectors, cameras, etc.) are typically needed to obtain flow data, which are generally unavailable in most roadway segments. Thus, we approximate the flow data with the data density provided along with TMC speed data. TMC data density depends on the number of probe vehicles when calculating segment speeds and is divided into three levels: A, B, and C, representing high, medium, and low probe vehicle numbers, respectively. This is obtained from INRIX data, but other data vendors provide similar measurements in terms of sampling rates or confidence levels. We will normalize the numbers by setting the data density for A, B, and C to 1, 2/3, and 1/3, respectively, and setting it to 0 when the probe speed data is missing. % reference???? 

\subsection{Weather and Time Information}
\textit{\textbf{Weather}}: Weather data may be obtained from a few possible sources, e.g., open-meteo.com, NOAA NWS, metomatics.com, etc. We use seven features in a numerical format: temperature, humidity, hourly precipitation, hourly snowfall, hourly snow depth, hourly wind speed, and hourly wind direction.  

\textit{\textbf{Time}}: Our time features include month, week, day of the week, and time of the day. Since these features are all periodic, we use sine and cosine functions to encode this information. The encoding formulas are as follows:
\begin{eqnarray}
\label{eq: SIN&COS}
SIN(t) = \sin\left(\frac{2\pi t}{T}\right)\\
COS(t) = \cos\left(\frac{2\pi t}{T}\right)
\end{eqnarray}
where \( t \) is the time feature value, and \( T \) is the period of the feature (for example, the period for month is 12).

\subsection{Data for experiments}
\label{section: data_source}
As an experiment implemented in the later part of this paper, incident reports from two regions, Howard County, Maryland, and Cranberry Township, Pennsylvania, were collected for analysis. The data from Howard County spans from 2022-02-14 to 2023-02-12, while data from Cranberry Township covers 2022-01-01 to 2024-01-31. Figure \ref{fig:test} shows the Traffic Message Channels (TMC) network (red lines) and the recorded incident locations (blue dots) for two regions. We select five segments from each region for our study (ten in total). The segment locations along with the number of incident reports for each segment are shown in Figure \ref{fig:incident_count}. It is important to note that the data from Cranberry Township spans two years, whereas the data from Howard County covers only one year, leading to a higher number of incidents reported in Cranberry Township.
\begin{figure}[htbp]
\centering
\begin{subfigure}[b]{0.5\textwidth} 
    \centering
    \includegraphics[width=\textwidth]{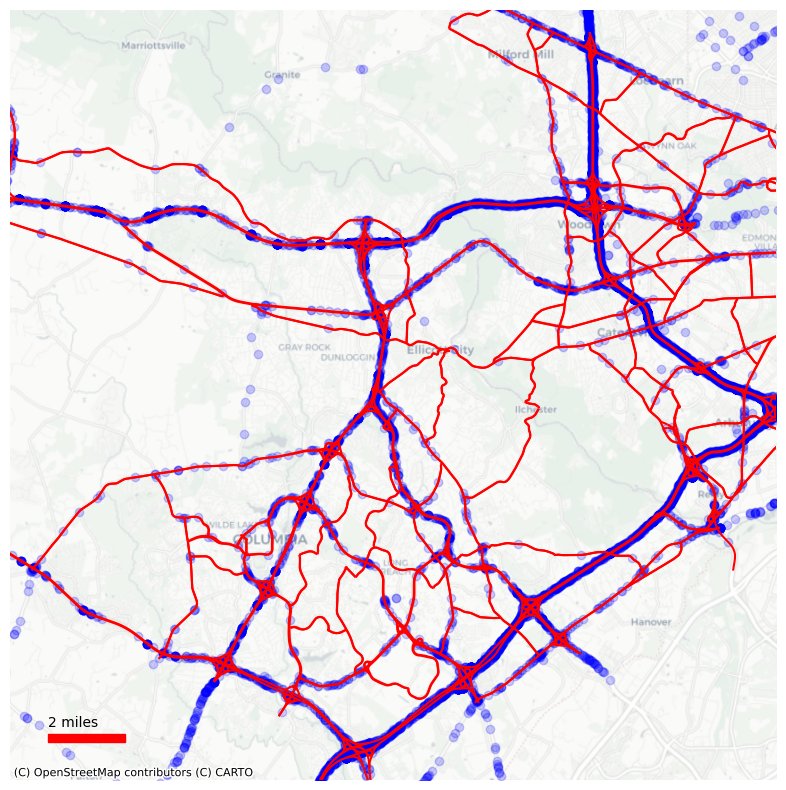} 
    \caption{Howard County, MD}
    \label{fig:sub1}
\end{subfigure}
\hfill 
\begin{subfigure}[b]{0.45\textwidth} 
    \centering
    \includegraphics[width=\textwidth]{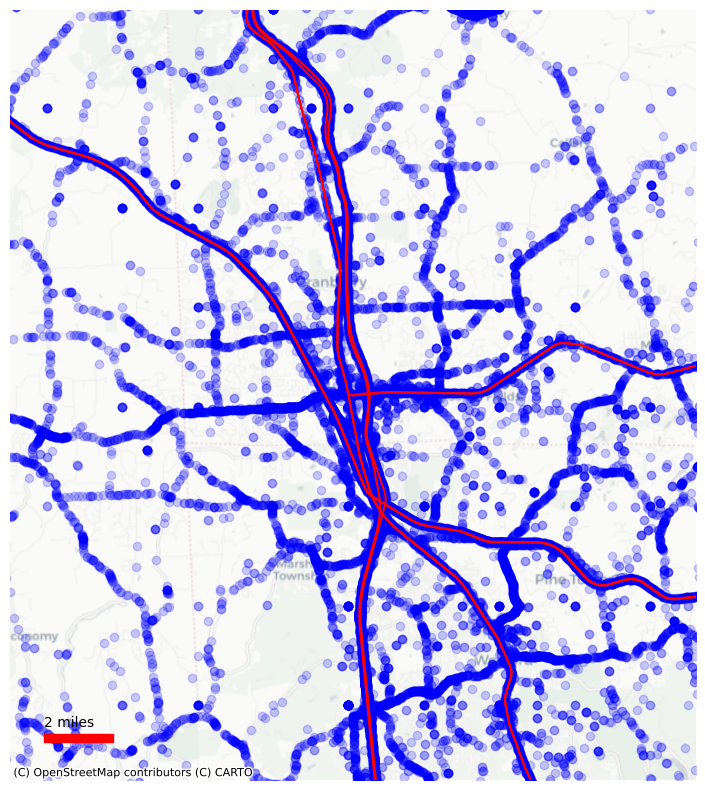} 
    \caption{Cranberry Township, PA}
    \label{fig:sub2}
\end{subfigure}
\caption{TMC Network and Incident Locations}
\label{fig:test}
\end{figure}
\begin{figure}[htbp]
\centering
\begin{subfigure}[b]{0.45\textwidth} 
    \centering
    \includegraphics[width=\textwidth]{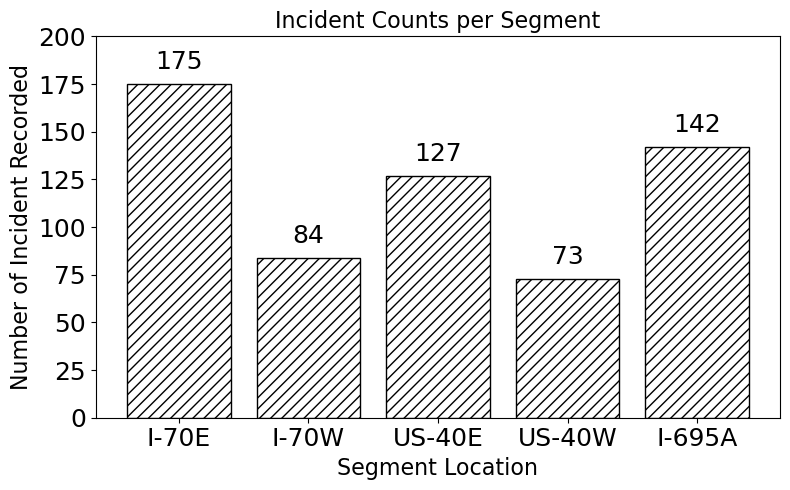} 
    \caption{Howard County, MD}
    \label{fig:sub3}
\end{subfigure}
\hfill 
\begin{subfigure}[b]{0.45\textwidth} 
    \centering
    \includegraphics[width=\textwidth]{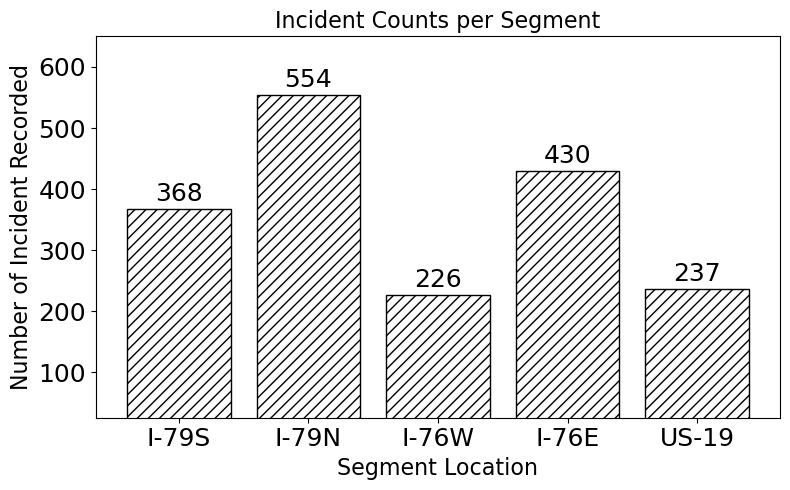} 
    \caption{Cranberry Township, PA}
    \label{fig:sub4}
\end{subfigure}
\caption{Incident Report Count Per Segment}
\label{fig:incident_count}
\end{figure}

The INRIX speed data used in this study includes all-vehicle speeds at both 1-minute and 5-minute granularities, along with personal vehicle and truck speeds at a 1-minute granularity. All data with 1-minute granularity are imputed to a 5-minute granularity using Equation~\ref{eq: SpaceMean}.
\section{Methodology}
\label{section: method}

Our goal is to train an effective model in early anomaly detection/prediction. The model's input consists of multi-source information over a past period, and the output is the anomaly status for a near future period. The incident reports are referenced for anomaly status labeling. To build an effective model, the following issues must be considered. 
\begin{itemize}
    \item \textbf{Rarity and randomness issues of the incident reports:} Regarding rarity, it leads to a severe imbalance between anomaly and normality data samples, which would hinder models from learning non-recurrent patterns. Regarding randomness, it may result in inconsistent data distributions between the training and validation/testing sets, posing challenges to model generalization.
    \item \textbf{Label quality issue:} Incident reports inherently have issues such as missing reports, delays, and false reports. Given that data points under anomalies are very scarce, directly using the presence of an incident report as an anomaly label can lead to substantial confusion between anomaly and non-anomaly data, making it difficult for the model to learn/train.
\end{itemize}
We propose a framework combining several strategies to address the above issues. The overall picture of the framework is depicted in Figure \ref{fig:overall}, where each gear pattern corresponds to a strategy we used. The framework first constructs a sub-graph for each link segment to establish the link-level anomaly detection/prediction model. Then, by referencing and analyzing incident reports, we generate our anomaly labels (i.e., ground truth) through prior knowledge and ahead-labeling. With the inputs and labels in place, the data split module generates training, validation (also including threshold-tuning set), and testing data, ensuring no data leakage while maximizing the number of training samples. The model training strategy includes the training techniques employed to deal with the anomaly rarity issue. In parallel with training, the alert threshold is calibrated to further enhance its performance. The details of each strategy will be introduced separately in this section.
\begin{figure}[!htb]
  \centering
  \includegraphics[width=1 \textwidth]{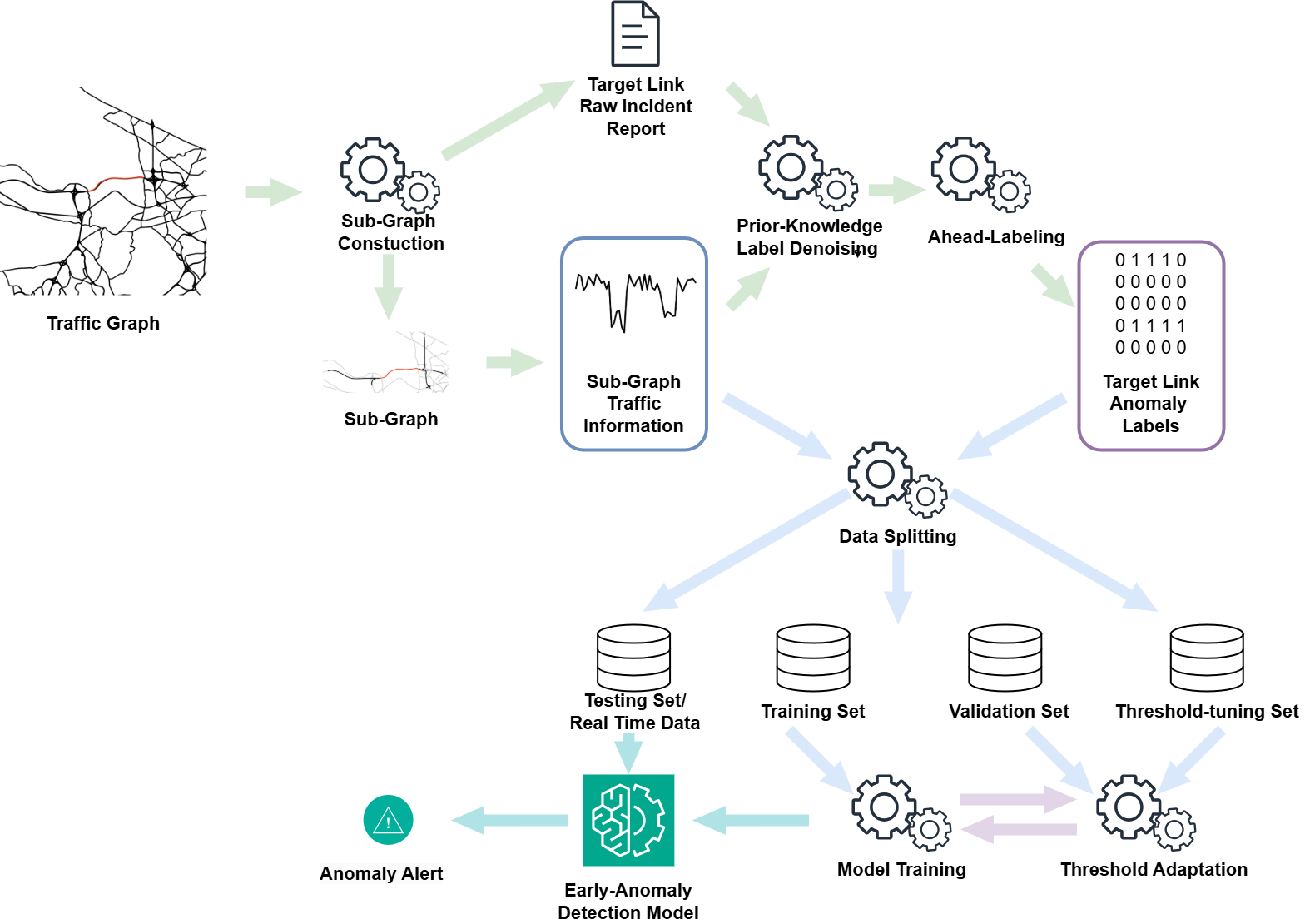}
  \caption{Framework Overview}
  \label{fig:overall}
\end{figure}
\subsection{Sub-Graph Construction}
\label{subsection: sub-graph}
Constructing a sub-graph for each link segment is our first strategy to address the anomaly rarity issue. Specifically, we build a separate machine learning model for each segment to predict its anomaly status instead of using a single machine learning model to predict all link segments' anomaly status. Though aiming for large-scale generalization, it is difficult to train a model to capture spatially or temporally variant incident probabilities and characteristics across different segments when anomaly data are scarce. Building separate models for each segment can reduce the complexity of the detection task. 

Besides reducing the complexity of the task, the model complexity is also reduced by controlling the input dimensions, which has been theoretically proven to reduce the demand for extensive training samples \citep{hastie2009elements, vapnik2013nature}. The model's input is limited to the traffic conditions within a certain range of upstream and downstream segments (which we refer to as a "sub-graph"). This is based on the observation that an incident typically affects the traffic conditions of its upstream and downstream segments \citep{karim2002incident}, and this effect generally diminishes with distance. By inputting the sub-graph information, the input-output relevance and model complexity can be balanced. 
\subsection{Label Denoising by Prior Knowledge}
\label{subsection: LabelDenoising}
Using incident reports directly as anomaly labels can introduce significant noise and hinder the convergence process. This is due to the inherent issues within incident reports, such as delays, missing, and inaccuracies. Additionally, not all incidents have a tangible impact on traffic conditions. For example, an incident on a multi-lane highway during off-peak hours might not affect speed or flow, thus resulting in an undetectable incident. This may not be of interest to traffic operators if there is no impact on mobility or safety for the time being. Consequently, using incident reports directly as labels can cause substantial confusion between anomaly and normality labels, diminishing their significance and learnability. To ensure the labels are meaningful and learnable, existing research typically uses a manual process to filter reported incidents; however, this introduces subjectivity and limits their scalability. Most importantly, manually selecting incident reports does not address the issue of missing reports, potentially causing unreported anomalies to be labeled as normal. Therefore, we propose a prior knowledge-based method for labeling anomalies. Besides filtering out insignificant or false reports, significant anomalies, which are potential unreported incidents, are also evaluated and labeled.

Given that traffic flow data often requires additional detectors and is not widely available, we use the slowdown speed described in Equation (\ref{eq: SD}) as the prior knowledge to infer whether an incident report corresponds to a traffic anomaly. Theoretically, as explained in Section \ref{section: speed}, a significant traffic incident of our interest should cause a significant slowdown speed. Table \ref{table:sd_table} empirically demonstrates the strong correlation between significant slowdown speeds and the occurrence of incidents. For instance, in the first row of Table \ref{table:sd_table}, data from Howard County reveals that only 1.27\% of timestamps on the I-70 westbound study segment have incident reports. However, of those timestamps associated with the top 3\% of highest slowdown speeds, 9.99\% have incident reports, and over 60\% of the entire incident reports fall within these top 3\% slowdown speed timestamps. Other study segments in Table \ref{table:sd_table} also exhibit similar characteristics, revealing a high correlation between significant slowdown speeds and the occurrence of incidents. 
\input{tables/sd_table}
\input{alg1}

Our proposed prior knowledge-based denoising method is outlined in Algorithm \ref{alg:1}. This method involves tuning a slowdown speed threshold for each link. If any point within the time frame of an incident report exceeds this threshold, the entire period is marked as an anomaly (Step 1). Additionally, any period during which the slowdown speed consistently exceeds this threshold is also marked as an anomaly (Step 2). Furthermore, we control the ratio of removal and addition anomaly labels by setting $\theta_1$ and $\theta_2$ (Step 3). It is important to note that the denoising process does not generate a new set of labels solely based on abnormal slowdown speed. The model is not trained to detect abnormal slowdown speeds. According to our rules, if an abnormal slowdown speed is detected at any timestamp within an incident report, all timestamps corresponding to that report will be marked as anomalies. This means not all points marked as anomalies necessarily exhibit abnormal slowdown speeds. Conversely, not every timestamp with abnormal slowdown speed will be labeled as an anomaly. The criteria for adding missing reports require that an abnormal slowdown speed must persist for at least $\theta_t$ steps to be labeled as an anomaly. In summary, abnormal slowdown speed is neither a sufficient nor a necessary condition for anomaly labeling. The default thresholds we set for $\theta_1$, $\theta_2$, and $\theta_t$ are 0.6, 1.0, and 3 (15 min as the prediction interval is 5 min), respectively. These values may be subject to slight adjustments based on the data collected.

\subsection{Ahead Labeling}
The algorithm \ref{alg:1} mainly extracts common features from incident reports. However, as shown in Figure \ref{fig: report_late}, incident reports often lack the early stage of incidents. While algorithm \ref{alg:1} can label some missing or delayed reports, they do not address the inherent delay issue in the reporting process. In addition, setting a fixed slowdown speed threshold may result in early-stage incidents—those that are developing but have not yet crossed the threshold—not being labeled. Therefore, we label anomalies starting from a few time steps prior to the reported incident/anomaly time to supplement the lack of early characteristics in the anomaly samples. For example, an anomaly from 7:30 a.m. to 8:30 a.m. can be ahead-labeled by 15 minutes to become from 7:15 a.m. to 8:30 a.m. The algorithm is shown in Algorithm \ref{alg:2}. The setting of $\theta_{ahead}$ balances the inclusion of early features and false anomalies. The larger the $\theta_{ahead}$, the more early incident features are included, but it also increases the likelihood of normal samples being labeled as anomalies. Practically, this would need to be tuned for each road segment or to the traffic operator's preference. In our experiment, we use $3$ as the default value of $\theta_{ahead}$, which is 15 minutes given the prediction interval is 15 minutes. 

\input{alg2}
\subsection{Multi-step Prediction \& Data Splitting}
\label{sec: multistep}
Conventional AID typically formulates its task as a single-step prediction, i.e., determining whether the road segment is currently in incident status. To promote early detection, in our model, we formulate an anomaly detection/prediction task as a multi-step prediction, predicting the anomaly status in the next few steps. In this case, if the model captures the early incident features, even if it may not cause anomaly in this step but a few steps later, the model will still be able to give an alert.  

Multi-step anomaly prediction/detection on past traffic conditions can be formulated as a sequence-to-sequence prediction task. To mitigate the issue of limited samples, we use a sliding window to increase sample numbers by partitioning the dataset into multiple subsections. As illustrated in Figure \ref{fig:data_spiliting}, suppose our study period is from 6 AM to 9 PM, and we use past 1 hour's data to predict the anomaly status for the next half hour, with predictions made every five minutes. Then, the first sample of the day uses data collected between 6:00 AM-6:55 AM to predict the anomaly status between 7:00 AM-7:25 AM; the second sample uses data collected between 6:05 AM-7:00 AM to predict the anomaly status between 7:05 AM-7:30 AM, and so on. Using a sliding window can increase the number of training samples to mitigate the limited data issues.
\begin{figure}[H]
  \centering
  \includegraphics[width=0.6 \textwidth]{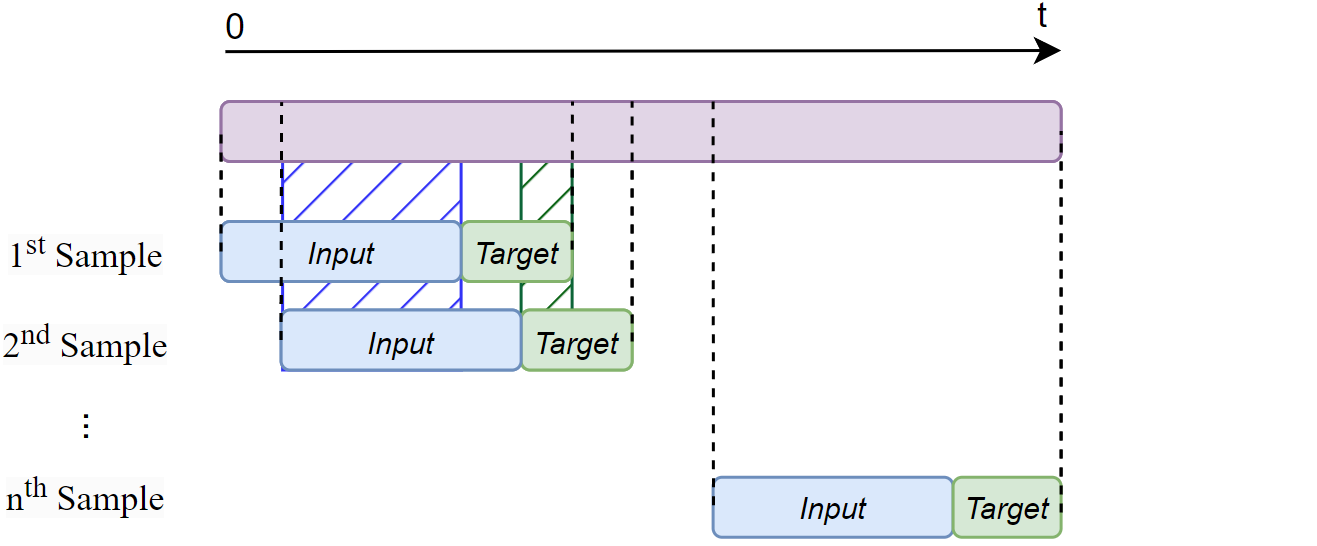}
  \caption{Sliding Window and Train-Test Contamination}
  \label{fig:data_spiliting}
\end{figure}

However, it is crucial to note that applying the sliding window must strictly occur after splitting the dataset. That is, we need to first divide the data into training, validation, and test sets and then apply the sliding window within each dataset. If the sliding window is applied to the entire dataset first and then randomly shuffled to create the training and test sets, it will result in data leakage, also known as train-test contamination. This occurs because information from the test set can inadvertently be mixed into the training set. As shown in Figure \ref{fig:data_spiliting}, the 1$^\text{st}$ and the 2$^\text{nd}$ samples, although not identical, have a high degree of overlap in both inputs and targets (see the hatched part). If they are assigned to the training and test sets respectively, even if the model overfits the 1$^\text{st}$ sample, testing using the 2$^\text{nd}$ sample may still produce good results due to its high similarity to the 1$^\text{st}$ sample, giving a false evaluation of model performance. We further verified by experiment that a model trained with train-test contamination performs poorly on a new dataset without data contamination.
\subsection{Model Training}
\label{sec: train}
This framework is applicable to various classical deep-learning models based on encoder-decoder structures, including Seq2Seq \citep{sutskever2014sequence}, Seq2Seq models with attention mechanisms \citep{bahdanau2014neural}, Transformer \citep{vaswani2017attention}, and GraphTrans \citep{wu2021representing}. As mentioned in Section \ref{subsection: sub-graph}, the traffic status of the sub-graph over the past period is fed into the encoder, and then the decoder outputs the anomaly status of the target link for the future period in an auto-regressive manner.

Weighted binary cross-entropy (WBCE), as shown in Eq.(\ref{eq: WBCE}), is used as the loss function for model training, where $w_{ano}$ denotes the additional weight for anomaly samples, $\hat{y}$ denotes the predicted value, and $y$ denotes the label value ($1$ for anomaly sample and $0$ for normality sample).
\begin{eqnarray}
\label{eq: WBCE}
L = -\frac{1}{N} \sum_{i=1}^{N} \left( w_{ano} y_i \log(\hat{y}_i) + (1 - y_i) \log(1 - \hat{y}_i) \right)
\end{eqnarray}

When $w_{ano}=1$, Eq.(\ref{eq: WBCE}) is the binary cross-entropy (BCE), which is commonly used for binary classification training due to its differentiability. The purpose of weighting is to overcome the imbalance in datasets caused by the scarcity of anomaly samples. Without weighting, the predominance of normality samples would bias the model to consistently output zero. For the setting of $w_{ano}$, the default practice is to divide the number of samples whose labels are 0 (normality samples) by the number of samples whose labels are 1 (anomaly samples). Further adjustment is a trade-off between recall and precision (definitions refer to Figure \ref{fig:TP}). Increasing the weight helps to improve recall, while decreasing it helps improve precision.
\begin{figure}[!htb]
  \centering
  \includegraphics[width=0.8 \textwidth]{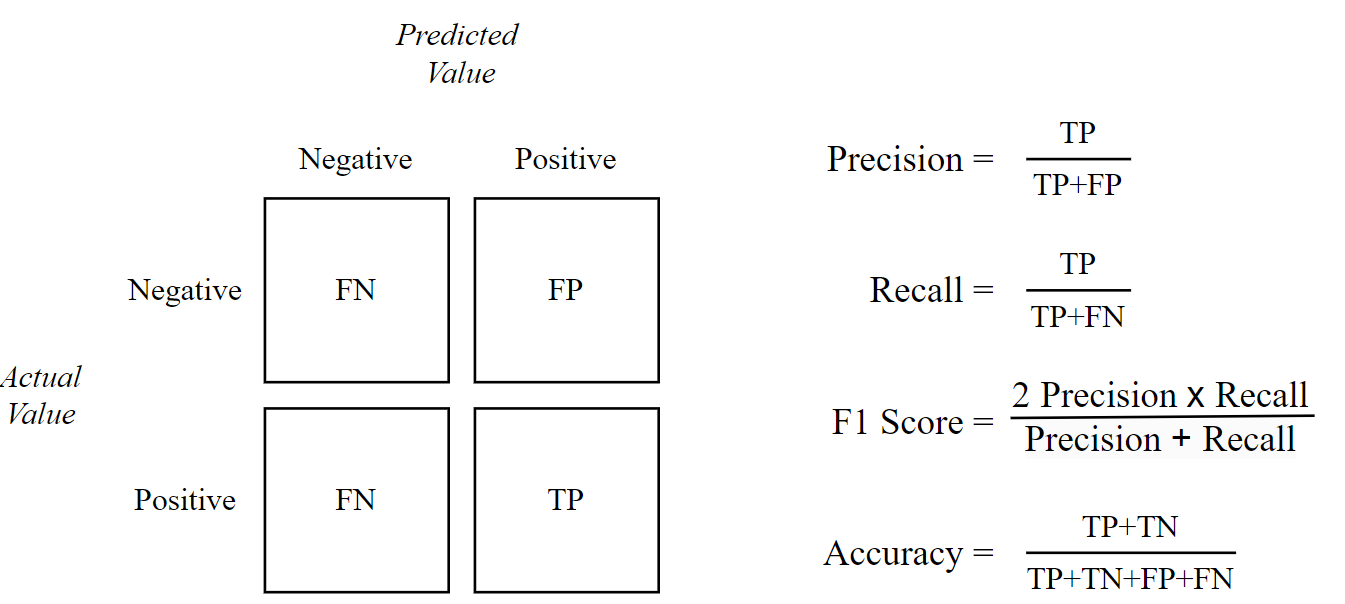}
  \caption{Recall, Precision, and F1 score}
  \label{fig:TP}
\end{figure}

During model training, we employ the teacher forcing technique \citep{williams1989learning} in the auto-regressive decoder to accelerate and enhance convergence. Teacher forcing is a widely adopted strategy in contemporary natural language processing (NLP) model training. This method involves substituting the decoder's prior outputs with actual labels during subsequent computations in the training phase. 

\subsection{Threshold Adaptation}
\label{section: threshold}
%The trained model outputs a number between 0 and 1, which can be interpreted as a probability of an anomaly; however, determining when to trigger an alert still requires a threshold. The practice of using 0.5 as a threshold in a balanced dataset assumes that the model can perfectly output values close to 0 and 1. However, due to the setting of $w_{ano}$ and the model's inability to converge perfectly, the 0.5 threshold lacks theoretical basis and may perform poorly. Therefore, it is necessary to tune an appropriate threshold for best model selection and development purposes.

%To prevent overfitting on the training data, threshold-tuning and model selection is performed on a validation set. Considering that incidents occur randomly, the distribution of anomaly samples in the validation set may differ from that in the training set. Therefore, we use the F1 score for model selection. The validation set is divided into two parts. The validation set 1 is used to find the threshold that maximizes the F1 score in validation set 1. With the tuned threshold, we can select the best model by optimizing the F1 score in validation set 2. Validation set 2 serves to simulate the test set. In this way, the model is selected based on a threshold that adapts to the model output, avoiding the issue of discarding well-performing models due to a fixed threshold. 

The model is optimized on the training set via a differentiable loss function, producing continuous output values ranging from 0 to 1. However, these continuous outputs themselves lack clear practical interpretation and can vary significantly across different segments, making them difficult to directly apply in real-world traffic management. To better meet practical needs, we introduce a threshold to determine when to trigger an alert. In other words, we need a threshold to convert the continuous outputs into meaningful binary classifications (0 or 1).

The practice of using 0.5 as a threshold might lead to suboptimal performance, particularly in imbalanced datasets. Additionally, traffic incidents exhibit seasonal and random characteristics—that is, their occurrence may be unevenly distributed over time. Therefore, we introduce a separate threshold tuning set. This dedicated dataset enables the optimization of threshold selection, ensuring that both the model and the chosen threshold generalize effectively across diverse incident distributions, thus improving model robustness. Our method for tuning thresholds and selecting models \input{alg3}

Similar to the setting of $w_{ano}$, depending on emphasis on precision and recall, other F-scores can be used. Here, $\beta$ is the hyper parameter, meaning that recall is $\beta$ time more important than precision. 
\begin{eqnarray}
\text{F}_{\beta} = (1 + \beta^2) \cdot \frac{\text{Precision} \times \text{Recall}}{(\beta^2 \cdot \text{Precision}) + \text{Recall}}
\end{eqnarray}

\section{Results and Discussion}
\label{section: results}
This section is divided into three parts: the results of label denoising, the results of early anomaly detection (with possible prediction), and incident report comparison. In the label denoising part, we compare our labeled anomalies with the incident report by analyzing speed deviation. In the early anomaly detection part, we use the generated anomaly label (ground truth) as a reference to assess different model's ability to detect or predict significant anomalies. In the incident report comparison part, we compare our prediction results, as well as baseline prediction results, to raw incident reports (with inaccurate, insignificant, missing, or delayed reports). 
\subsection{Label Denoising Results}
\label{section: label_results}
Given the extensive scale of the dataset, which spans over the years, expert-based manual labeling is infeasible. Therefore, we provide case studies to illustrate the effect of label denoising. As shown in Figure \ref{fig: noreport}, on the training set of the I-70W segment, we demonstrate how our algorithm \ref{alg:1} labels additional anomalies in time periods where no incidents were reported. In figure \ref{fig: noreport}, the black hatched areas represent the anomalies labeled by Algorithm \ref{alg:1}, while the blue hatched areas indicate those labeled by Algorithm \ref{alg:2} (i.e., ahead-labeling). The blue line represents the traffic speed of the segment on the target day, and the yellow and green lines correspond to the recurrent speed under similar conditions (i.e., the nearby same day of the week and time of the day). The comparison shows that our algorithms can label abnormal traffic conditions that significantly deviate from the recurrent reference, even without incident reports. It is worth noting that although our algorithm is primarily based on spatial features—specifically, the detection of speed anomalies via the slowdown speed defined in Equation (\ref{eq: SD}) —the labeled anomalies also exhibit significant deviation in the temporal dimension compared to the recurrent speed reference. The results show that the labeled anomalies are both temporal and spatial. Specifically, in cases (a), (b) and (d) of Figure \ref{fig: noreport}, we did not assign labels during the final phase of the deviation, which may be due to the slowdown speed already falling below the threshold. However, it is acceptable since this study does not focus on predicting the duration of incidents. Meanwhile, this also helps us predict the end time earlier. Additionally, in case (c), although there appears to be no deviation along the temporal dimension, our algorithm still assigns labels, as the spatial impact of the anomaly has not yet dissipated.

In addition, the results highlighted in the blue hatched area further demonstrate the necessity of incorporating ahead-labeling. Since Algorithm \ref{alg:1} derives the abnormal slowdown speed threshold based on incident reports, it inevitably inherits a key limitation of such reports—the lack of early-stage abnormal samples. If we rely solely on the labels generated by Algorithm \ref{alg:1}, many early abnormal patterns will be missing from the dataset. Consequently, models trained on this data may fail to effectively recognize the early stages of anomalies.
\begin{figure}[!htb]
    \centering
    \begin{minipage}{0.6\textwidth}
        \centering
        \includegraphics[width=\textwidth]{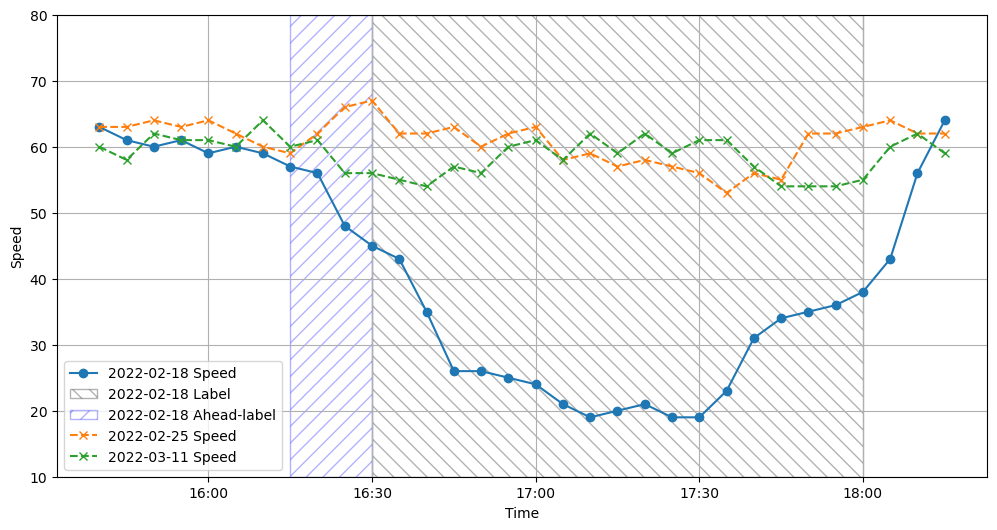}
        \subcaption[]{2022-02-18 Case}
    \end{minipage}
    \hfill
    \begin{minipage}{0.6\textwidth}
        \centering
        \includegraphics[width=\textwidth]{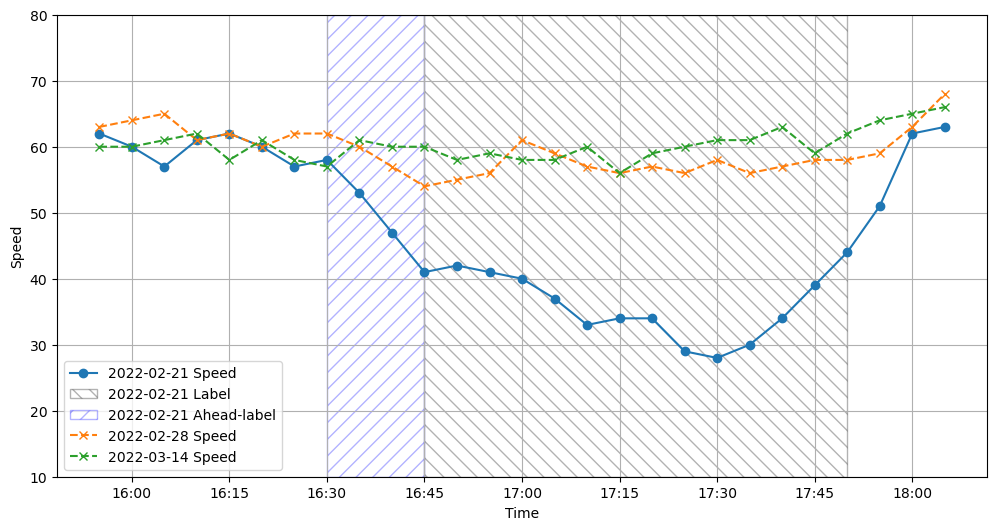}
        \subcaption[]{2022-02-21 Case}
    \end{minipage}
    \hfill
    \begin{minipage}{0.6\textwidth}
        \centering
        \includegraphics[width=\textwidth]{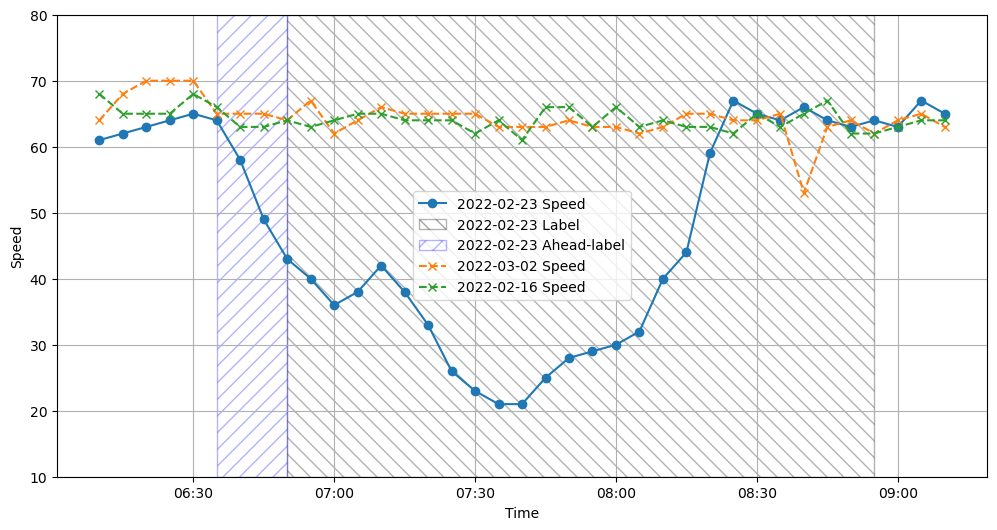}
        \subcaption[]{2022-02-23 Case}
    \end{minipage}
        \hfill
    \begin{minipage}{0.6\textwidth}
        \centering
        \includegraphics[width=\textwidth]{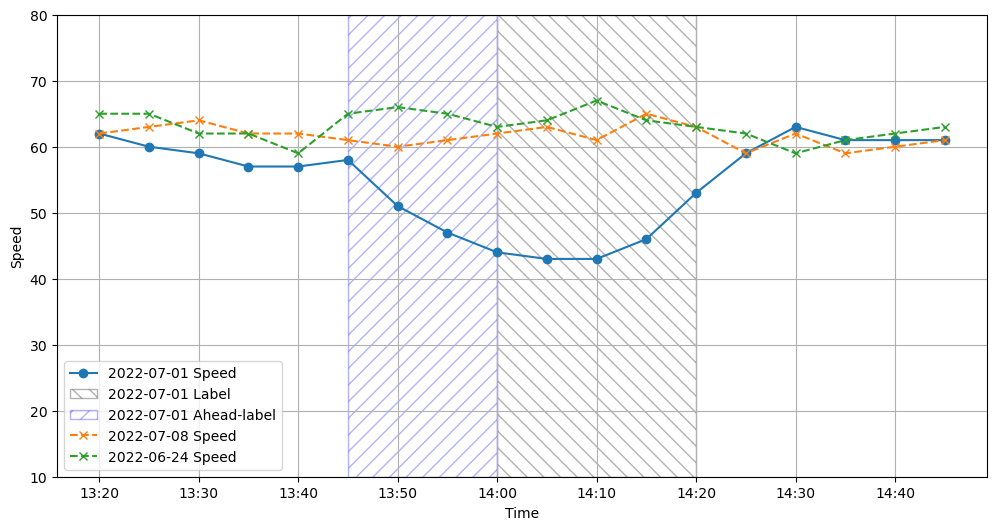}
        \subcaption[]{2022-07-01 Case}
    \end{minipage}
    \caption{No Report, Anomaly Labeled}
    \label{fig: noreport}
\end{figure}

\begin{figure}[!htb]
    \centering
    \begin{minipage}{0.6\textwidth}
        \centering
        \includegraphics[width=\textwidth]{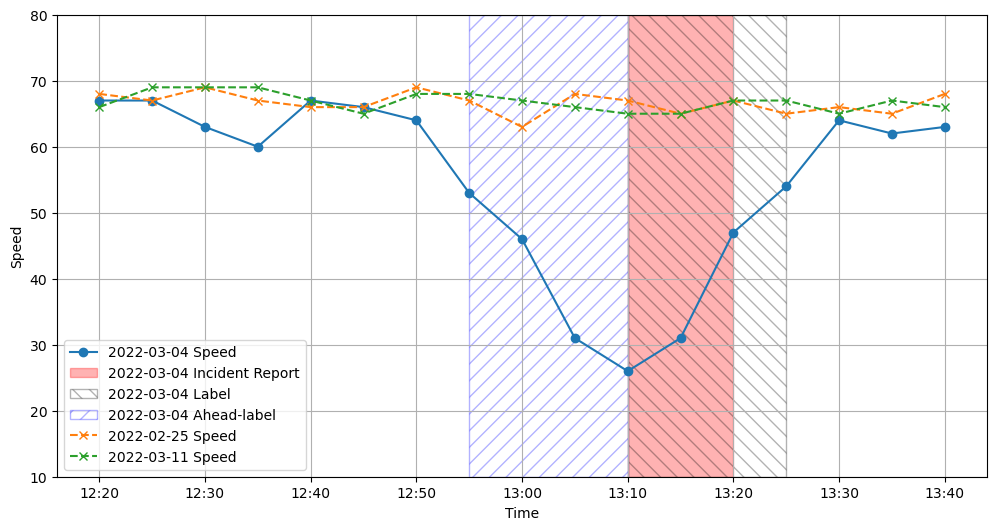}
        \subcaption[]{2022-03-04 Case}
    \end{minipage}
    \hfill
    \begin{minipage}{0.6\textwidth}
        \centering
        \includegraphics[width=\textwidth]{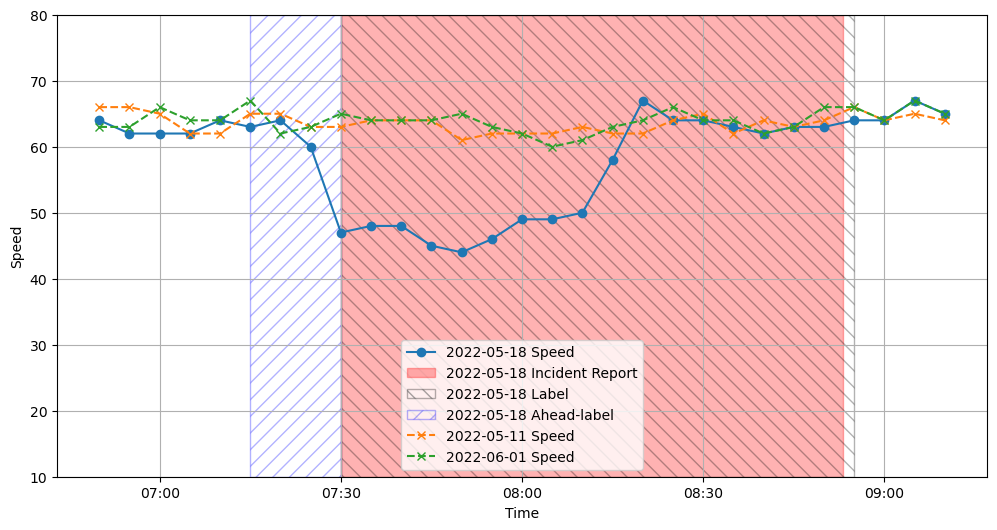}
        \subcaption[]{2022-05-18 Case}
    \end{minipage}
    \hfill
    \begin{minipage}{0.6\textwidth}
        \centering
        \includegraphics[width=\textwidth]{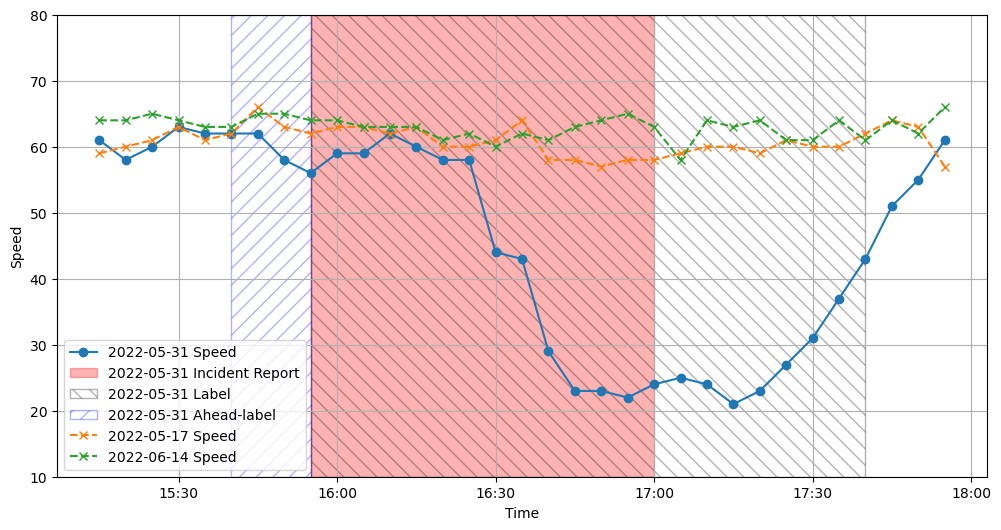}
        \subcaption[]{2022-05-31 Case}
    \end{minipage}
        \hfill
    \begin{minipage}{0.6\textwidth}
        \centering
        \includegraphics[width=\textwidth]{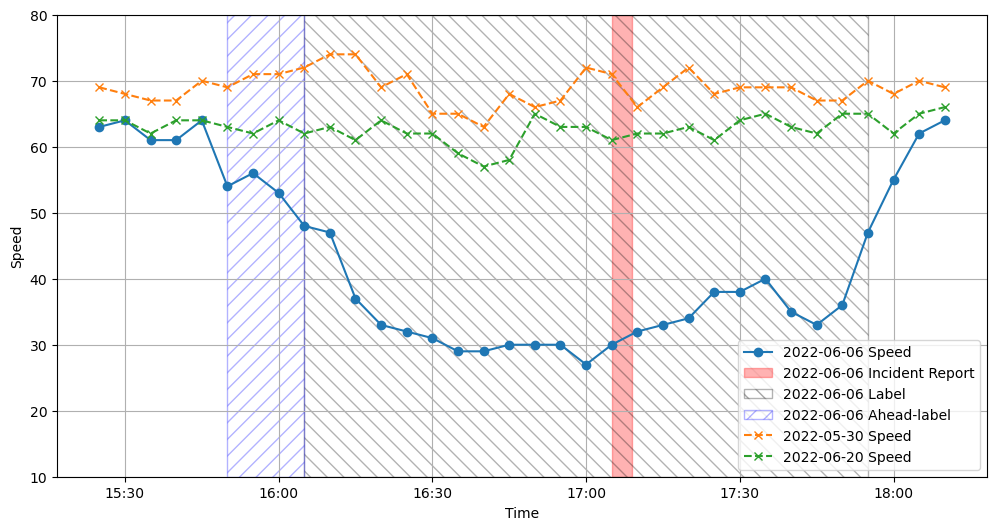}
        \subcaption[]{2022-06-06 Case}
    \end{minipage}
    \caption{Reported, Anomaly Labeled}
    \label{fig: badreport}
\end{figure}
Similarly, Figure \ref{fig: badreport} presents the anomalies labeled by our algorithm for time periods with available incident reports. The red-shaded regions correspond to the durations of the reported incidents. It can be clearly observed that the anomalies labeled by our method are more temporally consistent with the deviation of speed.

Despite the absence of expert-labeled ground truth, which makes it difficult to directly quantify the overall performance of our label-denoising and ahead-labeling strategies, we observe that models trained with our generated labels consistently converge and achieve superior performance in real-time prediction tasks. In contrast, models trained on original incident reports often fail to converge. The learnability of the labels indirectly highlights the rationality of our labeling approach. A more detailed analysis is provided in Section~\ref{section: results3}.
\subsection{Early Anomaly Detection/Prediction}
\label{section: annomlay_overall}
In this section, we discuss the predictive performance of different models on our newly generated labels. Our experiments were conducted on the two road networks introduced in Section 3. We divided our training, testing, and validation sets in a 7:2:1 ratio in chronological order as explained in section \ref{sec: multistep}. We chose the periods that are relatively more challenging to predict: Predictions were made on weekdays, from 6:00 a.m. to 8:30 p.m., forecasting the anomaly status for the next half hour. As discussed in Section~\ref{sec: multistep}, the problem can be formulated as a sequence-to-sequence (Seq2Seq) prediction task, for which the Transformer and its variants represent the current state-of-the-art. However, due to the inherent rarity and randomness of anomaly occurrences, effectively applying the Transformer model in our scenario necessitates further optimization and adaptation, as detailed in Sections~\ref{sec: train} and~\ref{section: threshold}.

We train the GraphTrans \citep{wu2021representing} using the introduced strategy and show the relationship between recall, precision, F1 score, and accuracy of the model as the threshold changes. Taking the Howard County I-70 segment as an example, Figure \ref{fig: recall_pre_without_ahead} shows the recall, precision, and F1 score for predicting anomaly status 5 minutes, 15 minutes, and 25 minutes ahead. The left side shows the results of the validation set, and the right side shows the results of the test set. From the figures, it can observed that the model's performance on the validation set is almost identical to that on the test set. As the prediction period increases, the best F1 score gradually decreases. As the threshold increases, recall increases while precision decreases. Selecting an appropriate threshold allows us to achieve high accuracy and a high F1 score. The range for this threshold is approximately between 0.5 and 0.7. Figure \ref{fig: recall_pre_with_ahead} shows the results trained with 15-minute ahead labeling. Similarly, as the prediction period increases, the best F1 score decreases. As the threshold increases, recall improves while precision drops. Selecting a threshold between 0.4 and 0.6 can balance recall and precision, ensuring high accuracy. It is worth noting that the ground truth differs between the cases with and without ahead labeling, so directly comparing these evaluation metrics between Figure \ref{fig: recall_pre_without_ahead} and Figure \ref{fig: recall_pre_with_ahead} lacks significance.
\input{images/recall_precision_F1}
\input{images/recall_precision_F1_15}

To further compare additional models, we conducted experiments on 10 highway segments across two road networks. Tables \ref{tab: tsmo_a} and \ref{tab: Cranbeery_a} show the results. To avoid the interference of $\theta_{ahead}$ settings affecting the quality of the labels and thus reducing the persuasiveness of the results, Tables \ref{tab: tsmo_a} and \ref{tab: Cranbeery_a} use results without the ahead label. The effect of the ahead label will be discussed in the next chapter. In addition to Transformer \citep{vaswani2017attention} and GraphTrans, we also evaluated models based on supervised learning algorithms that have demonstrated outstanding performance in AID tasks so far, including Random Forest, Light Gradient Boosting Machine, Support Vector Machine (SVM), and an SVM enhanced with data augmentation using a Generative Adversarial Network (GAN). All models were implemented with consideration for class imbalance and have been well-tuned and trained. The abbreviations represent the following: 
\begin{itemize}
    \item \textbf{RF}, Random Forest. Tree-based models are considered to perform well on imbalanced data and time-series predictions. We trained six RF models for prediction models ranging from 5 minutes to 30 minutes.
    \item \textbf{SVM}, Support Vector Machine, trained using the Radial Basis Function (RBF) kernel, demonstrates superior performance in conventional AID \citep{yuan2003incident}. Due to the impracticality of the computation time required for upsampling, we used downsampling for training. We trained six SVM models for prediction models ranging from 5 minutes to 30 minutes, in 5-min time increments. 
    \item \textbf{GAN}, Generative Adversarial Net \citep{goodfellow2014generative}. Inspired by the idea proposed by \citet{lin2020automated}, We first use GAN to learn the characteristics of anomaly samples and generate more anomaly samples to overcome the sample rareness issue. SVM is then used for classification. Similar to SVM, we need to build six models.
    \item \textbf{LightGBM}, Light Gradient Boosting Machine \citep{ke2017lightgbm}. LightGBM is an efficient gradient boosting framework that uses tree-based learning algorithms. It is designed to be distributed, efficient, and capable of handling large-scale data compared to the XGBoost, which has shown superiority in anomaly detection \citep{parsa2020toward}. For prediction models ranging from 5 minutes to 30 minutes, we trained six LightGBM models.
    \item \textbf{Trans}: Transformer \citep{vaswani2017attention} , trained with the optimization strategies described in Sections~\ref{sec: train} and~\ref{section: threshold}.
    \item \textbf{G-Trans}: GraphTrans \citep{wu2021representing} , trained with the optimization strategies described in Sections~\ref{sec: train} and~\ref{section: threshold}.
\end{itemize}
From the two tables, it is evident that the choice of model significantly impacts prediction performance. Specifically, SVM demonstrates poor precision performance across all road segments, which means that it generates a substantial number of false positives. After incorporating GAN-based data augmentation, precision significantly improves despite a slight decrease in recall. It is noteworthy that, on our anomaly detection task, the observed decrease in recall following GAN augmentation differs from \citet{lin2020automated}'s reported increase in detection rate. The observed performance gains from GAN-based augmentation in our anomaly detection task may be primarily attributed to the more balanced inclusion of both anomaly and normal samples during training. This balance potentially allows the model to capture the distribution of normal traffic conditions better. As a complementary analysis, training the SVM with a limited degree of upsampling—without introducing GAN-generated samples—could help clarify whether the improvements are mainly due to the increased representation of normal samples rather than the generative nature of GAN augmentation itself. Nevertheless, the positive impact of GAN augmentation on the overall model performance (F1 score) remains confirmed.

For tree-based models such as RF and LightGBM, LightGBM generally demonstrates a better balance between precision and recall while maintaining a high F1 score. In contrast, RF exhibits relatively poor recall performance, especially in regions such as I-695A, Howard County, and I-76W Cranberry Region. In these areas, the extremely low recall of RF leads to significantly worse overall performance compared to LightGBM and even SVM. This observation suggests that RF tends to make more conservative predictions, which aligns with the nature of bagging-based models—relying on majority voting, making the model more cautious in predicting positive cases. In contrast, LightGBM achieves superior overall performance due to its boosting-based architecture, which allows the model to iteratively correct misclassifications, especially for borderline samples, thereby improving its classification capability. By further comparing LightGBM with GAN, we found that LightGBM significantly outperformed GAN on the four road segments in Howard County. On other road segments, GAN performed slightly better, but the differences were not substantial.

Overall, Trans and G-Trans consistently outperformed the other models, primarily due to the superior data representation and learning capacity of their architectures. In particular, on the I-76W, I-76E, and US-19, Cranberry region, the Transformer model demonstrated a clear advantage in short-term prediction tasks. We attribute this performance to its effective temporal encoding mechanism. To further validate this hypothesis, we tested a variant of the Transformer model on the I-76E segment with the positional encoding removed. The results showed that the F1 scores dropped significantly to 0.41 and 0.29 for 5-minute and 10-minute predictions, respectively—substantially lower than those of the original model. This confirms that capturing and modeling temporal dependencies from the input data is essential. It also supports the necessity of our proposed strategy of incorporating temporal models into the prediction/detection framework. Compared with Trans, G-Trans performs slightly better on most segments, likely due to its ability to capture topological structures. Notably, in the results for I-79 and I-76 in Cranberry, Trans and G-Trans are outperformed by GAN in predicting anomalies at 20, 25, and 30 minutes. From the perspective of model architecture, this is probably because we use a shared decoder for all six predictions, whereas GAN trains six separate models. It can be observed that our model's prediction performance at 5, 10, and 15 minutes is significantly better than GAN's, likely because the shared decoder primarily learns to predict shorter time intervals rather than longer ones.

\input{tables/tsmo_anomaly}
\input{tables/cranberry_anomaly}

\subsection{Incident Report Comparison}
\label{section: results3}
% we present the results of our model in detecting raw incident reports. The common goal of incident detection is to determine whether a road is in an incident state (i.e., a binary judgment). As our prediction results cover six future time points, 
In this section, we will compare our model's prediction results with incident reports and conventional AID methods. Both incident reports and conventional AID methods use a single value to describe whether a road is in an incident state at a given time step. However, to achieve earlier detection and provide operators with a reference for road conditions over a future period (rather than just a single point in time), we employed multi-step predictions, forecasting road conditions for the next half hour. To compare multi-prediction with a single value, we adopted a conservative strategy by taking the minimum value among these six predictions as the measurement standard. If this minimum value exceeds the threshold, an alert is triggered; otherwise, no alert is triggered.
\input{tables/incident}
\subsubsection{Detection Examples}
Using I-70 as an example, our prediction process during the evening period on 2023-01-11 is detailed in Appendix \ref{section: appendixA}. In this example, we use the Transformer model trained with labels generated by Algorithms \ref{alg:1} and \ref{alg:2} and optimization method detailed in section \ref{sec: train} and \ref{section: threshold}. 
In Appendix \ref{section: appendixA}, the red dashed line at 0.56 represents the adaptive threshold the threshold-tuning set generated. At 17:15, our model's predictions for the next six time steps are all below the threshold, so no alert is triggered (the blue line is used to indicate no alert). However, at 17:25, the minimum value exceeds the threshold, triggering an alert (the red line is used to indicate an alert). This alert ends at 17:45, as the minimum predicted value falls below the threshold at that time. The black text in Appendix \ref{section: appendixA} represents the current speed at the predicted time, while the red shading indicates the report time. From this, it can be observed that our model triggered an alert earlier than the incident report and even before the speed significantly dropped.
\begin{figure}[!htb]
    \centering
    \includegraphics[width=0.8\textwidth]{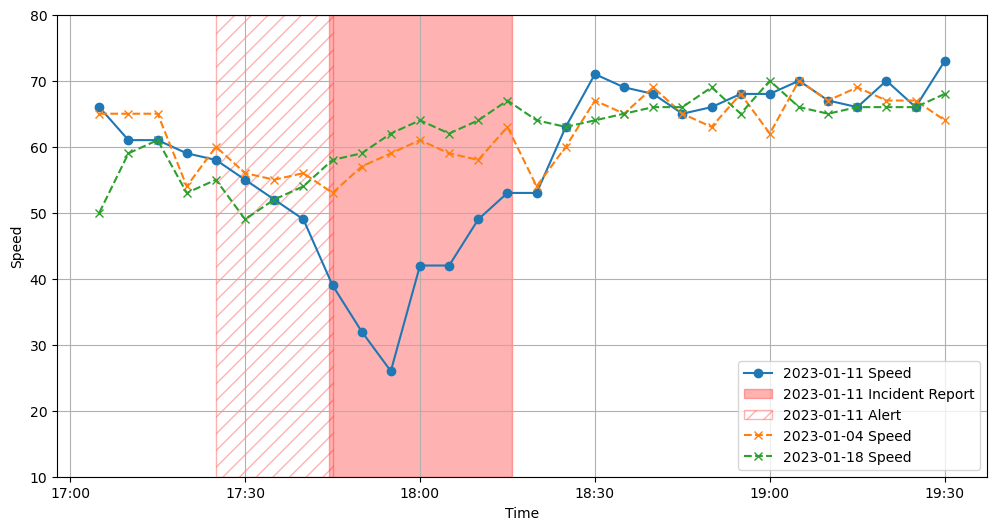}
    \caption{Speed, Alert, and Incident Report on 2023-01-11}
    \label{fig: TP_1}
\end{figure}

To facilitate observation, we compiled the alert, incident report, and speed reference for the case in Appendix \ref{section: appendixA} in Figure \ref{fig: TP_1}. By analyzing the speed on that day and comparing it with the speed on nearby days without incident reports (same day of the week, same time of day), we observed that at 17:25, although the speed of the target segment did not change significantly and seemed to exhibit a recurrent pattern, our model still triggered the alert. Figure \ref{fig: TP_1_a} shows the possible reasons for our model's early detection. While it is difficult to observe from the target segment alone, the speeds upstream and downstream had already exhibited clear incident characteristics at 17:25. The model likely captured these characteristics and triggered the alert. This example demonstrates that our model successfully learned the early characteristics of incidents.
\begin{figure}[!htb]
    \centering
    \includegraphics[width=0.8\textwidth]{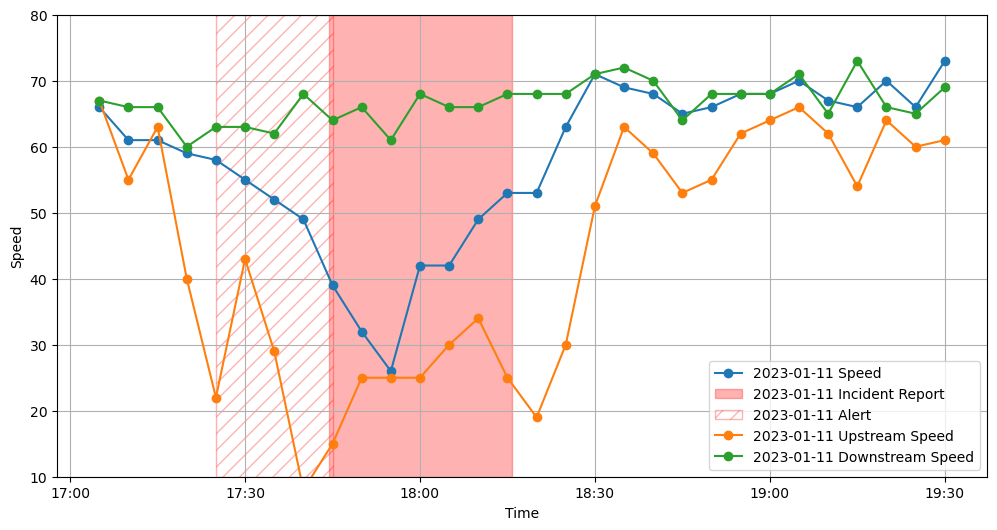}
    \caption{Upstream and Downstream speed on 2023-01-11}
    \label{fig: TP_1_a}
\end{figure}

Appendix \ref{section: appendixB} and Figure \ref{fig: TP_2} show our model's predictions in a more complex scenario involving multiple incidents on 2023-01-25. In this example, our model also predicted the occurrence of the incidents earlier. This further demonstrates our model's capability to effectively learn and identify early characteristics of incidents, even in more complex scenarios.
\begin{figure}[!htb]
    \centering
    \includegraphics[width=0.8\textwidth]{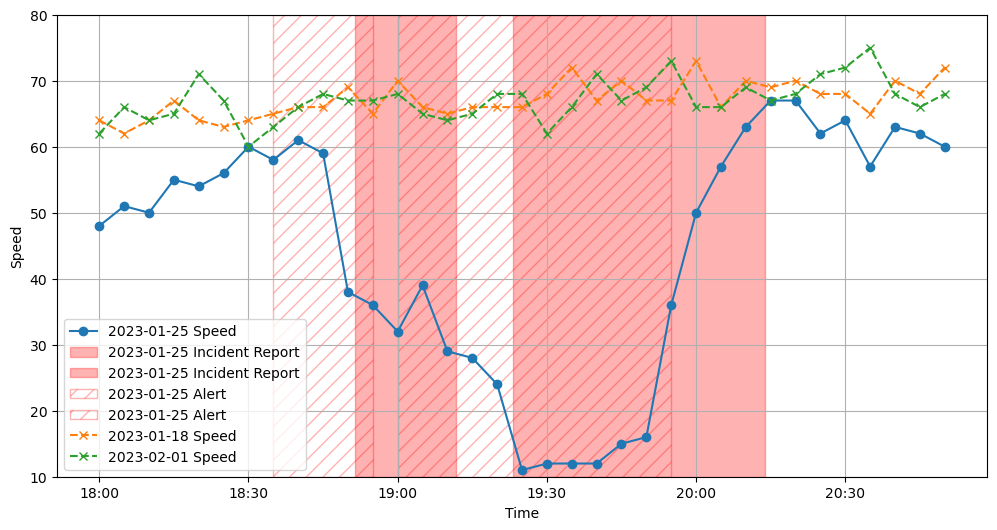}
    \caption{Speed, Alert, and Incident Report on 2023-01-25}
    \label{fig: TP_2}
\end{figure}

\subsubsection{Examples of detecting anomalies and possibly unreported incidents}
The above are two examples of direct prediction processes. During the periods mentioned, there were incident reports, and our model also issued alerts. However, there were also instances where no reports were present, but the alert was triggered (``False Positive" Cases), and cases where an incident report appeared later, but our model did not trigger an alert (``False Negative" Cases). Figures \ref{fig: FP_TSMO} and \ref{fig: FN_TSMO} illustrate these two kinds of cases, respectively.
\begin{figure}[!htb]
    \centering
    \begin{minipage}{0.6\textwidth}
        \centering
        \includegraphics[width=\textwidth]{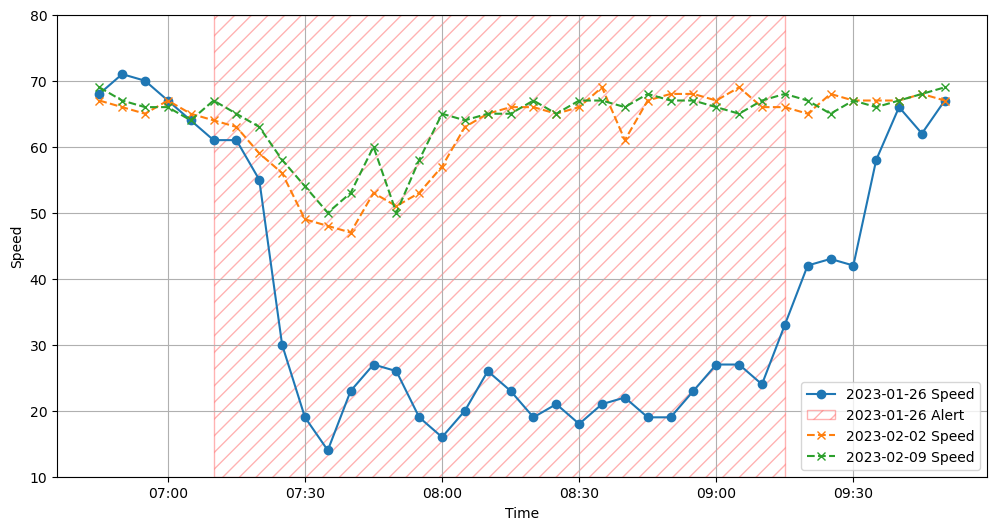}
        \subcaption[]{2023-01-26 Case}
    \end{minipage}
    \hfill
    \begin{minipage}{0.6\textwidth}
        \centering
        \includegraphics[width=\textwidth]{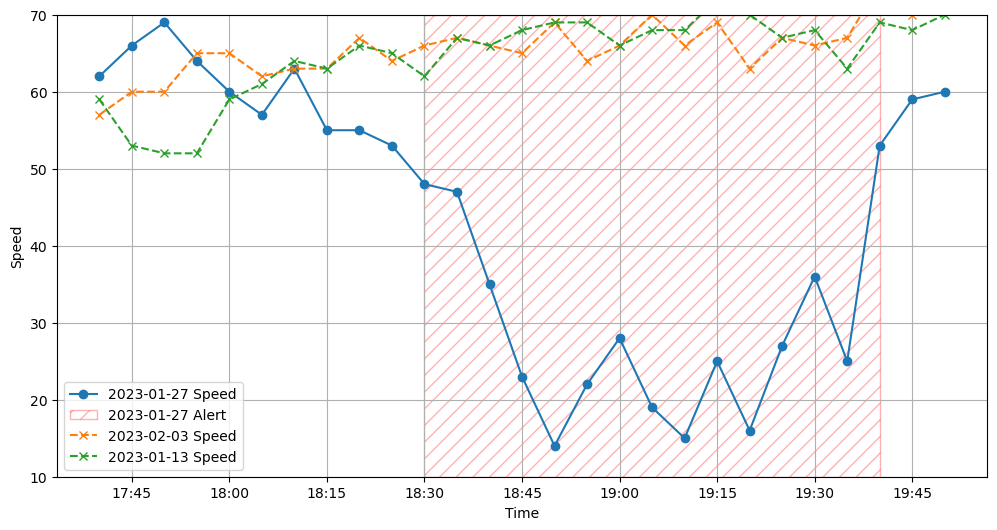}
        \subcaption[]{2023-01-27 Case}
    \end{minipage}
    \hfill
    \begin{minipage}{0.6\textwidth}
        \centering
        \includegraphics[width=\textwidth]{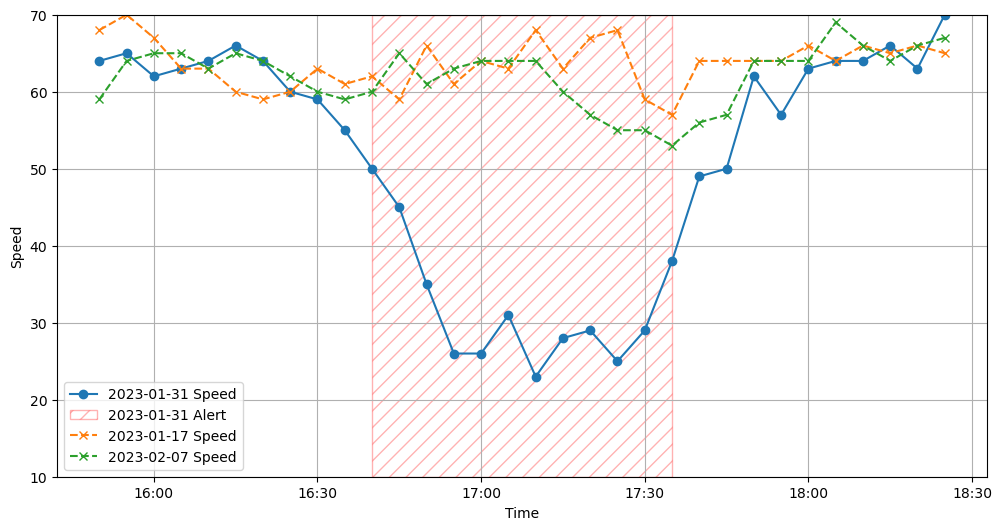}
        \subcaption[]{2023-01-31 Case}
    \end{minipage}
        \hfill
    \begin{minipage}{0.6\textwidth}
        \centering
        \includegraphics[width=\textwidth]{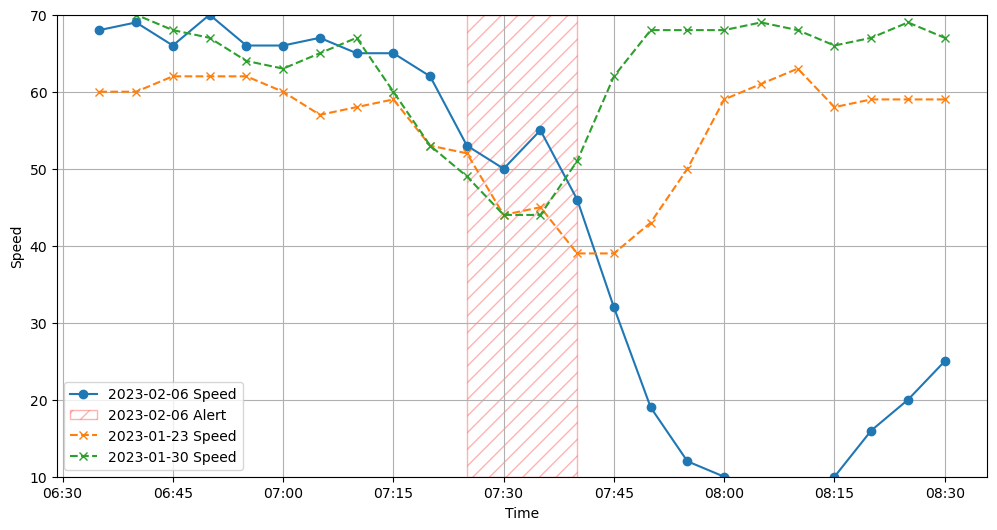}
        \subcaption[]{2023-02-06 Case}
    \end{minipage}
    \caption{The 'False Positive' Cases}
    \label{fig: FP_TSMO}
\end{figure}

Regarding false positive cases, we can observe in the four examples in Figure \ref{fig: FP_TSMO} that, despite no incident reports corresponding to the times on the dates indicated by the blue lines, there are clear anomalies when compared to recurrent patterns (represented by the orange and green dashed lines). Our model effectively captured these anomalies. As mentioned in the introduction, there is a significant missing issue with incident reports. These false alarms are likely due to the absence of incident reports, which underscores the importance of training with anomaly labels rather than relying solely on incident reports.

Regarding false negative cases, from Figure \ref{fig: FN_TSMO} , we can observe that compared to the recurrent pattern, the speed during the incident report period does not show significant changes, and most of these cases occur during non-peak hours. This could be because the incident reports are false, or do not have a significant impact, making it difficult for our model to detect them.
\begin{figure}[!htb]
    \centering
    \begin{minipage}{0.6\textwidth}
        \centering
        \includegraphics[width=\textwidth]{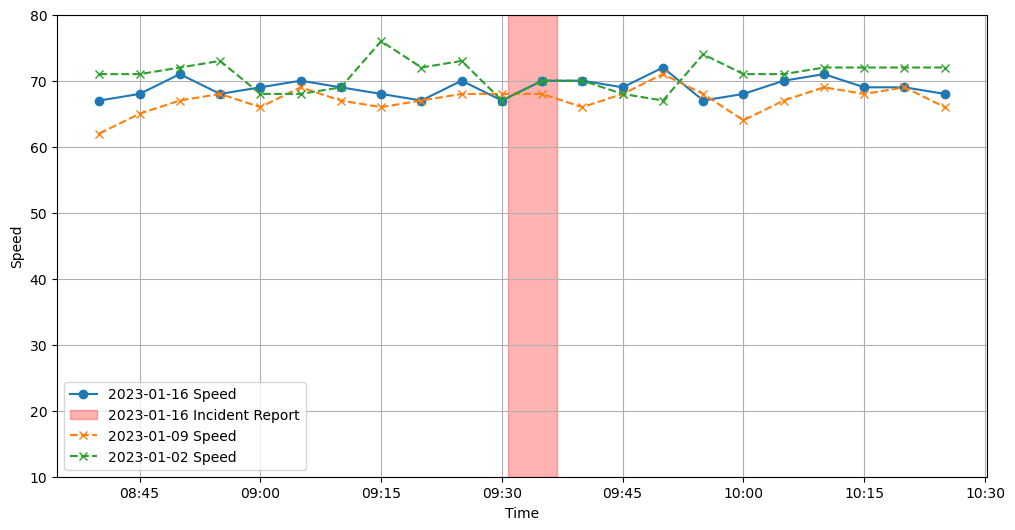}
        \subcaption[]{2023-01-16 Case}
    \end{minipage}
    \hfill
    \begin{minipage}{0.6\textwidth}
        \centering
        \includegraphics[width=\textwidth]{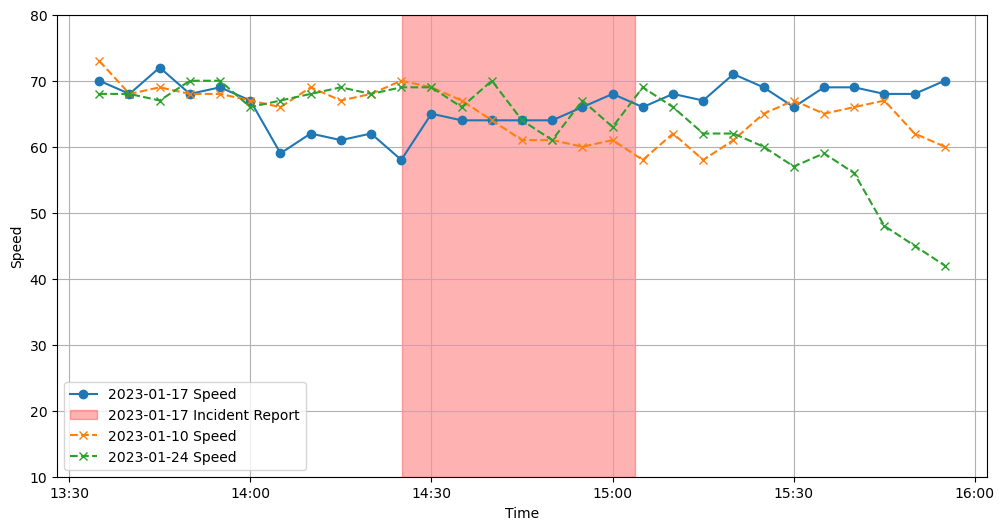}
        \subcaption[]{2023-01-17 Case}
    \end{minipage}
    \hfill
    \begin{minipage}{0.6\textwidth}
        \centering
        \includegraphics[width=\textwidth]{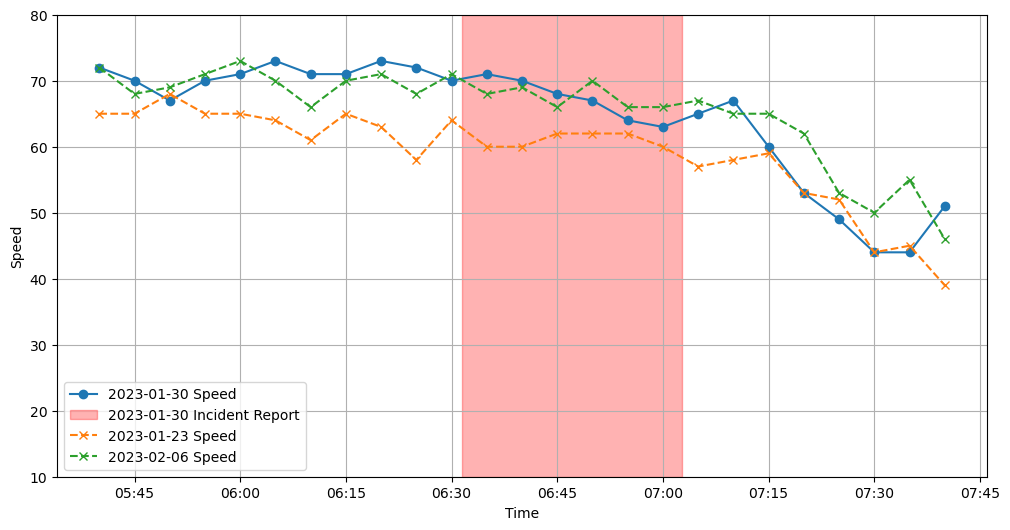}
        \subcaption[]{2023-01-30 Case}
    \end{minipage}
        \hfill
    \begin{minipage}{0.6\textwidth}
        \centering
        \includegraphics[width=\textwidth]{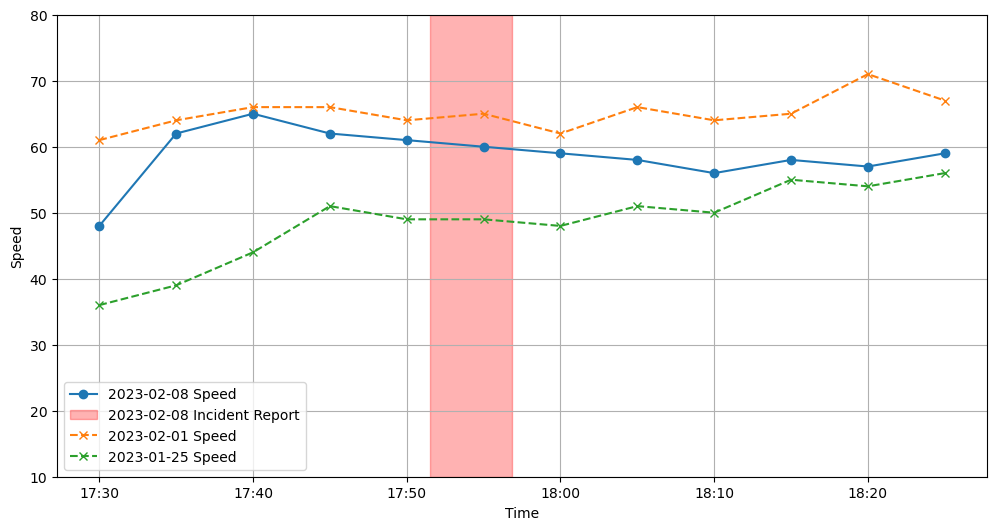}
        \subcaption[]{2023-02-08 Case}
    \end{minipage}
    \caption{The 'False Negative' Cases}
    \label{fig: FN_TSMO}
\end{figure}
\subsubsection{Overall Results}
\label{section: incident_overall}
The above are specific case analyses. To demonstrate the generalizability of our model, Table \ref{tab:Incident} shows the testing results for 10 segments across two road networks. The metrics used are defined as follows:
\begin{eqnarray}
\text{DR (Detection Rate)} &=& \frac{N_{\text{detected\_incident}}}{N_{\text{incident\_report}}}\label{eqn: DR} \\
\text{MTTD (Mean Time to Detection)} &=& \frac{\sum_{i=1}^{N_{\text{detected\_incident}}} (t_i^{\text{alarm}} - t_{i}^{\text{report}})}{N_{\text{detected\_incident}}}\label{eqn: MTTD2}\\
\text{FAR (False Alarm Rate)} &=& \frac{N_{\text{non-incident-non-anomaly}}}{N_{\text{alarm}}}\label{eqn: FAR}\\
\text{DR (S) (Detection Rate, Significant)} &=& \frac{N_{\text{detected\_significant\_incident}}}{N_{\text{significant\_incident\_report}}}\label{eqn: DRS}\\
\text{MTTD (S) (Mean Time to Detection, Significant)} &=& \frac{\sum_{i=1}^{N_{\text{detected\_significant\_incident}}} (t_i^{\text{alarm}} - t_{i}^{\text{report}})}{N_{\text{detected\_significant\_incident}}}\label{eqn: MTTDS}
\end{eqnarray}
The definitions of DR and MTTD are generally consistent with those used in other studies. The definition of detected incidents in Equation \ref{eqn: DR} is as follows: from the start to the end of an incident report, if the model triggers an alert during this period, it is considered detected. Note that detection after the report start time is typically less helpful if the incident has been reported to an operator. In Equation \ref{eqn: MTTD2}, $t_i^{\text{report}}$ refers to the start time of the report, and $t_i^{\text{alarm}}$ refers to the time when the model begins to trigger an alert. If there are multiple continuous alerts within one incident report, the earliest one is taken. Considering the incident report is typically after the actual (possibly unknown) occurrence time, negative values are allowed when calculating MTTD and indicate early detection/prediction. 

Considering the issues caused by incident missing report, as shown in Figures \ref{fig: report_missing} and \ref{fig: FP_TSMO}, we have optimized the definition of FAR (False Alarm Rate) in Equation \ref{eqn: FAR}. The numerator of FAR has been changed from the number of alarms not coinciding with any incident reports to the number of alarms not coinciding with any incident reports or anomalies. The anomalies here are those generated using the anomaly labels described in Section \ref{subsection: LabelDenoising}. This adjustment reduces the likelihood of alarms that captured anomalies, such as those in Figure \ref{fig: FP_TSMO}, being classified as false alarms.

The settings for DR (Significant) and MTTD (Significant) in Equations \ref{eqn: DRS} and \ref{eqn: MTTDS} aim to exclude the interference of some false reports or insignificant incidents in the evaluation process. A significant incident is defined as any incident that overlaps with our generated anomaly labels, i.e., the incidents retained in Step 1 of Algorithm \ref{alg:1}. The other calculation principles are consistent with those in Equations \ref{eqn: DR} and \ref{eqn: MTTD}.

According to Section \ref{section: literature}, we adopted the following conventional AID baselines, including statistical, comparative, and AI-based methods. It is important to note that, as we have already demonstrated the effectiveness of training with anomaly labels in the previous chapter, and since almost all existing studies use incident reports for training, all the baselines here, except for the statistical (OOD) and comparative method (CFOD), are trained using incident reports. Given the differences in input data, for example, we do not have loop detector data, we cannot directly use the original baseline models. However, we have made efforts to retain the proposed model's structure and adapt it by tuning it to our dataset.
\begin{itemize}
    \item \textbf{OOD}, Out of Distribution. OOD is based on the statistical method. we use the Seq2Seq with attention model \citep{bahdanau2014neural} to train speed predictions until convergence. The model will trigger an alert by measuring the SND (Standardized Normal Deviate) between observed and predicted values. The SND threshold is tuned based on the incidence rate of the incident reports. 
    \item \textbf{CFOD}, Chebyshev Distance First Order Directive. According to the early-detection method proposed by \citet{10287856}, for a given road segment, we first compute the speed difference between its recent input speed sequence and its corresponding monthly profile (i.e., the same day of the week and the time of the day over the past month). We then calculate the Chebyshev distance based on this difference. After normalizing the distance using parameters derived from the validation set, we compute its first-order derivative. The incident is considered to begin at a significant positive peak in the derivative and end at a substantial negative peak. 
    \item \textbf{SVM}, Support Vector Machine. Inspired by the idea proposed by \citet{yuan2003incident}, we train a binary classification SVM with an RBF kernel based on whether there is an incident report in the next time step. 
    \item \textbf{GAN}, Generative Adversarial Network. Inspired by the idea proposed by \citet{lin2020automated}, we first use a GAN (Generative Adversarial Network) to learn the characteristics of samples with incident reports and generate some fake samples. These fake samples, along with the real samples, are used to train the SVM.
    \item \textbf{TSSAE}, Temporal and Spatially Stacked Autoencoder. The model architecture is proposed by \citet{li2022real}. To adapt to our data, we fed all relevant input features of each segment within the sub-graph into a separate autoencoder. The autoencoders were first pre-trained in an unsupervised manner and then fine-tuned using incident labels. For each segment, we evaluated two settings—with and without GAN-based data augmentation—and reported the best-performing results. 
    \item \textbf{Ours}. The proposed method using the Transformer model trained without ahead-labeled anomaly labels.
    \item \textbf{Ours (a)}. The proposed method using the Transformer model trained with 15-minute ahead-labeled anomaly labels.
\end{itemize}

In Table \ref{tab:Incident}, we first crossed out some results that lack practical significance, such as extremely high FAR ($>$0.8) or cases where both DR and DR(S) are 0. The reasons for crossed out are marked in red.

Our model achieved a low FAR overall while obtaining high DR and low MTTD, especially with DR(S) almost all being 1. This is naturally related to our model is trained by anomaly labels and reflects that our model has successfully detected the incidents marked as significant. Furthermore, even without ahead labeling, all our MTTD values are negative, indicating detection before the incident reports. This shows that the addition of anomaly samples has already incorporated features that precede the incident reports.

After ahead labeling, our model further reduced MTTD in 8 out of 10 cases, demonstrating the significance of ahead labeling. In most cases, DR remained unchanged, but FAR generally increased slightly, though still within an acceptable low range. This is reasonable, as the ahead labeling process can label some non-anomaly samples as anomalies. For the I-79S case where DR, FAR, and MTTD all worsened, we may need to reduce the ahead labeling time $\theta_{ahead}$ to prevent the inclusion of too many non-anomaly samples.

Among baseline methods, OOD only shows some practical significance in Howard County US-40, while other results face very high FAR and low DR. Notably, they all converge well in the speed prediction model. This highlights the drawbacks of statistical methods, where outliers in SND are not necessarily anomalies but could be due to model prediction errors and probe vehicle speed observation errors. Although statistical methods have performed well in some studies, probe vehicle data may still be too noisy to support this method. 

Regarding CFOD, on the Howard County map, we observe that CFOD suffers from a significantly low detection rate and a relatively high false alarm rate (FAR), especially on I-70E, I-70W, and US-40W segments. This suggests that relying solely on temporal differences may be insufficient for effective incident detection. Furthermore, CFOD performs poorly on the Cranberry map, implying that incidents in Cranberry may exhibit less temporal fluctuation and instead manifest primarily through spatial patterns, as explained in section \ref{subsection: LabelDenoising}, which CFOD cannot capture. Although CFOD shows limited overall performance, it reveals an interesting insight: in the US-40W and I-695A segments of Howard County, CFOD is able to identify certain incident reports not labeled as significant in our dataset. This suggests that our labeling criteria may miss incidents with localized, segment-level temporal anomalies, which could be valuable for refining future labeling strategies. Moreover, given the strong theoretical interpretability of CFOD, we further analyzed its false alarms and found that most of them were caused by point anomalies—i.e., anomalies that occur at only a single time point. These point anomalies may result from noisy input data or may correspond to incidents that have limited impact on traffic conditions and were therefore not labeled as significant.

Regarding SVM, the results of the SVM are entirely lacking in practical significance. This may be due to the difficulty in learning from incident reports. This inference is supported by our observation that, when trained using anomaly labels (see Tables \ref{tab:T_Prediction} and \ref{tab:C_Prediction}), the models were able to converge in most cases—suggesting that the quality of incident report labels may hinder effective learning. GAN achieved good convergence across all cases in Howard County. This may be because GAN learned the anomaly characteristics from the incident reports, and its generator produced samples similar to those we generated using Algorithm \ref{alg:1}. 
By using incident reports directly for training, the DR of GAN in I-70E and I-695A was higher than our method, although their FAR was also relatively high. Regarding MTTD, the GAN achieved the best performance on I-76. However, it is important to note that the GAN's DR on this segment is very low, indicating that the low average MTTD may result from only a few detected incidents. Moreover, the FAR is high—approaching our unacceptable threshold of 0.8—which further limits the practical significance of these MTTD results. Overall, GAN's MTTD was still worse than our method, reflecting the lack of early features in the incident reports. In addition, GAN performed poorly in Cranberry Township, even worse than SVM in I-79S and I-79N, indicating that the samples generated by GAN actually added noise. This suggests that GAN still struggles to learn incident characteristics in the Cranberry Map. In combination with the CFOD results, it is reasonable to infer that the GAN may struggle to learn spatial anomalies effectively. In the future, we might consider using prior knowledge, such as slowdown speed, or by designing a discriminator with a graph-based structure, to guide the learning process of GAN.

As for TSSAE, the results are similar to those of SVM in that the model fails to converge. While we observed clear convergence during the pretraining stage, the model showed little to no convergence during finetuning. We also attempted to replace the architecture with the Transformer, but it similarly failed to converge. This further suggests that under noisy labeling conditions, simply improving the model architecture is insufficient to enhance performance. Notably, as in the original paper, we also attempted GAN-based data augmentation prior to training. However, the model still failed to converge and performed worse than the GAN-SVM combination. This likely reflects limitations in both data volume and label quality, which are insufficient to support effective deep learning model training even fine-tuning.
\subsection{Sensitivity Analysis on Slowdown Speed Threshold}
\label{section: sens}
\input{tables/senstivity}
In Algorithm \ref{alg:1}, we set $\theta_1$ and $\theta_2$ to determine the slowdown speed threshold $\theta_{\text{SD}}^i$. An incident report is marked as significant if it causes a slowdown speed value higher than this threshold. Additionally, if the slowdown speed continuously remains above $\theta_{\text{SD}}^i$, it is also considered anomalous. We now present how the choice of threshold impacts the model's performance. It is important to note that since the threshold selection directly affects the ground truth labels for anomalies, direct comparisons using metrics such as those in Table \ref{tab: tsmo_a} or Table \ref{tab: Cranbeery_a} are not meaningful. Instead, similar to Table \ref{tab:Incident}, we compare the detection results against incident reports. However, care must be taken when interpreting metrics like FAR, DR(S), and MTTD(S), as different thresholds result in different sets of significant anomalies or incident reports, which makes direct comparisons potentially misleading.

Take the I-70 segment as an example. As shown in Table \ref{tab:sen}, in the absence of ahead labeling, reducing $\theta_{\text{SD}}$ from 30 mph to 25 mph (i.e., comparing Row 1 and Row 3) introduces more incident reports into the training set. This leads to a notable increase in the Detection Rate (DR) but also causes a rise in the False Alarm Rate (FAR) since the inclusion of less severe incidents makes the model more likely to misclassify normal conditions as anomalies. Under the setting with ahead labeling, the DR also improves, while the FAR decreases numerically, which seems to be in contrast to the case with no ahead-labeling. Although the FAR does not appear to increase, this is mainly due to our definition, which states that the denominator of FAR is the total number of alerts. Lowering the threshold introduces more mild anomalies into the training set, making the model more sensitive and likely to trigger more alerts during inference. However, many of these alerts are considered anomalies under the relaxed threshold, thereby diluting the FAR.

When the slowdown speed threshold is further reduced to 20 mph, the model—without ahead labeling (Row 5)—fails to identify some incidents marked as significant (i.e., DR(S) is no longer 1). This is because the lower threshold introduces more anomalous samples that are difficult to distinguish from normal cases, even normal cases, into the training labels. This results in a drop in overall DR compared to the case where the threshold is set to 25 mph (Row 3). Compared to the threshold of 30 mph (Row 1), although the DR remains the same, the FAR is noticeably higher. When ahead-labeling is introduced, the proportion of incident samples in the training set increases, leading the model to make more aggressive predictions. However, this also results in more mild anomalies—or even normal samples—being mislabeled as anomalies, reducing the purity of training labels. Consequently, although DR increases significantly, the FAR also rises markedly.

Finally, MTTD does not show a significant change as the threshold decreases, which highlights the importance of ahead labeling for enabling early detection/prediction.
\section{Conclusion and Future Work}
\label{section: conclusion}
In this study, we address an early anomaly detection or prediction problem that is more practically useful and scalable than conventional AID models. We propose a method that combines incident reports and traffic domain knowledge, i.e., the correlation between slowdown speed and incident, to generate anomaly labels and train a deep learning model for early anomaly detection. This method aims to overcome inherent issues in incident reports, including false reports, insignificant incident reports, delayed reports, and missing reports. We adopted ahead labeling and multi-step prediction strategies, allowing the model to predict incidents earlier than traditional methods. We also applied various machine learning strategies to address the issues of insufficient anomaly samples and data imbalance. Unlike existing research in the field of AID, our method aims to capture a broader range of anomalies beyond \textit{after the fact} incident reports (including early incident characteristics and unreported anomalies) and, more importantly, it provides early alerts for anomalies that may disrupt traffic flow in the short term with regional impacts. Furthermore, our method is highly scalable as it relies solely on ubiquitously available data. We validated our model on ten road sections in two road networks, achieving results that surpassed the baselines. This experiment shows a good trade-off between precision and recall, resulting in anomaly alerts 5-42 min earlier than Waze reports. 

Although the data used in this study is theoretically available in real-time, we need to coordinate with data providers to ensure we can access the real-time data required by the model as early as possible. This will maximize the model's advantage in early prediction/detection for potential future applications. Besides, this study also has certain limitations that can be improved. First, regarding the selection of prior knowledge, we mainly chose edges with a high correlation between the occurrence of incidents and high slowdown speeds. In fact, due to the topological structure, not all edges have this correlation. Bases on our definition of slowdown speed, this high correlation requires that the segment be relatively long and that most incidents do not occur at the very beginning of each segment. For those that do not exhibit this high correlation, we may need to consider redefining the slowdown speed or using other values, such as the travel time index. Although most case studies demonstrate that our method effectively captures both spatial and temporal anomalies—and clearly outperforms approaches that focus solely on temporal patterns—comparisons with baselines suggest that some anomalies with purely temporal, segment-level effects may still be overlooked. Future research could explore more integrated labeling strategies that jointly capture both spatial and temporal anomalies in a cohesive manner. Semi-supervised learning or agent-based learning will be a potential solution for this issue. Moreover, our algorithm is also difficult to handle for segments where only one or two incidents occur per year, as the validation and test sets may not contain any incidents. This might require using simulator data to generate sufficient training samples and finetuning with real-world incident samples. Additionally, we can tune a reasonable ahead-labeling period, which is also a direction worth exploring in future research.

\section*{Acknowledgement}
This research is supported by US Department of Transportation Exploratory Advanced Research Award 693JJ321C000013. The contents of this paper reflect the views of the authors only, who are responsible for the facts and the accuracy of the information presented herein.

\clearpage
\bibliographystyle{johd}
\bibliography{bib}
\clearpage
\appendix
\section{Prediction/Detection Example on 2023-01-11}
\label{section: appendixA}
\input{images/detection_example_1}
\clearpage
\section{Prediction/Detection Example on 2023-01-25}
\label{section: appendixB}
\input{images/detection_example_2}
\end{flushleft}
\end{document}

%% file: tables/sd_table.tex
\begin{table}[!htb]
\scriptsize
\centering
\caption{Correlation between Incident Report and Slowdown Speed}
\label{table:sd_table}
\begin{tabular}{@{}ccccc@{}}
\toprule
\textbf{Study Segment} & \textbf{Region} & \textbf{Time Reported} & \textbf{Incident Reports in} & \textbf{Overlap with Incident} \\ 
\textbf{Location} & \textbf{} & \textbf{Incidents (\%)} & \textbf{Top Slowdown Speeds (\%)} & \textbf{Reports (\%)} \\ \midrule
I-70W   & Howard County, MD  &   1.27                                & 9.99 (Top 3\%)                                      & 60 (Top 3\%)                     \\
I-70E   & Howard County, MD  &   3.23                                & 15.34 (Top 5\%)                                      & 60 (Top 5\%)                     \\ 
I-79S   & Cranberry Township region, PA   &   3.81                                & 13.40 (Top 5\%)                                      & 65 (Top 5\%)                     \\
I-79N   & Cranberry Township region, PA   &  5.20                                & 10.36 (Top 7\%)                                      & 61 (Top 7\%)                     \\ 
\bottomrule
\end{tabular}
\end{table}

%% file: alg1.tex
\begin{algorithm}
\begin{footnotesize}
\begin{algorithmic}
\caption{Incident Label Denoising}
\label{alg:1}
\State \underline{\textbf{Inputs}}: \parbox[t]{\dimexpr\linewidth-\algorithmicindent}{incident report binary matrix $\mathbf{INC^i}\in \mathbf{R}^{number\_of\_days\times time\_of\_the\_day}$; \\
slowdown speed matrix $\mathbf{SD^i}\in \mathbf{R}^{number\_of\_days\times time\_of\_the\_day}$; \\
 removal percentage threshold $\mathbf{\theta_1}$;\\
 addition percentage threshold $\mathbf{\theta_2}$;\\
 minimum duration for a prolonged anomalies to be labeled $\theta_t$}
\State \underline{\textbf{Outputs}}: anomaly label matrix $\mathbf{ANO^i}\in \mathbf{R}^{number\_of\_days\times time\_of\_the\_day}$
\State \underline{\textbf{Initialization}}: \parbox[t]{\dimexpr\linewidth-\algorithmicindent}{top $\mathbf{n}\%$ slowdown speed is the indicator of anomalies
}
\State \underline{\textbf{Step 0: Generate the abnormal slowdown speed matrix}}:
\State obtain the abnormal slowdown speed threshold:  $\mathbf{\theta_{SD}^i}$ = Percentile(vec($\mathbf{SD^i}$), n)
\State compute the abnormal slowdown speed binary matrix: $\mathbf{ASD^{i}_{(p,q)}} = \mathbb{I}(\mathbf{SD^{i}_{(p,q)}}\geq\mathbf{\theta_{SD}^i})$ for any element \( (p, q) \) in the matrix 

\State \underline{\textbf{Step 1: Remove insignificant incident reports}}: 
\State initialize the significant incident report matrix $\mathbf{SIR^i}$ with zeros: $\mathbf{SIR^i} \gets \mathbf{0}^{number\_of\_days\times time\_of\_the\_day}$
\For {$p = 0, 1, \ldots, number\_of\_days -1$}
\For {$q = 0, 1, \ldots,  time\_of\_the\_day -1$}
    \If{($q=0$ and $\mathbf{INC^i_{(p,q)}}=1$) or ($\mathbf{INC^i_{(p,q-1)}}=0$ and $\mathbf{INC^i_{(p,q)}}=1$)}\Comment{find the start of an incident report}
    \State  $r=0$
    \While{$\mathbf{INC^i_{(p,q+r)}}=1$ \text{and} $q+r\leq time\_of\_the\_day$} \Comment{check how long the incident report lasted}
    \State $r\gets r+1$
    \EndWhile
    \If{\text{SUM}($\mathbf{ASD^i_{(p,q:q+r)}}$)$\geq$1} \Comment{check if the incident report is significant}
    \State set $\mathbf{SIR^i_{(p,q:q+r)}}=1$ 
    \EndIf
    \EndIf
\EndFor
\EndFor
\State \underline{\textbf{Step 2: Label prolonged significant anomalies}}:
\State initialize the prolonged significant anomalies matrix $\mathbf{PSA^i}$ with zeros: $\mathbf{PSA^i} \gets \mathbf{0}^{number\_of\_days\times time\_of\_the\_day}$
\For {$p = 0, 1, \ldots, number\_of\_days -1$}
\For {$q = 0, 1, \ldots,  time\_of\_the\_day -1$}
    \If{($q=0$ and $\mathbf{ASD^i_{(p,q)}}=1$) or ($\mathbf{ASD^i_{(p,q-1)}}=0$ and $\mathbf{ASD^i_{(p,q)}}=1$)}\Comment{find the start of an anomaly}
    \State  $r=0$
    \While{$\mathbf{ASD^i_{(p,q+r)}}=1$ \text{and} $q+r\leq time\_of\_the\_day$} \Comment{check how long the anomaly lasted}
    \State $r\gets r+1$
    \EndWhile
    \If{r$\geq$$\theta_t$} \Comment{check if the anomaly is prolonged}
    \State set $\mathbf{PSA^i_{(p,q:q+r)}}=1$ 
    \EndIf
    \EndIf
\EndFor
\EndFor
\State \underline{\textbf{Step 3: Check the removal and addition percentage threshold}}: 
\State compute the removal percentage $\mathbf{rm{\%}}$  = 1-($\text{SUM}(\mathbf{SIR^i})/\text{SUM}(\mathbf{INC^i})$)
\State determine which labels are newly added $\mathbf{ADD^i} = \text{ReLU}(\mathbf{ASD^i}-\mathbf{INC^i}, 0) = \text{ReLU}(\mathbf{ASD^i}-\mathbf{SIR^i}, 0)$
\State compute the addition percentage $\mathbf{add{\%}}$ = $\text{SUM}(\mathbf{ADD^i})/\text{SUM}(\mathbf{INC^i})$
\If{$\mathbf{rm{\%}}>\mathbf{\theta_1}$ and $\mathbf{add{\%}}\leq\mathbf{\theta_2}$}
\State $n\gets n+\alpha$
\State Go back to \textbf{Step 0}
\ElsIf{$\mathbf{rm{\%}}\leq\mathbf{\theta_1}$ and $\mathbf{add{\%}}>\mathbf{\theta_2}$}
\State $n\gets n-\alpha$
\State Go back to \textbf{Step 0}
\ElsIf{$\mathbf{rm{\%}}\leq\mathbf{\theta_1}$ and $\mathbf{add{\%}}\leq\mathbf{\theta_2}$}
\State Go to \textbf{Step 4}
\Else
\State the settings of $\mathbf{\theta_1}$ and $\mathbf{\theta_2}$ are unreasonable, terminate the algorithm and reinitialize

\EndIf

\State \underline{\textbf{Step 4: Generate anomaly labels}}: 
\State combine significant incident report and newly added labels by $\mathbf{ANO^i} = \mathbf{SIR^i}+\mathbf{ADD^i}$ 
\State output $\mathbf{ANO^i}$
\end{algorithmic}
\end{footnotesize}
\end{algorithm}

%% file: alg2.tex
\begin{algorithm}
\begin{footnotesize}
\begin{algorithmic}
\caption{Anomaly Ahead Labeling}
\label{alg:2}
\State \underline{\textbf{Inputs}}: \parbox[t]{\dimexpr\linewidth-\algorithmicindent}{anomaly label matrix $\mathbf{ANO^i}\in \mathbf{R}^{number\_of\_days\times time\_of\_the\_day}$\\
 ahead label duration $\theta_{ahead}$}
\State \underline{\textbf{Outputs}}: ahead-labeled anomaly label matrix $\mathbf{AAN^i}\in \mathbf{R}^{number\_of\_days\times time\_of\_the\_day}$

\State initialize the ahead-labeled anomaly label matrix $\mathbf{AAN^i}$ with the anomaly label matrix $\mathbf{AAN^i}\gets \mathbf{ANO^i}$
\For {$p = 0, 1, \ldots, number\_of\_days -1$}
\For {$q = 1, \ldots,  \theta_{ahead} -1$}
    \If{ $\mathbf{ANO^i_{(p,q)}}=1$}\Comment{find the start of an anomaly}
    \State  $\mathbf{AAN^i_{(p,:q)}}=1$
    \EndIf
\EndFor
\For {$q = \theta_{ahead} -1, \ldots, time\_of\_the\_day-1 $}
    \If{ $\mathbf{ANO^i_{(p,q)}}=1$ and $\mathbf{ANO^i_{(p,q-1)}}=0$}\Comment{find the start of an anomaly}
    \State  $\mathbf{AAN^i_{(p,q-\theta_{ahead}:q)}}=1$
    \EndIf
\EndFor
\EndFor
\State output $\mathbf{AAN^i}$
\end{algorithmic}
\end{footnotesize}
\end{algorithm}

%% file: alg3.tex
\begin{algorithm}
\begin{footnotesize}
\begin{algorithmic}
\caption{Threshold Adaptation}
\label{alg:3}
\State \underline{\textbf{Inputs}}: 
\parbox[t]{\dimexpr\linewidth-\algorithmicindent}{
Training set $\mathcal{D}_{\text{train}}$, \\
Threshold tuning set $\mathcal{D}_{\text{tune}}$, \\
Number of epochs $E$
}

\State \underline{\textbf{Outputs}}:
Best model parameters $\theta^*$, optimal threshold $\tau^*$

\State initialize model parameters $\theta$
\State initialize best validation F1 score F$1_{best} \gets 0$
\For{$epoch = 1$ to $E$}
    \State Train the model $\theta$ on $\mathcal{D}_{\text{train}}$ using WBCE loss
    \State initialize best threshold F1 score F$1_{\tau} \gets 0$
    \For{$\tau = 0.01, 0.02, \ldots, 0.99$}
        \State Apply model $\theta$ to $\mathcal{D}_{\text{tune}}$ with threshold $\tau$
        \State Compute F1 score $\text{F}1_{\tau'}$ under threshold $\tau$
        \If{F$1_{\tau'} > \text{F}1_{\tau}$}
            \State $\text{F}1_{\tau} \gets \text{F}1_{\tau'}$
            \State $\tau^* \gets \tau$
        \EndIf
    \EndFor
    \State {Apply model $\theta$ to $\mathcal{D}_{\text{val}}$ using threshold $\tau^*$}
    \State {Compute validation F1 score $\text{F}1_{val}$}
    \If{{$\text{F}1_{val} > \text{F}1_{best}$}}
        \State {$\text{F}1_{best} \gets \text{F}1_{val}$}
        \State {Save model parameters $\theta^* \gets \theta$}
    \EndIf
\EndFor

\end{algorithmic}
\end{footnotesize}
\end{algorithm}

%% file: images/recall_precision_F1.tex
\begin{figure}[!htb]
    \centering

    \begin{subfigure}[b]{0.45\textwidth}
        \centering
        \includegraphics[width=\textwidth]{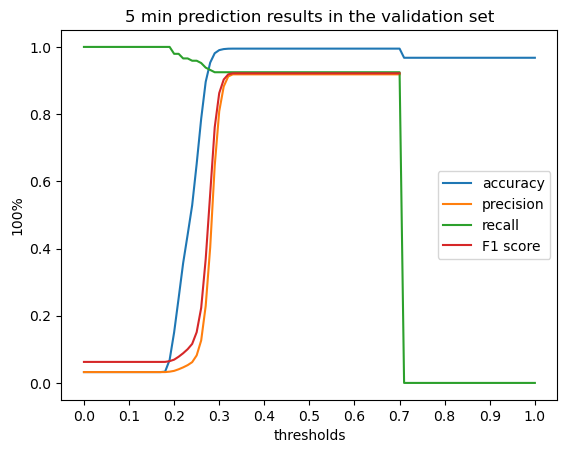}
    \end{subfigure}
    \hfill
    \begin{subfigure}[b]{0.45\textwidth}
        \centering
        \includegraphics[width=\textwidth]{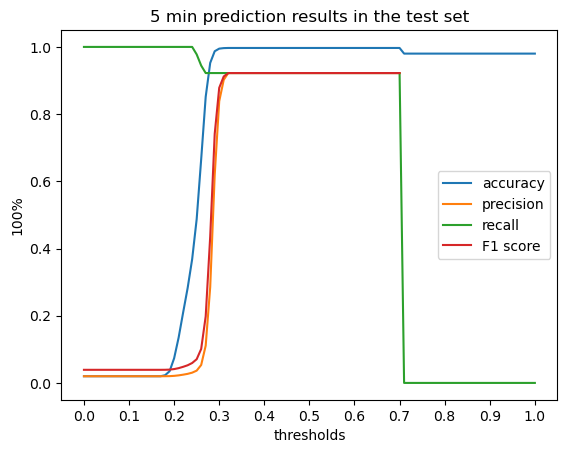}
    \end{subfigure}

    \vspace{0.5cm} % 添加垂直间距

    \begin{subfigure}[b]{0.45\textwidth}
        \centering
        \includegraphics[width=\textwidth]{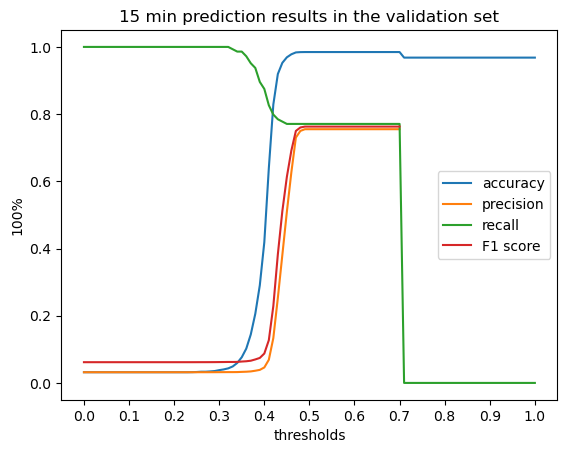}
    \end{subfigure}
    \hfill
    \begin{subfigure}[b]{0.45\textwidth}
        \centering
        \includegraphics[width=\textwidth]{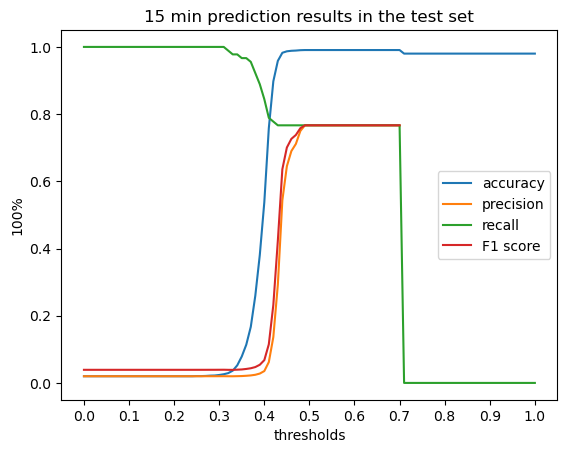}
    \end{subfigure}

    \vspace{0.5cm} % 添加垂直间距

    \begin{subfigure}[b]{0.45\textwidth}
        \centering
        \includegraphics[width=\textwidth]{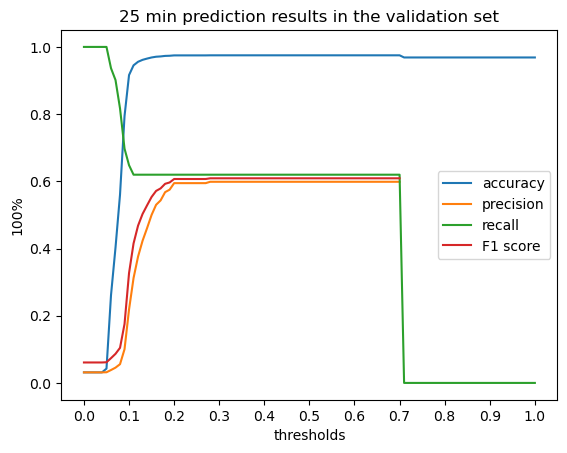}
    \end{subfigure}
    \hfill
    \begin{subfigure}[b]{0.45\textwidth}
        \centering
        \includegraphics[width=\textwidth]{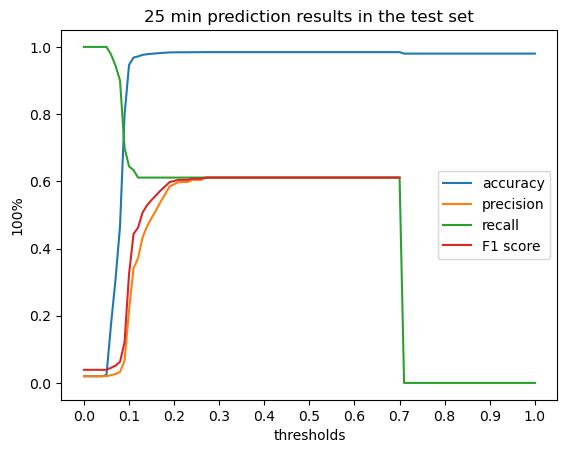}
    \end{subfigure}

    \caption{Recall, Precision, F1 score, Accuracy Plot (Without Ahead Labelling)}
    \label{fig: recall_pre_without_ahead}
\end{figure}

%% file: images/recall_precision_F1_15.tex
\begin{figure}[!htb]
    \centering

    \begin{subfigure}[b]{0.45\textwidth}
        \centering
        \includegraphics[width=\textwidth]{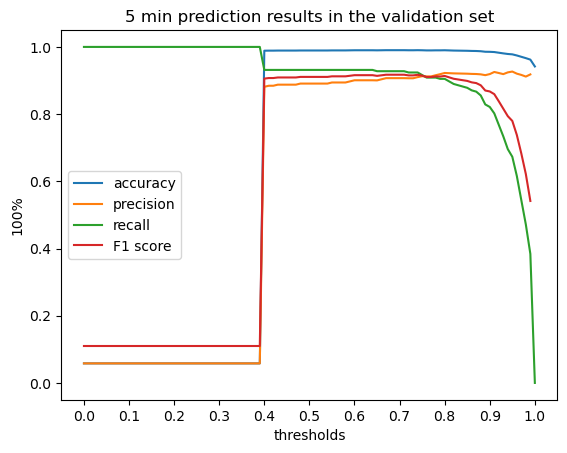}
    \end{subfigure}
    \hfill
    \begin{subfigure}[b]{0.45\textwidth}
        \centering
        \includegraphics[width=\textwidth]{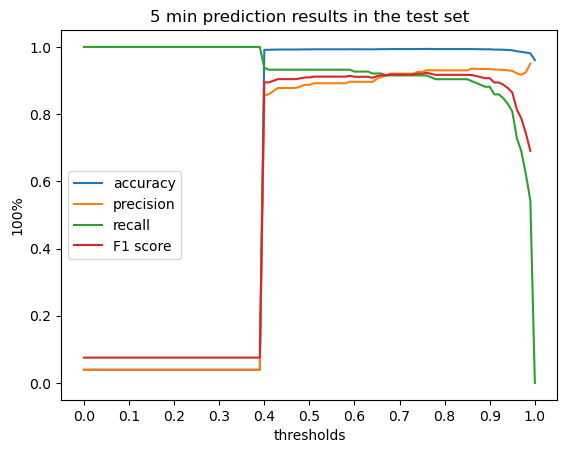}
    \end{subfigure}

    \vspace{0.5cm} % 添加垂直间距

    \begin{subfigure}[b]{0.45\textwidth}
        \centering
        \includegraphics[width=\textwidth]{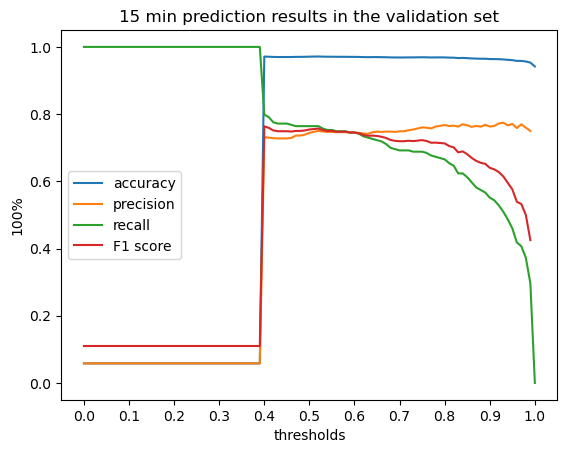}
    \end{subfigure}
    \hfill
    \begin{subfigure}[b]{0.45\textwidth}
        \centering
        \includegraphics[width=\textwidth]{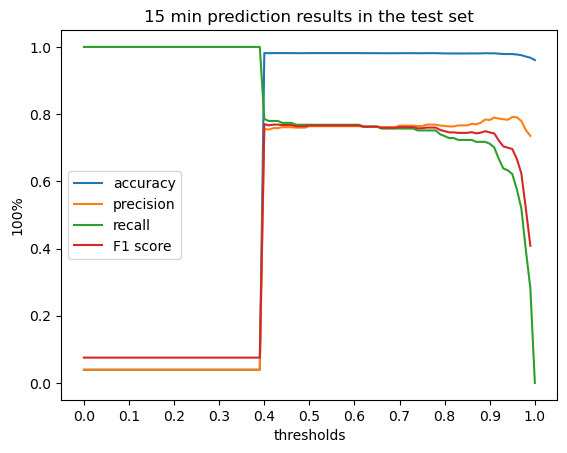}
    \end{subfigure}

    \vspace{0.5cm} % 添加垂直间距

    \begin{subfigure}[b]{0.45\textwidth}
        \centering
        \includegraphics[width=\textwidth]{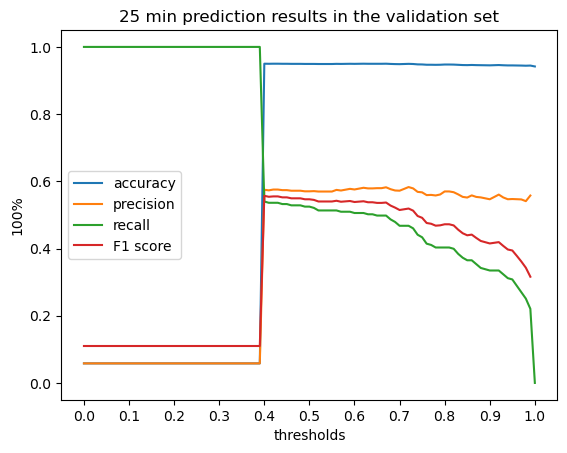}
    \end{subfigure}
    \hfill
    \begin{subfigure}[b]{0.45\textwidth}
        \centering
        \includegraphics[width=\textwidth]{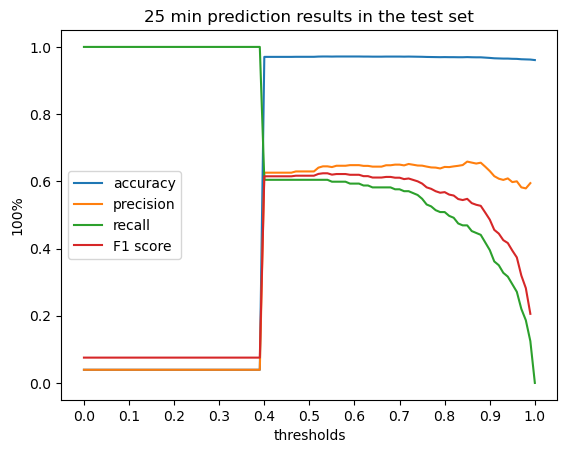}
    \end{subfigure}

    \caption{Recall, Precision, F1 score, Accuracy Plot (With Ahead Labelling)}
    \label{fig: recall_pre_with_ahead}
\end{figure}

%% file: tables/tsmo_anomaly.tex
\begin{table}[!htb]
%\scriptsize
%\tiny
\fontsize{6}{6.8}\selectfont
	\begin{center}
		\begin{tabular}{c|c|c|c|c|c|c|c|c|c}
		    \hline 
                \multicolumn{2}{c|}{Location} & \multirow{2}*{Metrics} & \multirow{2}*{Model} & \multicolumn{6}{|c}{Time}\\
                \cline{1-2}\cline{5-10}
		      County & Road &  &  & 5min & 10min & 15min& 20min & 25min & 30min\\\hline % Title
              \multirow{18}*{Howard, MD} & 
                \multirow{18}*{I-70E} & 
                     \multirow{6}*{Recall} &
                     RF & 0.63 & 0.52 & 0.38 & 0.26 & 0.18 & 0.08 \\
                    &&& SVM & 0.89 & 0.87 & 0.86 & 0.84 & 0.80 & 0.75 \\
                    &&& GAN & 0.87 & 0.86 & 0.83 & 0.78 & 0.73 & 0.69 \\
                    &&& LightGBM & 0.79 & 0.76 & 0.70 & 0.58 & 0.53 & 0.46 \\
                    &&& Trans   & 0.91 & 0.83 & 0.74 & 0.67 & 0.59 & 0.52 \\
                    &&& G-Trans & 0.91 & 0.83 & 0.74 & 0.67 & 0.62 & 0.52 \\
                    \cline{3-10}
                && \multirow{6}*{Precision} &
                        RF & 0.79 & 0.73 & 0.68 & 0.62 & 0.54 & 0.42 \\
                    &&& SVM & 0.38 & 0.34 & 0.29 & 0.26 & 0.21 & 0.19 \\
                    &&& GAN & 0.47 & 0.43 & 0.40 & 0.36 & 0.31 & 0.28 \\
                    &&& LightGBM  & 0.84 & 0.75 & 0.68 & 0.70 & 0.6 & 0.55 \\
                    &&& Trans   & 0.91 & 0.77 & 0.63 & 0.49 & 0.38 & 0.33 \\
                    &&& G-Trans & 0.91 & 0.83 & 0.74 & 0.67 & 0.60 & 0.52 \\
                    \cline{3-10}
                && \multirow{6}*{F1 Score} &
                        RF & 0.70 & 0.61 & 0.48 & 0.37 & 0.28 & 0.13 \\
                    &&& SVM & 0.54 & 0.49 & 0.43 & 0.40 & 0.34 & 0.30 \\
                    &&& GAN & 0.61 & 0.57 & 0.54 & 0.49 & 0.43 & 0.39 \\
                    &&& LightGBM & 0.81 & 0.75 & 0.68$^\dagger$ & 0.64$^\dagger$ & 0.56$^\dagger$ & 0.50$^\dagger$ \\
                    &&& Trans   & 0.91$^*$ & 0.80$^\dagger$ & 0.68$^\dagger$ & 0.57 & 0.47 & 0.41 \\
                    &&& G-Trans & 0.91$^*$ & 0.83$^*$ & 0.74$^*$ & 0.67$^*$ & 0.61$^*$ & 0.52$^*$ \\
                \hline  % ID <<<
              \multirow{18}*{Howard, MD} &  
                \multirow{18}*{I-70W} & 
                     \multirow{6}*{Recall} &
                        RF & 0.52 & 0.35 & 0.22 & 0.07 & 0.01 & - \\
                    &&& SVM & 0.94 & 0.84 & 0.80 & 0.70 & 0.61 & 0.52 \\
                    &&& GAN & 0.86 & 0.75 & 0.67 & 0.59 & 0.52 & 0.45 \\
                    &&& LightGBM & 0.84 & 0.84 & 0.77 & 0.68 & 0.59 & 0.58 \\
                    &&& Trans   & 0.90 & 0.77 & 0.67 & 0.58 & 0.51 & 0.43 \\
                    &&& G-Trans & 0.90 & 0.80 & 0.68 & 0.49 & 0.41 & 0.38 \\
                    \cline{3-10}
                && \multirow{6}*{Precision} &
                        RF & 0.77 & 0.65 & 0.60 & 0.36 & 0.14 & - \\ 
                    &&& SVM & 0.38 & 0.28 & 0.23 & 0.19 & 0.17 & 0.13 \\
                    &&& GAN & 0.60 & 0.46 & 0.36 & 0.31 & 0.24 & 0.20 \\
                    &&& LightGBM & 0.68 & 0.59 & 0.40 & 0.37 & 0.31 & 0.21 \\
                    &&& Trans   & 0.90 & 0.80 & 0.71 & 0.62 & 0.54 & 0.46 \\
                    &&& G-Trans & 0.81 & 0.79 & 0.68 & 0.60 & 0.51 & 0.48 \\
                    \cline{3-10}
                && \multirow{6}*{F1 Score} &
                     RF & 0.62 & 0.45 & 0.32 & 0.12 & 0.03 & - \\
                    &&& SVM & 0.54 & 0.42 & 0.36 & 0.30 & 0.27 & 0.21 \\
                    &&& GAN & 0.71 & 0.57 & 0.46 & 0.40 & 0.33 & 0.28 \\
                    &&& LightGBM & 0.75 & 0.69 & 0.53 & 0.48 & 0.41 & 0.30 \\
                    &&& Trans   & 0.90$^*$ & 0.79$^*$ & 0.69$^*$ & 0.60$^*$ & 0.52$^*$ & 0.45$^*$ \\
                    &&& G-Trans & 0.85$^\dagger$ & 0.79$^*$ & 0.68$^\dagger$  & 0.54$^\dagger$  & 0.45$^\dagger$  & 0.42$^\dagger$ \\
                \hline  % ID <<<

              \multirow{18}*{Howard, MD} & 
                \multirow{18}*{US-40E} & 
                     \multirow{6}*{Recall} &
                        RF & 0.31 & 0.21 & 0.13 & 0.10 & 0.06 & 0.05 \\
                    &&& SVM & 0.81 & 0.80 & 0.82 & 0.75 & 0.74 & 0.71 \\
                    &&& GAN & 0.57 & 0.49 & 0.43 & 0.37 & 0.33 & 0.30 \\
                    &&& LightGBM & 0.61 & 0.50 & 0.42 & 0.39 & 0.31 & 0.30 \\
                    &&& Trans   & 0.85 & 0.70 & 0.57 & 0.50 & 0.45 & 0.40 \\
                    &&& G-Trans & 0.85 & 0.70 & 0.57 & 0.50 & 0.45 & 0.40 \\
                    \cline{3-10}
                && \multirow{6}*{Precision} &
                       RF & 0.84 & 0.77 & 0.81 & 0.87 & 0.75 & 0.69 \\
                    &&& SVM & 0.17 & 0.11 & 0.09 & 0.10 & 0.08 & 0.07 \\
                    &&& GAN & 0.66 & 0.61 & 0.57 & 0.61 & 0.53 & 0.55 \\
                    &&& LightGBM & 0.55 & 0.46 & 0.38 & 0.36 & 0.41 & 0.40 \\
                    &&& Trans   & 0.85 & 0.71 & 0.58 & 0.51 & 0.45 & 0.41 \\
                    &&& G-Trans & 0.85 & 0.71 & 0.58 & 0.51 & 0.45 & 0.41 \\
                    \cline{3-10}
                && \multirow{6}*{F1 Score} &
                        RF & 0.45 & 0.33 & 0.23 & 0.19 & 0.12 & 0.09 \\
                    &&& SVM & 0.28 & 0.19 & 0.16 & 0.17 & 0.15 & 0.12 \\
                    &&& GAN & 0.61 & 0.55 & 0.49 & 0.46 & 0.41 & 0.39 \\
                    &&& LightGBM & 0.58 & 0.48 & 0.40 & 0.37 & 0.36 & 0.34 \\
                    &&& Trans   & 0.85$^*$ & 0.71$^*$ & 0.57$^*$ & 0.50$^*$ & 0.45$^*$ & 0.40$^*$ \\
                    &&& G-Trans & 0.84$^\dagger$ & 0.71$^*$ & 0.57$^*$ & 0.50$^*$ & 0.45$^*$ & 0.40$^*$ \\
                \hline  % ID <<<

              \multirow{18}*{Howard, MD} &
                \multirow{18}*{US-40W} & 
                     \multirow{6}*{Recall} &
                     RF & 0.43 & 0.35 & 0.20 & 0.17 & 0.07 & 0.05 \\
                    &&& SVM & 0.81 & 0.76 & 0.74 & 0.73 & 0.70 & 0.68 \\ 
                    &&& GAN & 0.62 & 0.58 & 0.52 & 0.50 & 0.47 & 0.45 \\
                    &&& LightGBM & 0.60 & 0.55 & 0.55 & 0.37 & 0.28 & 0.26 \\
                    &&& Trans   & 0.86 & 0.73 & 0.60 & 0.51 & 0.45 & 0.42 \\
                    &&& G-Trans & 0.86 & 0.73 & 0.60 & 0.51 & 0.45 & 0.42 \\
                    \cline{3-10}
                && \multirow{6}*{Precision} &
                       RF & 0.71 & 0.56 & 0.38 & 0.35 & 0.22 & 0.23 \\
                    &&& SVM & 0.26 & 0.19 & 0.14 & 0.11 & 0.10 & 0.16 \\
                    &&& GAN & 0.38 & 0.31 & 0.25 & 0.21 & 0.16 & 0.12 \\
                    &&& LightGBM & 0.68 & 0.40 & 0.33 & 0.29 & 0.33 & 0.31 \\
                    &&& Trans   & 0.82 & 0.73 & 0.61 & 0.53 & 0.47 & 0.44 \\
                    &&& G-Trans & 0.86 & 0.74 & 0.61 & 0.53 & 0.47 & 0.44 \\
                    \cline{3-10}
                && \multirow{6}*{F1 Score} &
                        RF & 0.53 & 0.43 & 0.26 & 0.22 & 0.10 & 0.08 \\      
                    &&& SVM & 0.40 & 0.30 & 0.23 & 0.19 & 0.17 & 0.09 \\
                    &&& GAN & 0.47 & 0.40 & 0.33 & 0.30 & 0.24 & 0.19 \\
                    &&& LightGBM & 0.64 & 0.46 & 0.41 & 0.33 & 0.30 & 0.28 \\
                    &&& Trans   & 0.84$^\dagger$ & 0.73$^*$ & 0.60$^*$ & 0.52$^*$ & 0.46$^*$ & 0.43$^*$ \\
                    &&& G-Trans & 0.86$^*$ & 0.73$^*$ & 0.60$^*$ & 0.52$^*$ & 0.46$^*$ & 0.43$^*$ \\
                \hline % ID <<<

              \multirow{18}*{Howard, MD} & 
                \multirow{18}*{I-695A} & 
                     \multirow{6}*{Recall} &
                     RF & 0.07 & 0.01 & 0.01 & 0.01 & - & - \\
                    &&& SVM & 0.91 & 0.85 & 0.83 & 0.78 & 0.74 & 0.66 \\
                    &&& GAN & 0.59 & 0.83 & 0.75 & 0.70 & 0.66 & 0.61 \\
                    &&& LightGBM & 0.75 & 0.63 & 0.48 & 0.42 & 0.32 & 0.60 \\
                    &&& Trans   & 0.87 & 0.75 & 0.65 & 0.60 & 0.55 & 0.50 \\
                    &&& G-Trans & 0.87 & 0.74 & 0.65 & 0.59 & 0.55 & 0.50 \\
                    \cline{3-10}
                && \multirow{6}*{Precision} &
                    RF & 0.89 & 1.00 & 1.00 & 1.00 & - & - \\
                    &&& SVM & 0.26 & 0.23 & 0.21 & 0.19 & 0.18 & 0.16 \\
                    &&& GAN & 0.44 & 0.40 & 0.38 & 0.38 & 0.37 & 0.36 \\
                    &&& LightGBM & 0.77 & 0.75 & 0.49 & 0.51 & 0.59 & 0.36 \\
                    &&& Trans   & 0.87 & 0.76 & 0.66 & 0.62 & 0.56 & 0.51 \\
                    &&& G-Trans & 0.88 & 0.76 & 0.67 & 0.61 & 0.57 & 0.52 \\
                    \cline{3-10}
                && \multirow{6}*{F1 Score} &
                    RF & 0.13 & 0.02 & 0.02 & 0.02 & - & - \\
                    &&& SVM & 0.40 & 0.36 & 0.34 & 0.31 & 0.29 & 0.26 \\
                    &&& GAN & 0.59 & 0.54 & 0.50 & 0.49 & 0.48 & 0.45 \\
                    &&& LightGBM & 0.76 & 0.68 & 0.49 & 0.46 & 0.41 & 0.45 \\
                    &&& Trans   & 0.87$^\dagger$ & 0.76$^*$ & 0.66$^*$ & 0.61$^*$ & 0.55$^\dagger$ & 0.51$^*$ \\
                    &&& G-Trans & 0.88$^*$ & 0.75$^\dagger$ & 0.66$^*$ & 0.60$^\dagger$ & 0.56$^*$ & 0.51$^*$ \\
                \hline  % ID <<<
		\end{tabular}
	\end{center}
 \caption{Howard County Anomaly Detection/Prediction Results}\label{tab:T_Prediction}
 \label{tab: tsmo_a}
\end{table}

%% file: tables/cranberry_anomaly.tex
\begin{table}[!htb]
%\scriptsize
%\tiny
\fontsize{6}{6.8}\selectfont
	\begin{center}
		\begin{tabular}{c|c|c|c|c|c|c|c|c|c}
		    \hline 
                \multicolumn{2}{c|}{Location} & \multirow{2}*{Metrics} & \multirow{2}*{Model} & \multicolumn{6}{|c}{Time}\\
                \cline{1-2}\cline{5-10}
		      County & Road &  &  & 5min & 10min & 15min& 20min & 25min & 30min\\\hline % Title
              \multirow{18}*{Cranberry, PA} & 
                \multirow{18}*{I-79S} &
                     \multirow{6}*{Recall} &
                        RF & 0.36 & 0.26 & 0.24 & 0.19 & 0.20 & 0.18 \\
                    &&& SVM & 0.62 & 0.51 & 0.43 & 0.38 & 0.45 & 0.37 \\
                    &&& GAN & 0.43 & 0.36 & 0.31 & 0.29 & 0.27 & 0.25 \\
                    &&& LightGBM & 0.37 & 0.37 & 0.28 & 0.21 & 0.22 & 0.19 \\
                    &&& Trans & 0.80 & 0.64 & 0.35 & 0.28 & 0.26 & 0.27 \\
                    &&& G-Trans & 0.81 & 0.64 & 0.46 & 0.37 & 0.32 & 0.32 \\
                    \cline{3-10}
                && \multirow{6}*{Precision} &
                        RF & 0.76 & 0.67 & 0.65 & 0.52 & 0.53 & 0.52 \\
                    &&& SVM & 0.08 & 0.07 & 0.07 & 0.07 & 0.07 & 0.06 \\
                    &&& GAN & 0.39 & 0.44 & 0.44 & 0.42 & 0.41 & 0.44 \\
                    &&& LightGBM & 0.38 & 0.29 & 0.31 & 0.36 & 0.27 & 0.26 \\
                    &&& Trans & 0.79 & 0.62 & 0.46 & 0.39 & 0.36 & 0.38 \\
                    &&& G-Trans & 0.78 & 0.63 & 0.46 & 0.37 & 0.32 & 0.32 \\
                    \cline{3-10}
                && \multirow{6}*{F1 Score} &
                        RF & 0.44 & 0.38 & 0.35 & 0.28 & 0.29 & 0.27 \\
                    &&& SVM & 0.15 & 0.13 & 0.11 & 0.11 & 0.12 & 0.10 \\
                    &&& GAN & 0.41 & 0.40 & 0.36 & 0.35$^\dagger$ & 0.32$^*$ & 0.32$^*$ \\
                    &&& LightGBM & 0.38 & 0.33 & 0.30 & 0.27 & 0.24 & 0.22 \\
                    &&& Trans & 0.79$^\dagger$ & 0.58$^\dagger$ & 0.40$^\dagger$ & 0.32 & 0.30 & 0.31 \\
                    &&& G-Trans & 0.80$^*$ & 0.63$^*$ & 0.46$^*$ & 0.37$^*$ & 0.32$^*$ & 0.32$^*$\\
                \hline  % ID <<<

              \multirow{18}*{Cranberry, PA} & 
                \multirow{18}*{I-79N} &
                     \multirow{6}*{Recall} &
                        RF & 0.19 & 0.11 & 0.10 & 0.09 & 0.09 & 0.11 \\
                    &&& SVM & 0.71 & 0.62 & 0.59 & 0.58 & 0.61 & 0.57 \\
                    &&& GAN & 0.52 & 0.39 & 0.33 & 0.29 & 0.27 & 0.23 \\
                    &&& LightGBM & 0.36 & 0.31 & 0.38 & 0.37 & 0.22 & 0.25 \\
                    &&& Trans   & 0.77 & 0.49 & 0.32 & 0.23 & 0.16 & 0.13 \\
                    &&& G-Trans & 0.78 & 0.58 & 0.38 & 0.28 & 0.22 & 0.18 \\
                    \cline{3-10}
                && \multirow{6}*{Precision} &
                        RF & 0.44 & 0.31 & 0.28 & 0.29 & 0.32 & 0.38 \\
                    &&& SVM & 0.15 & 0.15 & 0.13 & 0.13 & 0.12 & 0.11 \\
                    &&& GAN & 0.28 & 0.25 & 0.22 & 0.22 & 0.21 & 0.20 \\
                    &&& LightGBM & 0.35 & 0.22 & 0.15 & 0.18 & 0.17 & 0.18 \\
                    &&& Trans   & 0.64 & 0.41 & 0.29 & 0.22 & 0.15 & 0.13 \\
                    &&& G-Trans & 0.78 & 0.58 & 0.38 & 0.28 & 0.22 & 0.18 \\
                    \cline{3-10}
                && \multirow{6}*{F1 Score} &
                        RF & 0.26 & 0.16 & 0.15 & 0.14 & 0.14 & 0.17 \\
                    &&& SVM & 0.25 & 0.25 & 0.22 & 0.21 & 0.21 & 0.19 \\
                    &&& GAN & 0.37 & 0.31 & 0.26 & 0.25$^\dagger$ & 0.23$^*$ & 0.21$^*$ \\
                    &&& LightGBM & 0.36 & 0.25 & 0.22 & 0.24 & 0.19 & 0.21$^*$ \\
                    &&& Trans   & 0.70$^\dagger$ & 0.45$^\dagger$ & 0.30$^\dagger$ & 0.22 & 0.16 & 0.13 \\
                    &&& G-Trans & 0.78$^*$ & 0.58$^*$ & 0.38$^*$ & 0.28$^*$ & 0.22$^\dagger$ & 0.18 \\
                \hline  % ID <<<

              \multirow{18}*{Cranberry, PA} & 
                \multirow{18}*{I-76W} & 
                     \multirow{6}*{Recall} &
                        RF & 0.01 & - & - & - & - & - \\
                    &&& SVM & 0.66 & 0.50 & 0.40 & 0.36 & 0.31 & 0.36 \\
                    &&& GAN & 0.53 & 0.38 & 0.29 & 0.29 & 0.30 & 0.29 \\
                    &&& LightGBM & 0.45 & 0.36 & 0.23 & 0.34 & 0.25 & 0.25 \\
                    &&& Trans   & 0.58 & 0.45 & 0.31 & 0.27 & 0.23 & 0.20 \\
                    &&& G-Trans & 0.75 & 0.52 & 0.30 & 0.23 & 0.19 & 0.15 \\
                    \cline{3-10}
                && \multirow{6}*{Precision} &
                        RF & 1.00 & - & - & - & - & - \\
                    &&& SVM & 0.08 & 0.06 & 0.05 & 0.05 & 0.04 & 0.05 \\
                    &&& GAN & 0.16 & 0.13 & 0.12 & 0.13 & 0.13 & 0.13 \\
                    &&& LightGBM & 0.16 & 0.12 & 0.18 & 0.07 & 0.13 & 0.17 \\
                    &&& Trans   & 0.72 & 0.41 & 0.27 & 0.19 & 0.15 & 0.12 \\
                    &&& G-Trans & 0.76 & 0.53 & 0.31 & 0.24 & 0.19 & 0.15 \\
                    \cline{3-10}
                && \multirow{6}*{F1 Score} &
                        RF & 0.02 & - & - & - & - & - \\
                    &&& SVM & 0.14 & 0.10 & 0.10 & 0.08 & 0.07 & 0.08 \\
                    &&& GAN & 0.25 & 0.20 & 0.17 & 0.18 & 0.18$^\dagger$ & 0.18$^\dagger$ \\
                    &&& LightGBM & 0.23 & 0.18 & 0.20 & 0.12 & 0.17 & 0.20$^*$ \\
                    &&& Trans   & 0.65$^\dagger$ & 0.43$^\dagger$ & 0.29$^\dagger$ & 0.23$^\dagger$ & 0.18$^\dagger$ & 0.15 \\
                    &&& G-Trans & 0.76$^*$ & 0.52$^*$ & 0.30$^*$ & 0.24$^*$ & 0.19$^*$ & 0.15 \\
                \hline  % ID <<<

              \multirow{18}*{Cranberry, PA} &
                \multirow{18}*{I-76E} & 
                     \multirow{6}*{Recall} &
                        RF & 0.05 & 0.03 & 0.07 & 0.10 & 0.10 & 0.10 \\
                    &&& SVM & 0.42 & 0.35 & 0.33 & 0.33 & 0.33 & 0.37 \\
                    &&& GAN & 0.34 & 0.23 & 0.20 & 0.23 & 0.27 & 0.27 \\
                    &&& LightGBM & 0.54 & 0.42 & 0.38 & 0.32 & 0.30 & 0.33 \\
                    &&& Trans   & 0.74 & 0.44 & 0.15 & 0.08 & 0.08 & 0.08 \\
                    &&& G-Trans & 0.74 & 0.45 & 0.15 & 0.08 & 0.08 & 0.08 \\
                    \cline{3-10}
                && \multirow{6}*{Precision} &
                        RF & 0.21 & 0.12 & 0.22 & 0.50 & 0.38 & 0.40 \\
                    &&& SVM & 0.08 & 0.07 & 0.06 & 0.06 & 0.06 & 0.07 \\
                    &&& GAN & 0.14 & 0.10 & 0.09 & 0.10 & 0.12 & 0.12 \\
                    &&& LightGBM & 0.10 & 0.08 & 0.07 & 0.07 & 0.06 & 0.07 \\
                    &&& Trans   & 0.71 & 0.43 & 0.15 & 0.08 & 0.09 & 0.09 \\
                    &&& G-Trans & 0.72 & 0.41 & 0.14 & 0.08 & 0.08 & 0.08 \\
                    \cline{3-10}
                && \multirow{6}*{F1 Score} &
                        RF & 0.08 & 0.05 & 0.10 & 0.17$^*$ & 0.16$^*$ & 0.16$^*$ \\
                    &&& SVM & 0.14 & 0.11 & 0.11 & 0.10 & 0.10 & 0.10 \\
                    &&& GAN & 0.20 & 0.13 & 0.12 & 0.14$^\dagger$ & 0.16$^*$ & 0.16$^*$ \\
                    &&& LightGBM & 0.16 & 0.13 & 0.11 & 0.11 & 0.10 & 0.11 \\
                    &&& Trans   & 0.72$^*$ & 0.43$^*$ & 0.15$^*$ & 0.08 & 0.09 & 0.09 \\
                    &&& G-Trans & 0.72$^*$ & 0.43$^*$ & 0.14$^\dagger$ & 0.08 & 0.08 & 0.08 \\
                \hline % ID <<<

              \multirow{18}*{Cranberry, PA} & 
                \multirow{18}*{US-19} & 
                     \multirow{6}*{Recall} &
                        RF & - & - & - & - & - & - \\
                    &&& SVM & 0.47 & 0.36 & 0.39 & 0.39 & 0.45 & 0.51 \\
                    &&& GAN & 0.27 & 0.24 & 0.22 & 0.21 & 0.24 & 0.19 \\
                    &&& LightGBM & 0 .11& 0.16 & 0.13 & 0.11 & 0.13 & 0.09 \\
                    &&& Trans   & 0.74 & 0.35 & 0.24 & 0.21 & 0.18 & 0.16 \\
                    &&& G-Trans & 0.77 & 0.54 & 0.31 & 0.22 & 0.19 & 0.16 \\
                    \cline{3-10}
                && \multirow{6}*{Precision} &
                        RF & - & - & - & - & - & - \\
                    &&& SVM & 0.04 & 0.03 & 0.03 & 0.03 & 0.04 & 0.04 \\
                    &&& GAN & 0.06 & 0.06 & 0.05 & 0.05 & 0.05 & 0.04 \\
                    &&& LightGBM & 0.08 & 0.09 & 0.08 & 0.06 & 0.06 & 0.04 \\
                    &&& Trans   & 0.76 & 0.61 & 0.47 & 0.44 & 0.40 & 0.35 \\
                    &&& G-Trans & 0.76 & 0.52 & 0.29 & 0.21 & 0.17 & 0.14 \\
                    \cline{3-10}
                && \multirow{6}*{F1 Score} &
                        RF & - & - & - & - & - & - \\
                    &&& SVM & 0.08 & 0.06 & 0.06  & 0.06  & 0.07 & 0.08 \\
                    &&& GAN & 0.10 & 0.09 & 0.08 & 0.08 & 0.07 & 0.07 \\
                    &&& LightGBM & 0.09 & 0.11 & 0.10 & 0.08 & 0.08 & 0.06 \\
                    &&& Trans   & 0.75$^\dagger$ & 0.45$^\dagger$ & 0.31$^*$ & 0.28$^*$ & 0.25$^*$ & 0.22$^*$ \\
                    &&& G-Trans & 0.77$^*$ & 0.53$^*$ & 0.30$^\dagger$ & 0.21$^\dagger$ & 0.18$^\dagger$ & 0.15$^\dagger$ \\
                \hline % ID <<<
		\end{tabular}
	\end{center}
 \caption{Cranberry Township Anomaly Detection/Prediction Results}\label{tab:C_Prediction}
 \label{tab: Cranbeery_a}
\end{table}

%% file: tables/incident.tex
\begin{table}[!htb]
%\scriptsize
%\tiny
\fontsize{6}{7.2}\selectfont
	\begin{center}
		\begin{tabular}{c|c|c|c|c|c|c|c}
		    \hline 
                \multicolumn{2}{c|}{Location}  & \multirow{2}*{Model} & \multirow{2}*{DR}& \multirow{2}*{MTTD} &\multirow{2}*{FAR} &\multirow{2}*{DR(S)} &\multirow{2}*{MTTD(S)}\\
                \cline{1-2}
		      County & Road & &  & & &  & \\\hline % Title
            \multirow{7}*{Cranberry, PA} & 
                \multirow{7}*{I-79S} &
                     OOD & \cancel{0.14}  & \st{0} & \cancel{\textcolor{red}{0.99}} & \cancel{0.00}  & - \\
                     && {CFOD} & \cancel{\textcolor{red}{0.00}}  & - & \cancel{0.63}       & \cancel{\textcolor{red}{0.00}}  & -\\ 
                    && SVM & \cancel{0.00} &  - & \cancel{\textcolor{red}{0.95}} & \cancel{0.00}  & -\\ 
                    && GAN & \cancel{\textcolor{red}{0.00}}  & - & \st{0} & \cancel{\textcolor{red}{0.00}}  & -\\ 
                    && {TSSAE} & \cancel{\textcolor{red}{0.00}}  &  - & \cancel{\textcolor{red}{1.00}} &  \cancel{\textcolor{red}{0.00}}  & -\\ 
                    && Ours & 0.50  & -1 & 0.08 & 1.00  & -1   \\
                    && Ours (a) & 0.50  & -12  & 0.00 & 1.00 & -12 \\
                \hline  % ID <<<

            \multirow{7}*{Cranberry, PA} &
                \multirow{7}*{I-79N} & 
                     OOD & \cancel{0.05}  & \cancel{27} & \cancel{\textcolor{red}{0.94}}& \cancel{0.09}  & \cancel{27}\\
                      && {CFOD} & \cancel{\textcolor{red}{0.00}}  & - & \cancel{\textcolor{red}{0.91}} & \cancel{\textcolor{red}{0.00}}  & -\\ 
                    && SVM & \cancel{0.21}  & \cancel{-14} & \cancel{\textcolor{red}{0.94}}& \cancel{0.18}  & \cancel{-30}\\
                    && GAN & \cancel{\textcolor{red}{0.00}}  & - & \st{0} & \cancel{\textcolor{red}{0.00}}  & -\\ 
                      && {TSSAE} & \cancel{0.16}  &  \st{2} & \cancel{\textcolor{red}{0.98}} &  \cancel{0.00}  & -\\ 
                    && Ours & 0.58  & -3 & 0.00& 1.00  & -3   \\
                    && Ours (a) & 0.58 & -14 & 0.13& 1.00  & -14 \\
                \hline  % ID <<<

            \multirow{7}*{Cranberry, PA} & 
                \multirow{7}*{I-76W} & 
                     OOD & \cancel{0.33}  & \cancel{14} & \cancel{\textcolor{red}{0.99}} & \cancel{0.33}  & \cancel{14} \\
                      && {CFOD} & \cancel{\textcolor{red}{0.00}}  & - & \cancel{\textcolor{red}{0.86}} & \cancel{\textcolor{red}{0.00}}  & -\\ 
                    && SVM & \cancel{0.33}  & \cancel{-76} & \cancel{\textcolor{red}{0.92}} & \cancel{0.33}  & \cancel{-76}\\
                    && GAN & 0.33  &  -76 & 0.77 & 0.33  & -76\\
                     && {TSSAE} & \cancel{\textcolor{red}{0.00}}  &  - & \cancel{\textcolor{red}{1.00}} &  \cancel{\textcolor{red}{0.00}}  & -\\ 
                    && Ours & 1.00  & -2 & 0.00 & 1.00  & -2   \\
                    && Ours (a) & 1.00 & -17 & 0.10 & 1.00  & -17 \\
                \hline  % ID <<<

            \multirow{7}*{Cranberry, PA} & 
                \multirow{7}*{I-76E} & 
                     OOD & \cancel{0.60}  & \cancel{-30} & \cancel{\textcolor{red}{0.94}} & \cancel{0.75}  & \cancel{-30}\\
                      && {CFOD} & \cancel{0.20}  & \cancel{43} & \cancel{\textcolor{red}{0.97}} & \cancel{0.25}  & \cancel{43}\\ 
                    && SVM & \cancel{0.80} & \cancel{-27} & \cancel{\textcolor{red}{0.89}} & \cancel{1.00}  & \cancel{-27}\\
                    && GAN & 0.60  & -58  & 0.76 & 0.75 & -58\\
                     && {TSSAE} & \cancel{0.20}  &  \cancel{23} & \cancel{\textcolor{red}{0.98}} &  \cancel{0.25}  & \cancel{23}\\ 
                    && Ours & 0.80  & -2 & 0.00 & 1.00  & -2   \\
                    && Ours (a) & 0.80  & -18 & 0.10 & 1.00  & -18 \\
                \hline  % ID <<<
            
            \multirow{7}*{Cranberry, PA} & %
                \multirow{7}*{US-19} &
                     OOD & \cancel{0.25}  & \st{7} & \cancel{\textcolor{red}{0.98}} & \cancel{0.25}  & \st{0} \\
                     && {CFOD} & \cancel{\textcolor{red}{0.00}}  & - & \cancel{\textcolor{red}{0.95}} & \cancel{\textcolor{red}{0.00}}  & -\\ 
                    && SVM & \cancel{0.78}  & \cancel{-5} & \cancel{\textcolor{red}{0.95}} & \cancel{1.00}  & \cancel{-5}\\
                    && GAN & \cancel{0.56} & \cancel{-10} & \cancel{\textcolor{red}{0.92}} & \cancel{0.75} & \cancel{-5}\\
                    && {TSSAE} & \cancel{0.11}  &  \st{0} & \cancel{\textcolor{red}{0.97}} &  \cancel{0.00}  & -\\ 
                    && Ours & 0.44  & -3 & 0.00 & 1.00  & -3   \\
                    && Ours (a) & 0.44  & -17 & 0.08  & 0.80 & -17 \\
                \hline  % ID <<<
            
            \multirow{7}*{Howard, MD} & 
                \multirow{7}*{I-70E} & 
                     OOD &  \cancel{0.00}  & - & \cancel{\textcolor{red}{1.00}} & \cancel{0.00}  & - \\
                      && {CFOD} & 0.23  &  -11 & 0.67 & 0.50  & -11\\ 
                       %&& \textcolor{blue}{RL} & \cancel{0.86}  & - & \cancel{0.63}       & \cancel{\textcolor{red}{0.00}}  & -\\ 
                    && SVM & \cancel{0.69}  & \st{4} & \cancel{\textcolor{red}{0.92}} &  \cancel{0.13}  & \st{4} \\
                    && GAN & 0.61  & 1 & 0.10 & 1.00  & 2 \\
                     && {TSSAE} & \cancel{0.08}  &  \cancel{-27} & \cancel{\textcolor{red}{0.86}} & \cancel{0.00}  & -\\ 
                    && Ours & 0.46 & -7 & 0.00 & 1.00  & -7 \\
                    && Ours (a) & 0.46 & -12 & 0.23 & 1.00  & -12 \\
                \hline  % ID <<<
            \multirow{7}*{Howard, MD} &
                \multirow{7}*{I-70W} & 
                     OOD & \cancel{0.14} & \cancel{-8} & \cancel{\textcolor{red}{0.95}} & \cancel{0.25}  & \cancel{-8} \\
                      && {CFOD} &  0.28  &  -27 & 0.67 & 0.5  & -27\\ 
                      %&& \textcolor{blue}{RL} & \cancel{0.86}  & - & \cancel{0.63}       & \cancel{\textcolor{red}{0.00}}  & -\\ 
                    && SVM & \cancel{0.71} & \cancel{-33} & \cancel{\textcolor{red}{0.95}} & \cancel{1.00}  & \cancel{-25} \\
                    && GAN & 0.43 & -14 & 0.2 & 0.75  & -14\\
                     && {TSSAE} & \cancel{0.14}  &  \cancel{-1} & \cancel{\textcolor{red}{0.99}} & \cancel{0.00}  & -\\ 
                    && Ours & 0.57 & -28 & 0.54 & 0.75  & -6 \\
                    && Ours (a) & 0.28 & -8 & 0.13 & 0.5  & -8 \\
                \hline  % ID <<<

            \multirow{7}*{Howard, MD} &
                \multirow{7}*{US-40E} & 
                     OOD &   \cancel{0.00}  & - & \cancel{\textcolor{red}{1.00}} & \cancel{0.00}  & - \\
                     && {CFOD} &  0.33  &  6 & 0.33 & 0.40  & 6\\ 
                    && SVM & \cancel{1.00}   & \cancel{-19} & \cancel{\textcolor{red}{0.94}} & \cancel{1.00}  & \cancel{-27}\\
                    && GAN & 0.67  & 2 & 0.00 & 0.80  & 2 \\
                     && {TSSAE} & \cancel{\textcolor{red}{0.00}}  &  - & \st{0} & \cancel{\textcolor{red}{0.00}}  & -\\ 
                    && Ours & 0.83  & -17 & 0.00 & 1.00  & -17 \\
                    && Ours (a) & 0.83  & -42 & 0.00 & 1.00  & -42 \\
                \hline  % ID <<<

            \multirow{7}*{Howard, MD} & 
                \multirow{7}*{US-40W} & 
                     OOD & 0.33  & -7 & 0.40 & 0.40  & -4 \\
                     && {CFOD} &  0.22  &  -2 & 0.67 & 0  & -\\ 
                    && SVM & \cancel{0.67}  & \st{0} & \cancel{\textcolor{red}{0.88}} & \cancel{0.80}  & \st{0}\\
                    && GAN & 0.44 & 8 & 0.00 & 0.60  & 11\\
                     && {TSSAE} & \cancel{0.22}  &  \st{1} & \cancel{\textcolor{red}{0.99}}  & \cancel{\textcolor{red}{0.00}}  & -\\ 
                    && Ours & 0.56  & -5 & 0.00 & 1.00  & -5 \\
                    && Ours (a) & 0.44  & -4 & 0.35 & 0.80  & -4 \\
                \hline  % ID <<<

            \multirow{7}*{Howard, MD} & 
                \multirow{7}*{I-695A} & 
                     OOD & \cancel{0.00}  & - & \cancel{\textcolor{red}{0.96}} & \cancel{0.00}  & - \\
                     && {CFOD} &  0.22  &  0 & 0 & 0.50  & 0\\ 
                    && SVM & \cancel{0.67}  & \cancel{-15} & \cancel{\textcolor{red}{0.92}} & \cancel{1.00}  & \st{2} \\
                    && GAN & 0.67 & -2 & 0.47 & 1.00  & 12 \\
                    && {TSSAE} &  \cancel{0.44}  &  \st{0} & \cancel{\textcolor{red}{0.99}} & \cancel{0.00}  & -\\ 
                    && Ours & 0.22  & -8 & 0.00 & 1.00  & -8 \\
                    && Ours (a) & 0.22  & -23 & 0.00 & 1.00  & -23 \\
                \hline  % ID <<<

		\end{tabular}
	\end{center}
 \caption{Overall performance of early anomaly detection/prediction}\label{tab:Incident}
\end{table}

%% file: tables/senstivity.tex
\begin{table}[!htb]
%\scriptsize
%\tiny
\fontsize{6}{7.2}\selectfont
	\begin{center}
		\begin{tabular}{c|c|c|c|c|c|c}
		    \hline 
                \multirow{2}*{$\theta_{\text{SD}}^{\text{I-70E}}$ (mph)}  & \multirow{2}*{Ahead-Labeling} & \multirow{2}*{DR}& \multirow{2}*{MTTD} &\multirow{2}*{FAR} &\multirow{2}*{DR(S)} &\multirow{2}*{MTTD(S)}\\
		        & &  & & &  & \\\hline % Title
                      \multirow{2}*{30} & 
                      \multirow{2}*{\xmark} & \multirow{2}*{0.46} &  \multirow{2}*{-7} & \multirow{2}*{0.00} & \multirow{2}*{1.00}  & \multirow{2}*{-7} \\ 
                      & &  & & &  & \\
                     \multirow{2}*{30} &  \multirow{2}*{\cmark} & \multirow{2}*{0.46} &  \multirow{2}*{-12} & \multirow{2}*{0.23} & \multirow{2}*{1.00}  & \multirow{2}*{-12} \\
                    & &  & & &  & \\
                    \multirow{2}*{25} &  \multirow{2}*{\xmark} & \multirow{2}*{0.61}  & \multirow{2}*{-7} & \multirow{2}*{0.15} & \multirow{2}*{1.00}  & \multirow{2}*{-7}   \\
                    & &  & & &  & \\
                    \multirow{2}*{25} &  \multirow{2}*{\cmark} & \multirow{2}*{0.61}  & \multirow{2}*{-9}  & \multirow{2}*{0.20  } & \multirow{2}*{1.00} & \multirow{2}*{-9} \\
                    & &  & & &  & \\
                    \multirow{2}*{20} & 
                      \multirow{2}*{\xmark} & \multirow{2}*{0.46} &  \multirow{2}*{-6} & \multirow{2}*{0.11} & \multirow{2}*{0.75}  & \multirow{2}*{-6} \\ 
                      & &  & & &  & \\
                    \multirow{2}*{20} & 
                      \multirow{2}*{\cmark} & \multirow{2}*{0.92} &  \multirow{2}*{-17} & \multirow{2}*{0.34} & \multirow{2}*{1.00}  & \multirow{2}*{-24} \\ 
                      & &  & & &  & \\
                \hline  % ID <<<
		\end{tabular}
	\end{center}
 \caption{Sensitivity Analysis}\label{tab:sen}
\end{table}

%% file: images/detection_example_1.tex
\begin{figure}[!htb]
    \centering
    \captionsetup[subfigure]{labelformat=empty}
    \begin{subfigure}[t]{0.45\textwidth}
        \centering
        \includegraphics[width=\textwidth]{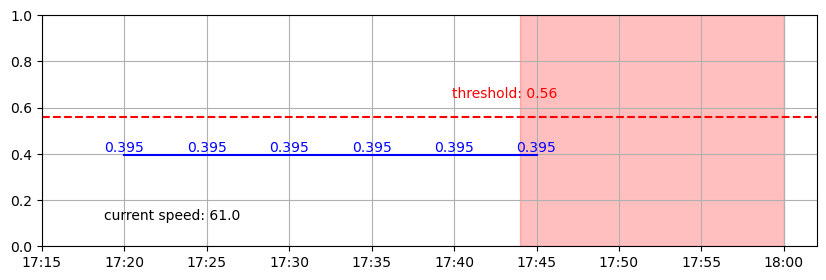}
        \caption{Prediction at 2023-01-11 17:15}
    \end{subfigure}
    \hfill
    \begin{subfigure}[t]{0.45\textwidth}
        \centering
        \includegraphics[width=\textwidth]{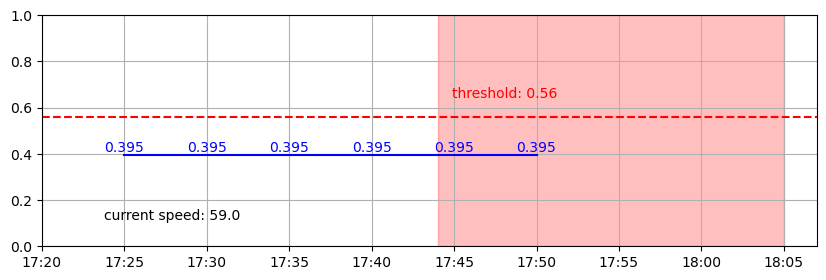}
        \caption{Prediction at 2023-01-11 17:20}
    \end{subfigure}
\end{figure}

\begin{figure}[!htb]
    \centering
    \captionsetup[subfigure]{labelformat=empty}
    \begin{subfigure}[t]{0.45\textwidth}
        \centering
        \includegraphics[width=\textwidth]{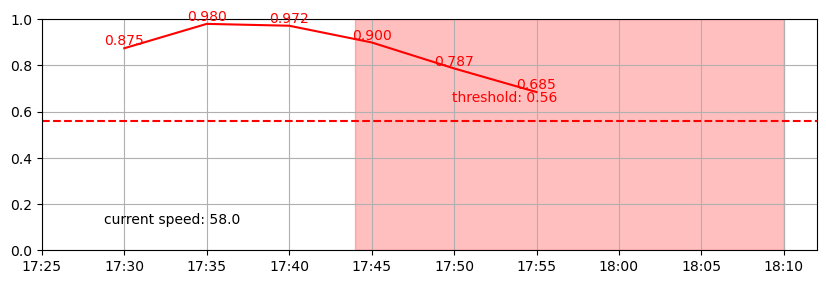}
        \caption{Prediction at 2023-01-11 17:25}
    \end{subfigure}
    \hfill
    \begin{subfigure}[t]{0.45\textwidth}
        \centering
        \includegraphics[width=\textwidth]{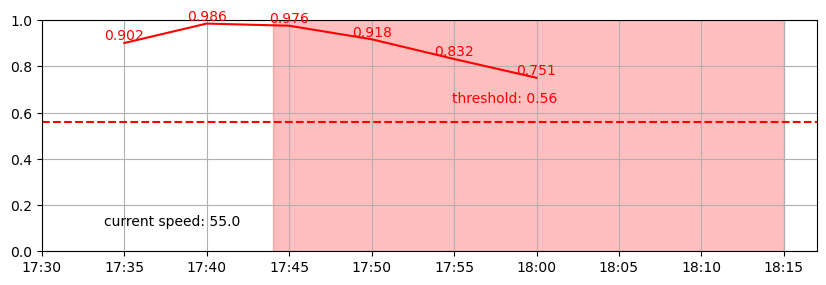}
        \caption{Prediction at 2023-01-11 17:30}
    \end{subfigure}
\end{figure}

\begin{figure}[!htb]
    \centering
    \captionsetup[subfigure]{labelformat=empty}
    \begin{subfigure}[t]{0.45\textwidth}
        \centering
        \includegraphics[width=\textwidth]{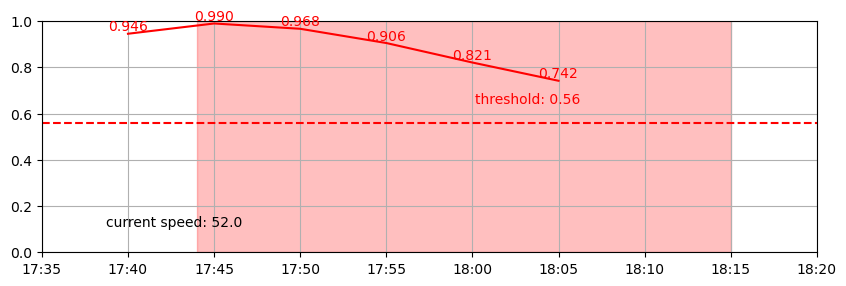}
        \caption{Prediction at 2023-01-11 17:35}
    \end{subfigure}
    \hfill
    \begin{subfigure}[t]{0.45\textwidth}
        \centering
        \includegraphics[width=\textwidth]{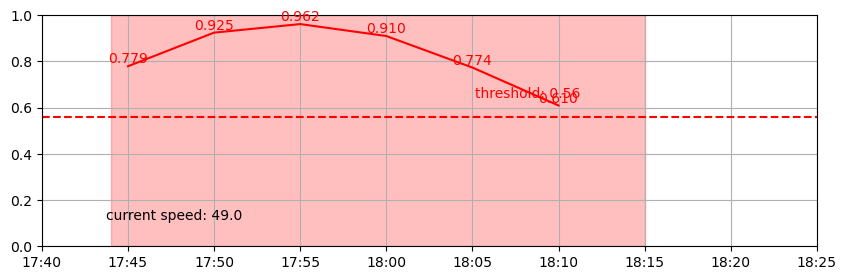}
        \caption{Prediction at 2023-01-11 17:40}
    \end{subfigure}
\end{figure}

\begin{figure}[!htb]
    \centering
    \captionsetup[subfigure]{labelformat=empty}
    \begin{subfigure}[t]{0.45\textwidth}
        \centering
        \includegraphics[width=\textwidth]{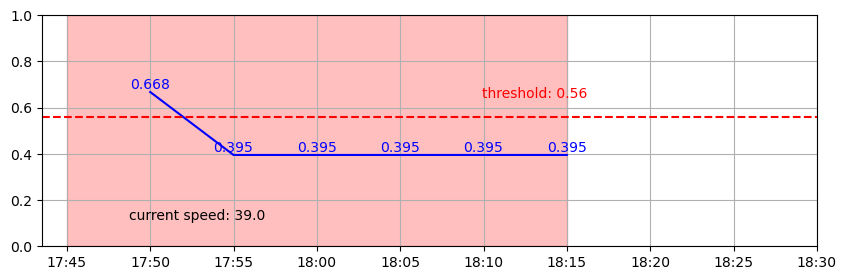}
        \caption{Prediction at 2023-01-11 17:45}
    \end{subfigure}
    \hfill
    \begin{subfigure}[t]{0.45\textwidth}
        \centering
        \includegraphics[width=\textwidth]{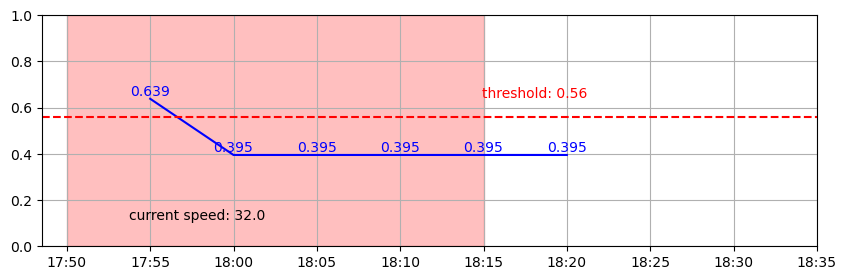}
        \caption{Prediction at 2023-01-11 17:50}
    \end{subfigure}
\end{figure}

%% file: images/detection_example_2.tex
\begin{figure}[H]
    \centering
    \captionsetup[subfigure]{labelformat=empty}
    \begin{subfigure}[t]{0.45\textwidth}
        \centering
        \includegraphics[width=\textwidth]{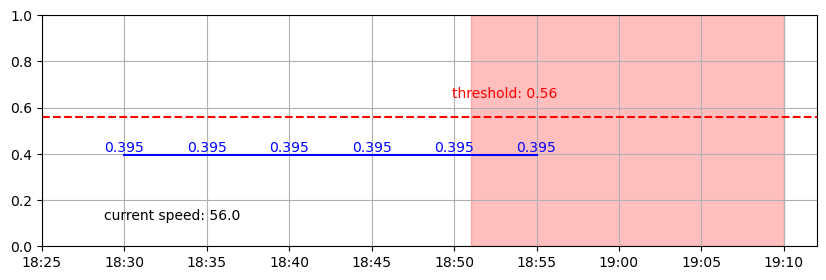}
        \caption{Prediction at 2023-01-25 18:25}
    \end{subfigure}
    \hfill
    \begin{subfigure}[t]{0.45\textwidth}
        \centering
        \includegraphics[width=\textwidth]{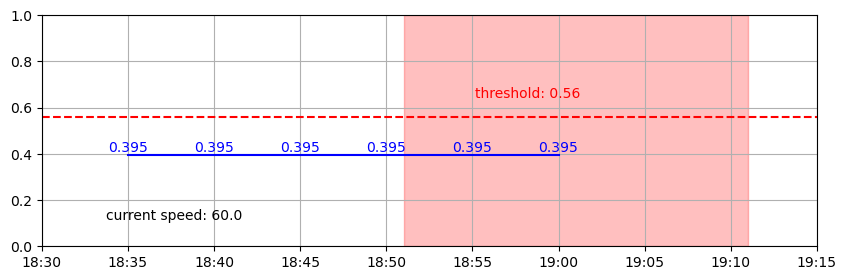}
        \caption{Prediction at 2023-01-25 18:30}
    \end{subfigure}
\end{figure}

\begin{figure}[H]
    \centering
    \captionsetup[subfigure]{labelformat=empty}
    \begin{subfigure}[t]{0.45\textwidth}
        \centering
        \includegraphics[width=\textwidth]{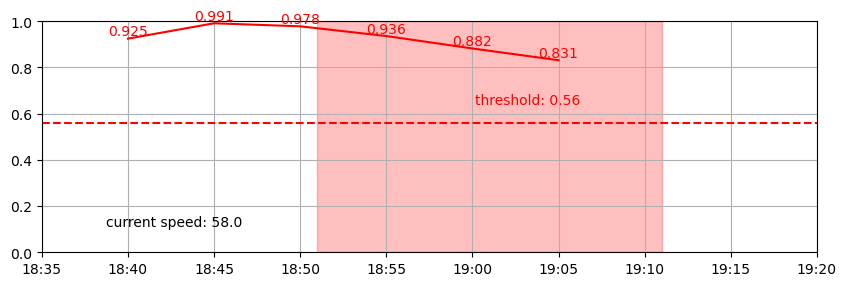}
        \caption{Prediction at 2023-01-25 18:35}
    \end{subfigure}
    \hfill
    \begin{subfigure}[t]{0.45\textwidth}
        \centering
        \includegraphics[width=\textwidth]{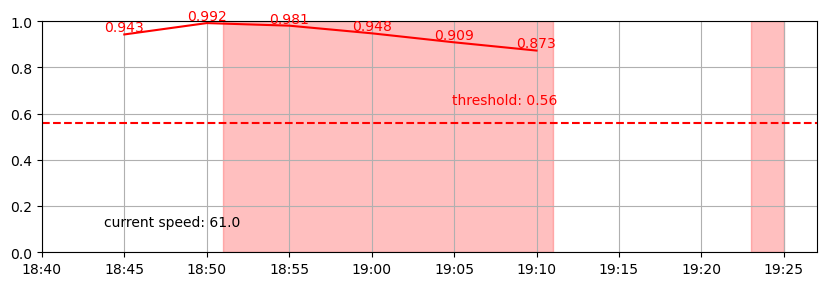}
        \caption{Prediction at 2023-01-25 18:40}
    \end{subfigure}
\end{figure}

\begin{figure}[H]
    \centering
    \captionsetup[subfigure]{labelformat=empty}
    \begin{subfigure}[t]{0.45\textwidth}
        \centering
        \includegraphics[width=\textwidth]{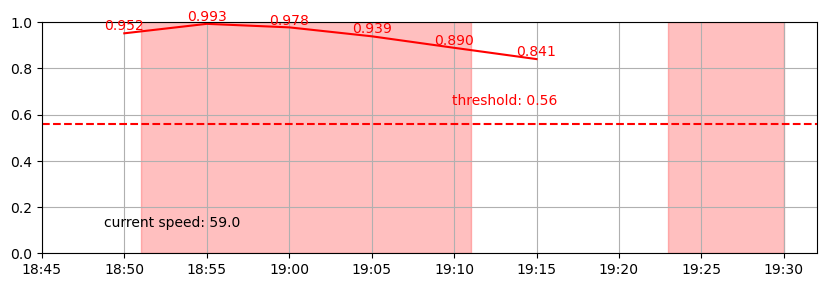}
        \caption{Prediction at 2023-01-25 18:45}
    \end{subfigure}
    \hfill
    \begin{subfigure}[t]{0.45\textwidth}
        \centering
        \includegraphics[width=\textwidth]{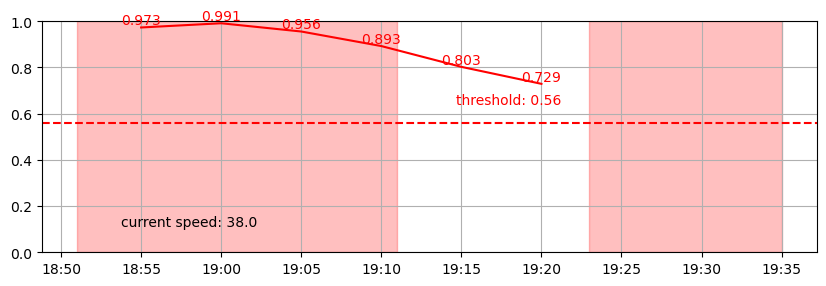}
        \caption{Prediction at 2023-01-25 18:50}
    \end{subfigure}
\end{figure}

\begin{figure}[H]
    \centering
    \captionsetup[subfigure]{labelformat=empty}
    \begin{subfigure}[t]{0.45\textwidth}
        \centering
        \includegraphics[width=\textwidth]{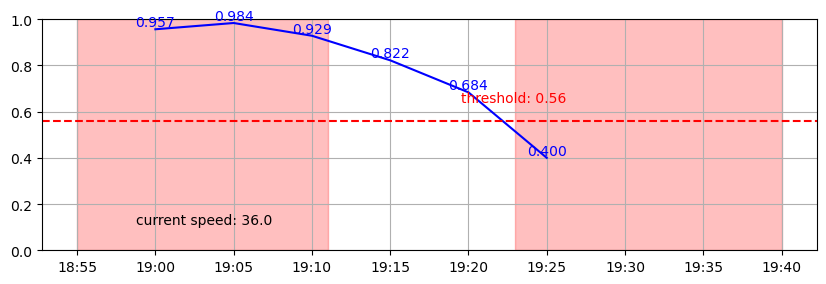}
        \caption{Prediction at 2023-01-25 18:55}
    \end{subfigure}
    \hfill
    \begin{subfigure}[t]{0.45\textwidth}
        \centering
        \includegraphics[width=\textwidth]{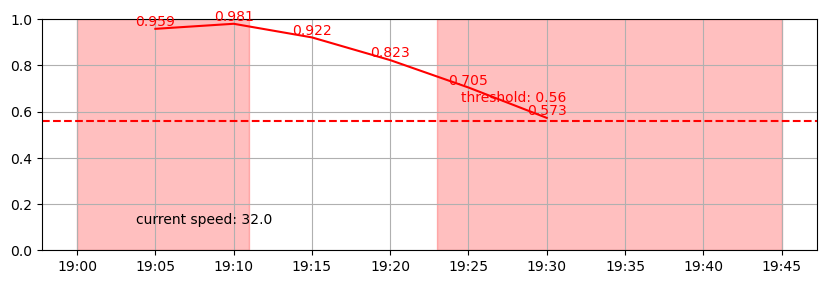}
        \caption{Prediction at 2023-01-25 19:00}
    \end{subfigure}
\end{figure}

\begin{figure}[H]
    \centering
    \captionsetup[subfigure]{labelformat=empty}
    \begin{subfigure}[t]{0.45\textwidth}
        \centering
        \includegraphics[width=\textwidth]{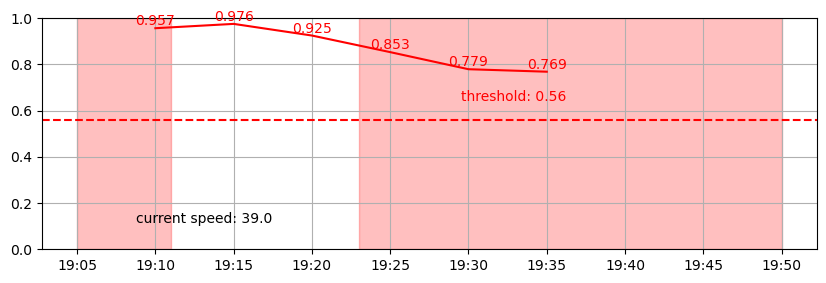}
        \caption{Prediction at 2023-01-25 19:05}
    \end{subfigure}
    \hfill
    \begin{subfigure}[t]{0.45\textwidth}
        \centering
        \includegraphics[width=\textwidth]{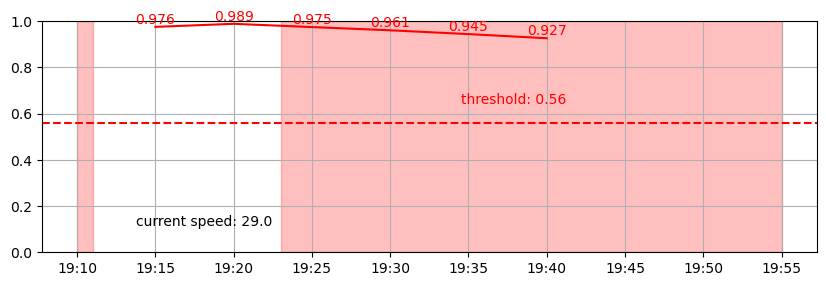}
        \caption{Prediction at 2023-01-25 19:10}
    \end{subfigure}
\end{figure}

\begin{figure}[H]
    \centering
    \captionsetup[subfigure]{labelformat=empty}
    \begin{subfigure}[t]{0.45\textwidth}
        \centering
        \includegraphics[width=\textwidth]{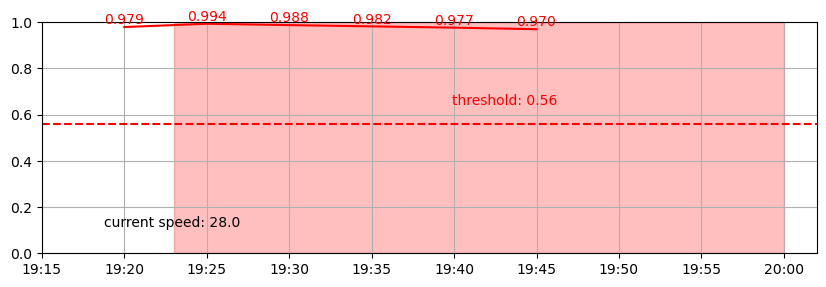}
        \caption{Prediction at 2023-01-25 19:15}
    \end{subfigure}
    \hfill
    \begin{subfigure}[t]{0.45\textwidth}
        \centering
        \includegraphics[width=\textwidth]{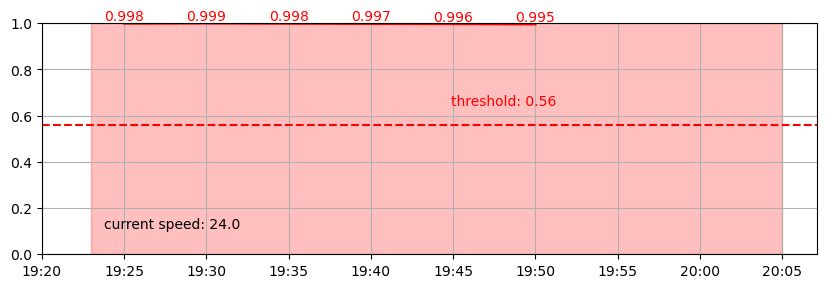}
        \caption{Prediction at 2023-01-25 19:20}
    \end{subfigure}
\end{figure}

\begin{figure}[H]
    \centering
    \captionsetup[subfigure]{labelformat=empty}
    \begin{subfigure}[t]{0.45\textwidth}
        \centering
        \includegraphics[width=\textwidth]{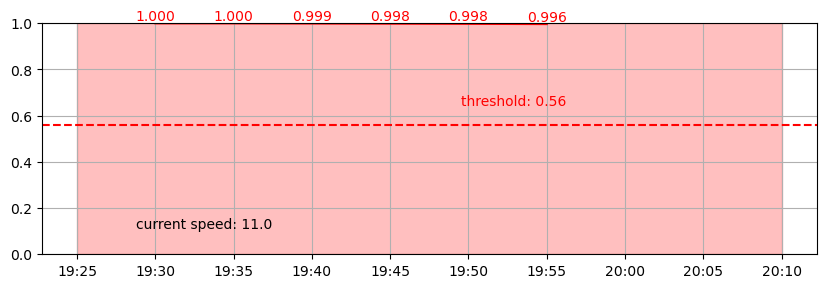}
        \caption{Prediction at 2023-01-25 19:25}
    \end{subfigure}
    \hfill
    \begin{subfigure}[t]{0.45\textwidth}
        \centering
        \includegraphics[width=\textwidth]{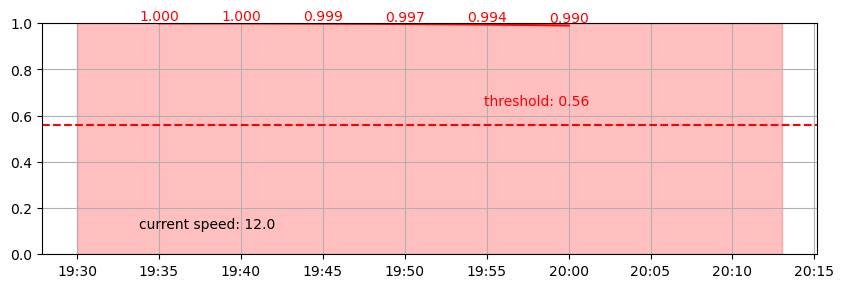}
        \caption{Prediction at 2023-01-25 19:30}
    \end{subfigure}
\end{figure}

\begin{figure}[H]
    \centering
    \captionsetup[subfigure]{labelformat=empty}
    \begin{subfigure}[t]{0.45\textwidth}
        \centering
        \includegraphics[width=\textwidth]{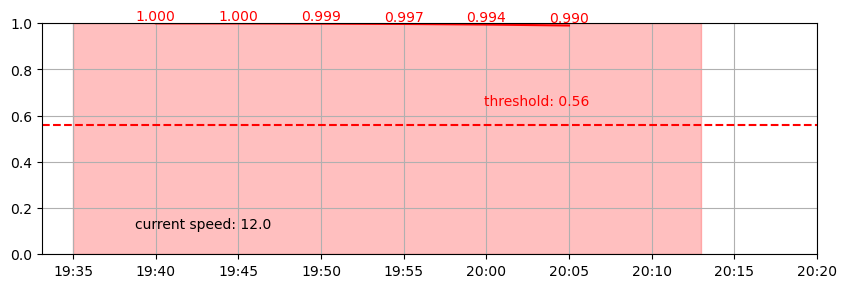}
        \caption{Prediction at 2023-01-25 19:35}
    \end{subfigure}
    \hfill
    \begin{subfigure}[t]{0.45\textwidth}
        \centering
        \includegraphics[width=\textwidth]{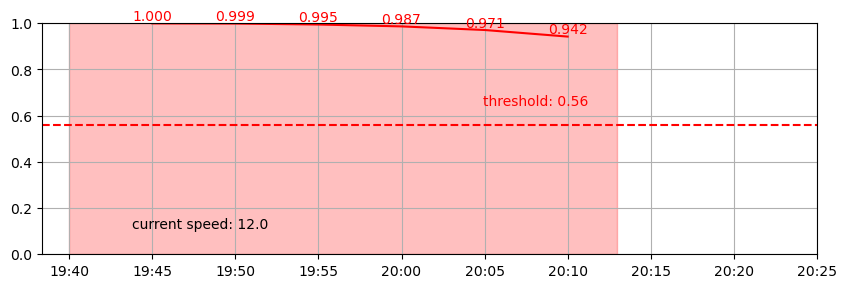}
        \caption{Prediction at 2023-01-25 19:40}
    \end{subfigure}
\end{figure}

\begin{figure}[H]
    \centering
    \captionsetup[subfigure]{labelformat=empty}
    \begin{subfigure}[t]{0.45\textwidth}
        \centering
        \includegraphics[width=\textwidth]{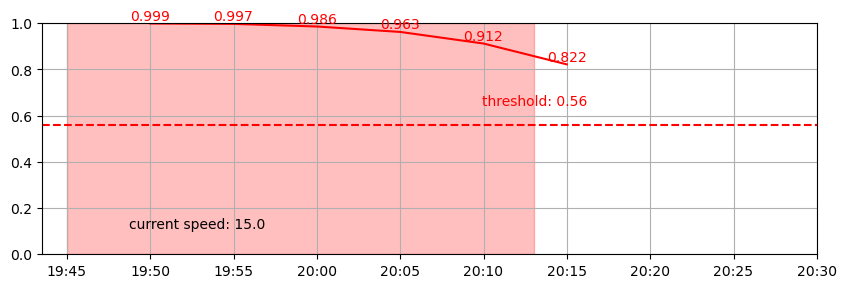}
        \caption{Prediction at 2023-01-25 19:45}
    \end{subfigure}
    \hfill
    \begin{subfigure}[t]{0.45\textwidth}
        \centering
        \includegraphics[width=\textwidth]{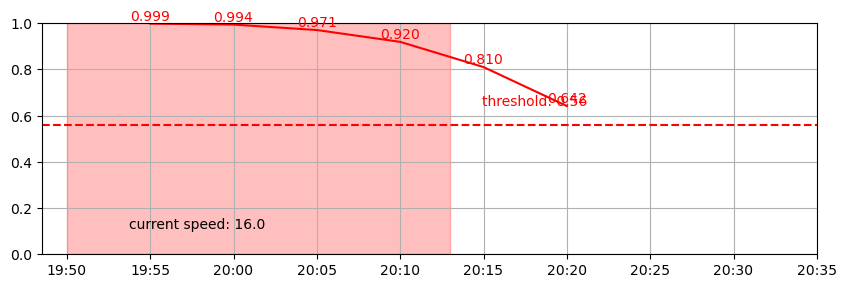}
        \caption{Prediction at 2023-01-25 19:50}
    \end{subfigure}
\end{figure}

\begin{figure}[H]
    \centering
    \captionsetup[subfigure]{labelformat=empty}
    \begin{subfigure}[t]{0.45\textwidth}
        \centering
        \includegraphics[width=\textwidth]{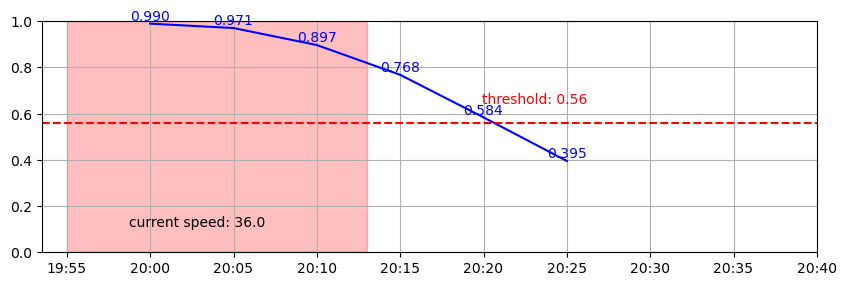}
        \caption{Prediction at 2023-01-25 19:55}
    \end{subfigure}
    \hfill
    \begin{subfigure}[t]{0.45\textwidth}
        \centering
        \includegraphics[width=\textwidth]{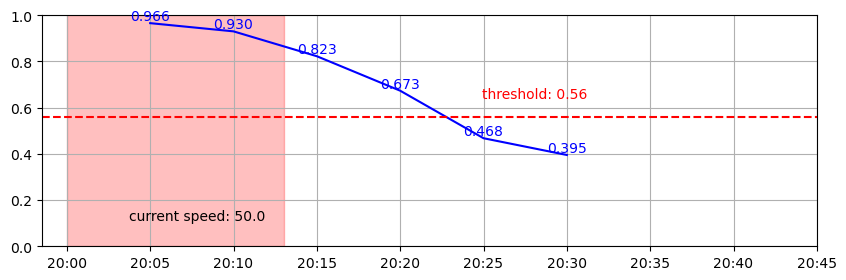}
        \caption{Prediction at 2023-01-25 20:00}
    \end{subfigure}
\end{figure}